\documentclass{article}

\usepackage[nonatbib,preprint]{neurips_2025}
\usepackage{graphicx}

\usepackage{wrapfig}

\usepackage[utf8]{inputenc} 
\usepackage[T1]{fontenc}    
\usepackage{hyperref}       
\usepackage{url}            
\usepackage{booktabs}       
\usepackage{amsfonts}       
\usepackage{nicefrac}       
\usepackage{microtype}      
\usepackage{xcolor}         
\usepackage{multirow}
\usepackage{amssymb}
\usepackage{pifont}
\usepackage{float}
\def\our{HuSc3D}

\title{\our{}: Human Sculpture dataset for 3D object reconstruction}

\newcommand{\cmark}{\ding{51}}%
\newcommand{\xmark}{\ding{55}}%

\author{%
  Weronika Smolak-Dyżewska\thanks{ \texttt{weronika.smolak@doctoral.uj.edu.pl}},
  Dawid Malarz, 
  Grzegorz Wilczyński, 
  Rafał Tobiasz,  
 \and
 \bf Joanna Waczy\'nska,  
  Piotr Borycki,  
  Przemysław Spurek\\
  Jagiellonian University\\
  IDEAS Research Institute\\
}

\begin{document}

\maketitle
\vspace{-0.9cm}
\begin{figure}[H]
    \centering
        \includegraphics[trim={150 50 150 50},clip, width=0.85\linewidth]{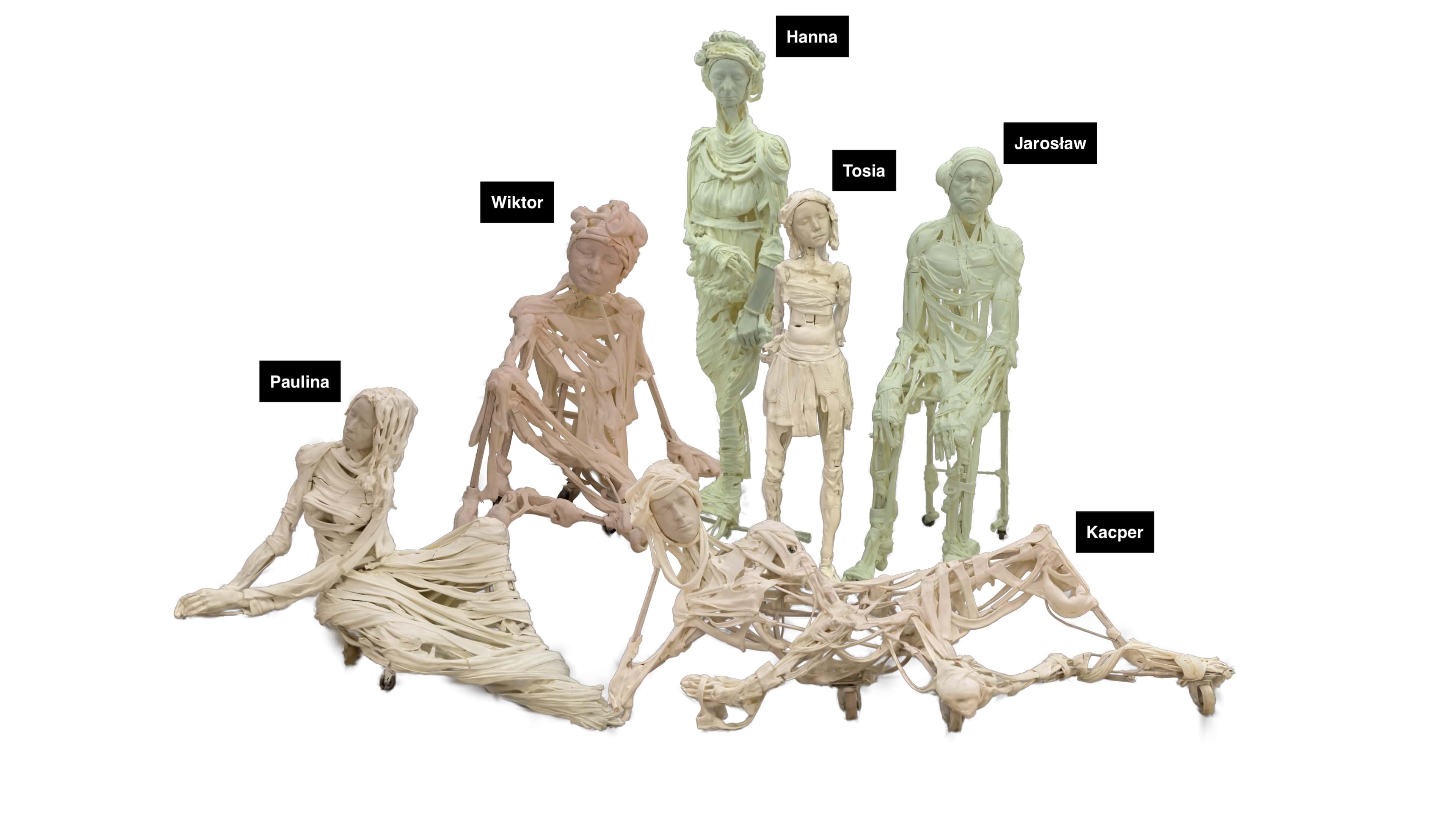}
        \vspace{-0.8cm}
    \caption{The \our{} dataset is designed for 3D reconstruction from 2D images, simulating a realistic acquisition process of an inexperienced user. The dataset includes six human-like sculptures created by Paweł Althamer. It is intended to support benchmarking and research in reconstruction tasks. It emphasizes challenges such as varying lighting conditions, dynamic background, and inconsistencies in image acquisition parameters.}
    \label{fig:demo}
     \vspace{-0.3cm}
\end{figure}

\begin{abstract}
3D scene reconstruction from 2D images is one of the most important tasks in computer graphics. Unfortunately, existing datasets and benchmarks concentrate on idealized synthetic or meticulously captured realistic data.  Such benchmarks fail to convey the inherent complexities encountered in newly acquired real-world scenes. In such scenes especially those acquired outside, the background is often dynamic, and by popular usage of cell phone cameras, there might be discrepancies in, e.g., white balance. To address this gap, we present \our{}, a novel dataset specifically designed for rigorous benchmarking of 3D reconstruction models under realistic acquisition challenges. Our dataset uniquely features six highly detailed, fully white sculptures characterized by intricate perforations and minimal textural and color variation. Furthermore, the number of images per scene varies significantly, introducing the additional challenge of limited training data for some instances alongside scenes with a standard number of views. By evaluating popular 3D reconstruction methods on this diverse dataset, we demonstrate the distinctiveness of \our{} in effectively differentiating model performance, particularly highlighting the sensitivity of methods to fine geometric details, color ambiguity, and varying data availability – limitations often masked by more conventional datasets. The code and dataset are available \href{https://wmito.github.io/HuSc3D/}{https://wmito.github.io/HuSc3D/}.
\end{abstract}

\section{Introduction}
The burgeoning field of 3D scene reconstruction has witnessed remarkable advancements, largely propelled by novel neural representations such as Neural Radiance Fields (NeRF) \cite{mildenhall2020nerf} and, more recently, 3D Gaussian Splatting (3DGS) \cite{kerbl3Dgaussians}. These methods have demonstrated impressive capabilities in generating photorealistic novel views and detailed 3D geometries from collections of 2D images. Consequently, a significant body of research now focuses on improving the fidelity, speed, and robustness of these techniques. Central to this progress is the availability of diverse and challenging datasets for training and, critically, for benchmarking the performance of these algorithms. 

However, majority of existing datasets, while valuable, present a somewhat sanitized view of the data acquisition process. Synthetic datasets, by their nature, offer perfect camera poses and controlled environments. Even datasets captured in the real world often undergo meticulous curation, extensive calibration, or are captured under studio-like conditions, resulting in data that is cleaner and more consistent than what an average user might produce. This discrepancy is particularly relevant as methods like 3DGS become more accessible, inviting users to capture scenes spontaneously with readily available devices like smartphones. Such "in-the-wild" captures, often intended for casual 3D reconstruction, inherently possess a different set of challenges not fully represented by current benchmarks.

To address this gap, we introduce \our{} dataset which consists of 6 scenes of white sculptures. The dataset was made by three different methods and showcases several challenges that the average user encounters when starting with the reconstruction task. 
Our dataset exhibits challenges such as:
\begin{itemize}
    \setlength\itemsep{-.1em}
    \item Low number of matched photos by COLMAP \cite{schoenberger2016mvs,schoenberger2016sfm} resulting in small training sample. 
    \item Difference in automatic white balance (AWB), especially when capturing as a video with a phone camera. 
    \item Non-static background elements like people passing behind when capturing in public spaces.
    \item Challenging objects of interest with numerous details but low contrast compared to the background.
\end{itemize}

\section{Related Works}

Currently, one of the most popular tasks in computer vision and graphics is reconstructing 3D scene from a series of 2D images. 
In recent years, the field has been revolutionized by learning-based approaches, particularly with the advent of Neural Radiance Fields (NeRF) \cite{mildenhall2020nerf} and 3D Gaussian Splatting (3DGS) \cite{kerbl3Dgaussians}, which have demonstrated state-of-the-art results in novel view synthesis and geometry representation. In the last years, there has been a surge in other methods based on NeRF \cite{multinerf2022,verbin2022refnerf,pumarola2020d,Hong_2022_CVPR,Deng_2022_CVPR} and 3DGS \cite{wu20234dgaussians,huang2023sc,waczynska2024d,Huang2DGS2024,MALARZ2025104273}. The progress and rigorous evaluation of such methods are intrinsically linked to the availability and nature of benchmarking datasets.

While numerous datasets exist, facilitating progress across various facets of 3D reconstruction, the majority of these benchmarks are synthetically created or with real data, but with professional acquisition. The discussion in this section is focused on a selection of widely recognized benchmarks. These chosen datasets are are particurarly relevant as they exemplify distinct types of data and associated challenges that have shaped the field, thereby providing context for the specific gap addressed by \our{}.

For instance, synthetic datasets like NeRF-Synthetic \cite{mildenhall2020nerf} represent the category of perfectly controlled environments. Usually, these scenes are created in Blender. They offer ideal camera parameters with their ground truth positions and object models. While invaluable for initial algorithm development and debugging, they inherently bypass real-world complexities like imperfect camera calibrations, imperfect camera position estimation made by COLMAP, noisy sensor data, lighting variations, and dynamic elements - issues that \our{} is designed to emphasize.

Moving to real-world data, datasets like the Mip-NeRF 360 dataset \cite{multinerf2022} exemplify the challenge of large-scale, unbounded scenes. This dataset significantly advanced the handling of 360-degree captures, presenting complex outdoor and indoor environments with detailed backgrounds. Its primary focus is on achieving high fidelity in large-scale reconstructions. In contrast, while also featuring real captures, \our{} concentrates on different complexities: specifically, objects with challenging material properties (highly detailed, low variety in texture, white sculptures) captured with consumer-grade devices, leading to artifacts like white balance inconsistencies and dynamic background.

\begin{figure}[t]
    \centering
    \renewcommand{\arraystretch}{0.2}
    \setlength{\tabcolsep}{0.4pt}
    \begin{tabular}{c@{}c@{}c@{}}
        \\[0.2cm]
        Jaroslaw &
        Kacper & 
        Wiktor \\
        
        \includegraphics[width=0.33\linewidth]{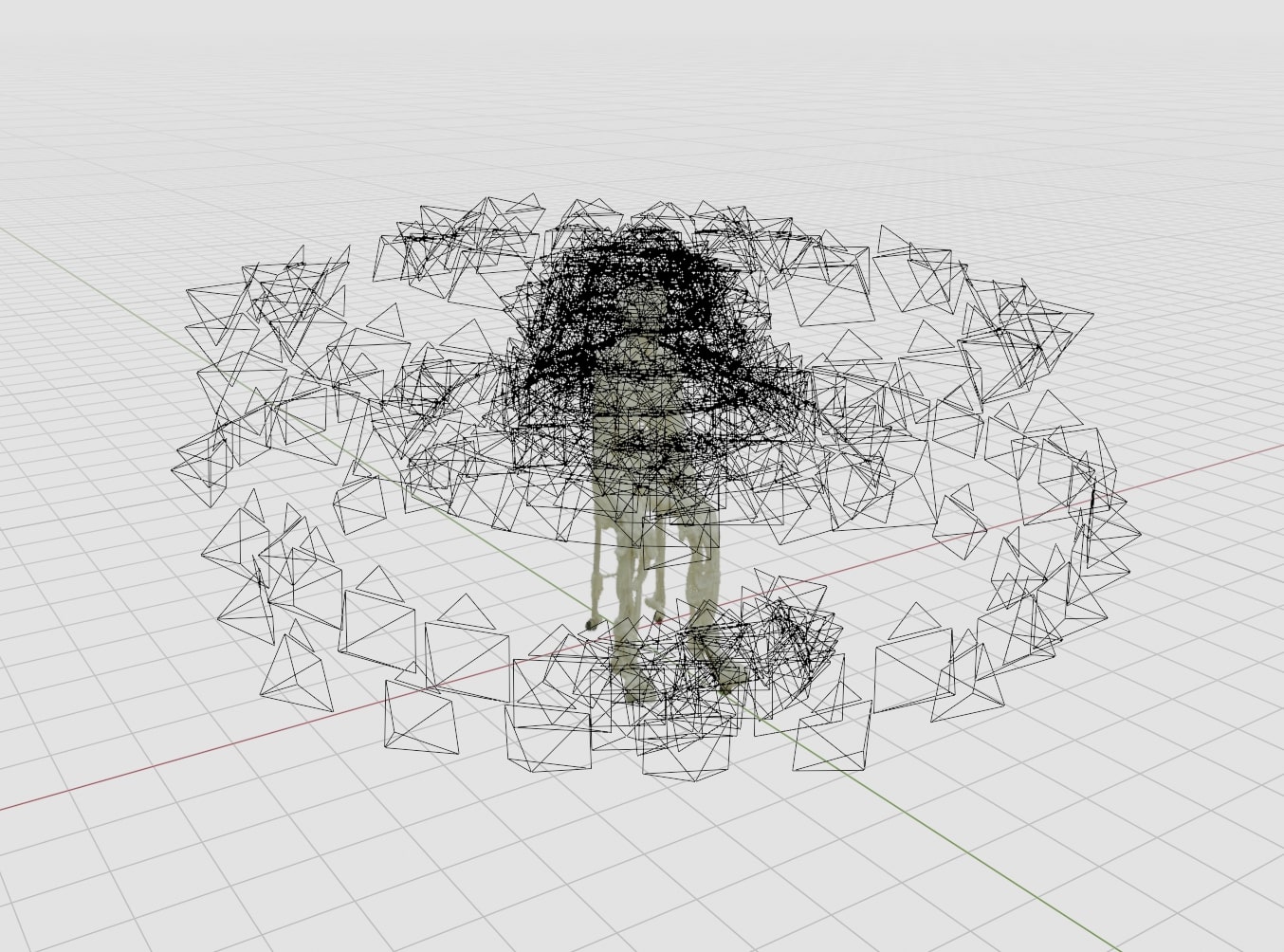} &       
        \includegraphics[width=0.33\linewidth]{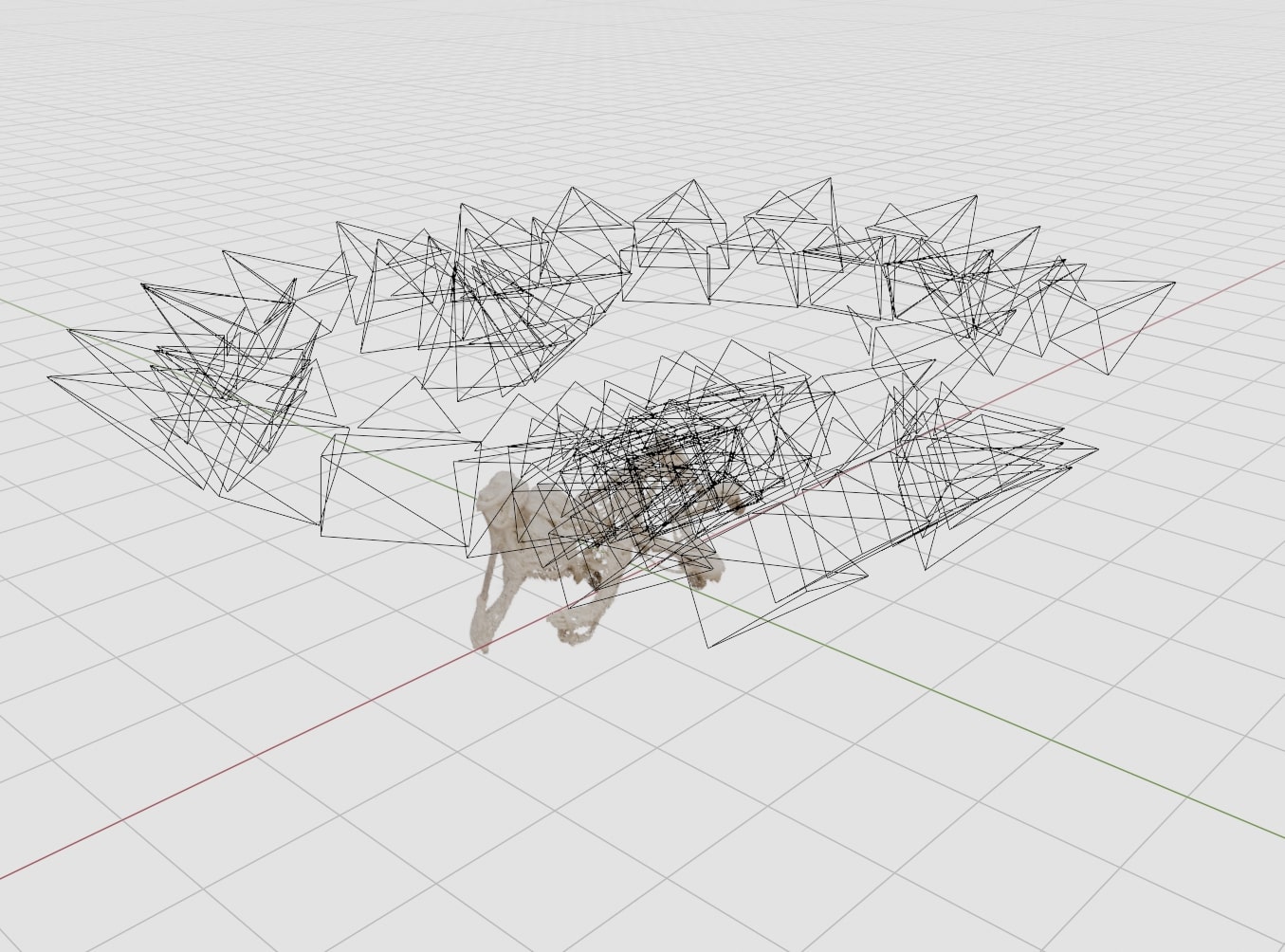} & 
        \includegraphics[width=0.33\linewidth]{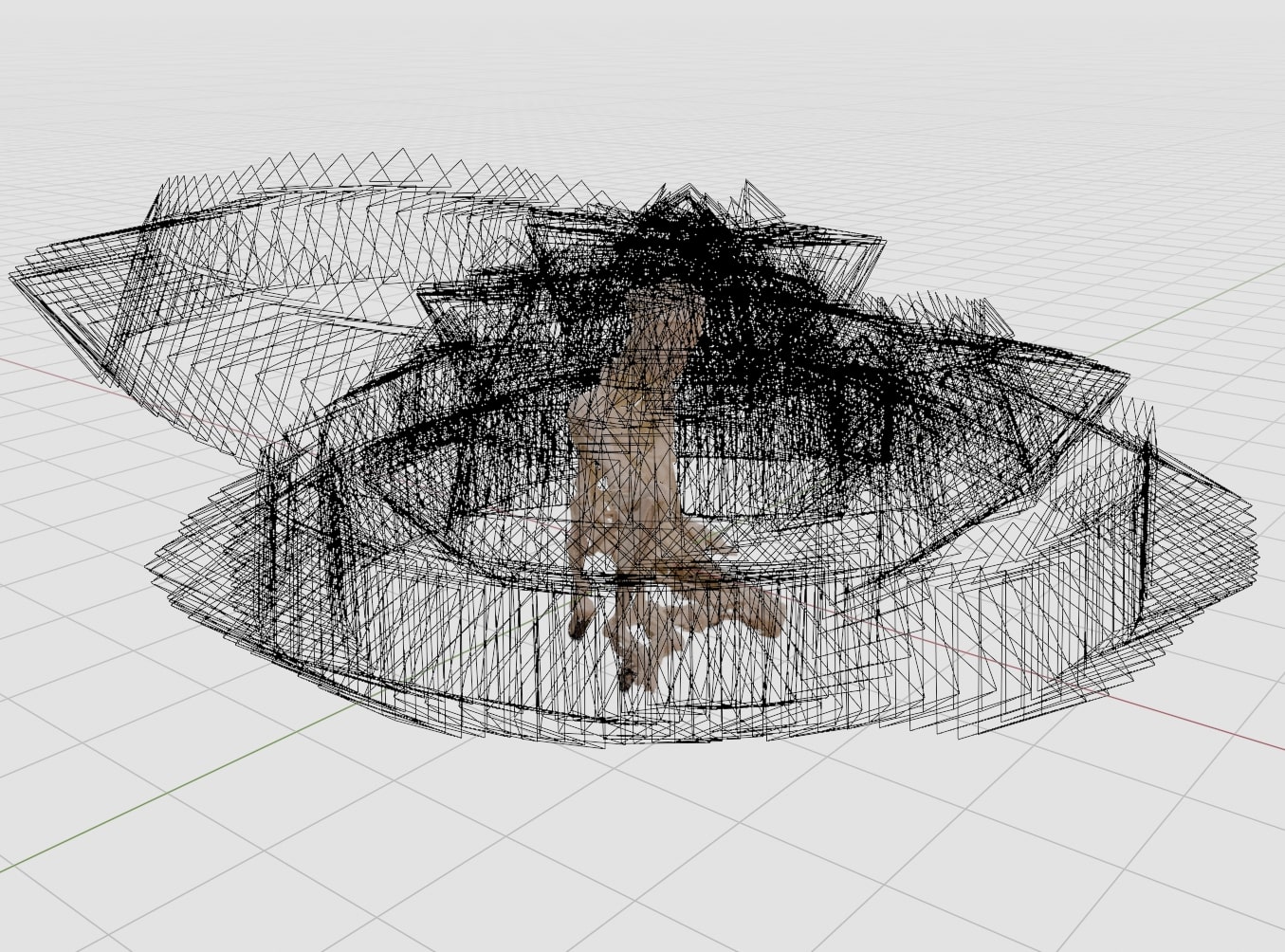}
        \\ [0.2cm]
        Hanna & Paulina & Tosia \\ [0.05cm]
        \includegraphics[width=0.33\linewidth]{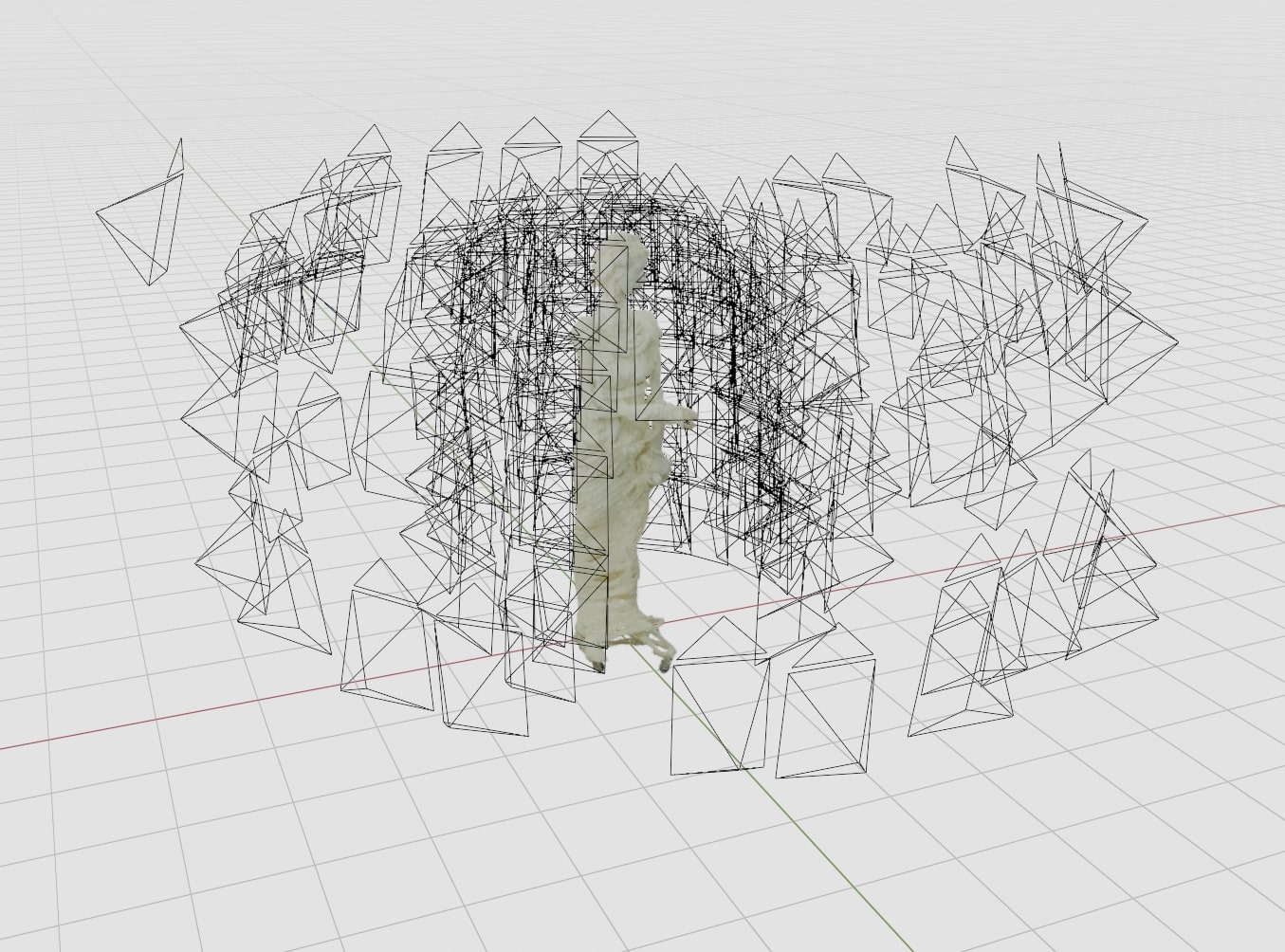} &         
        \includegraphics[width=0.33\linewidth]{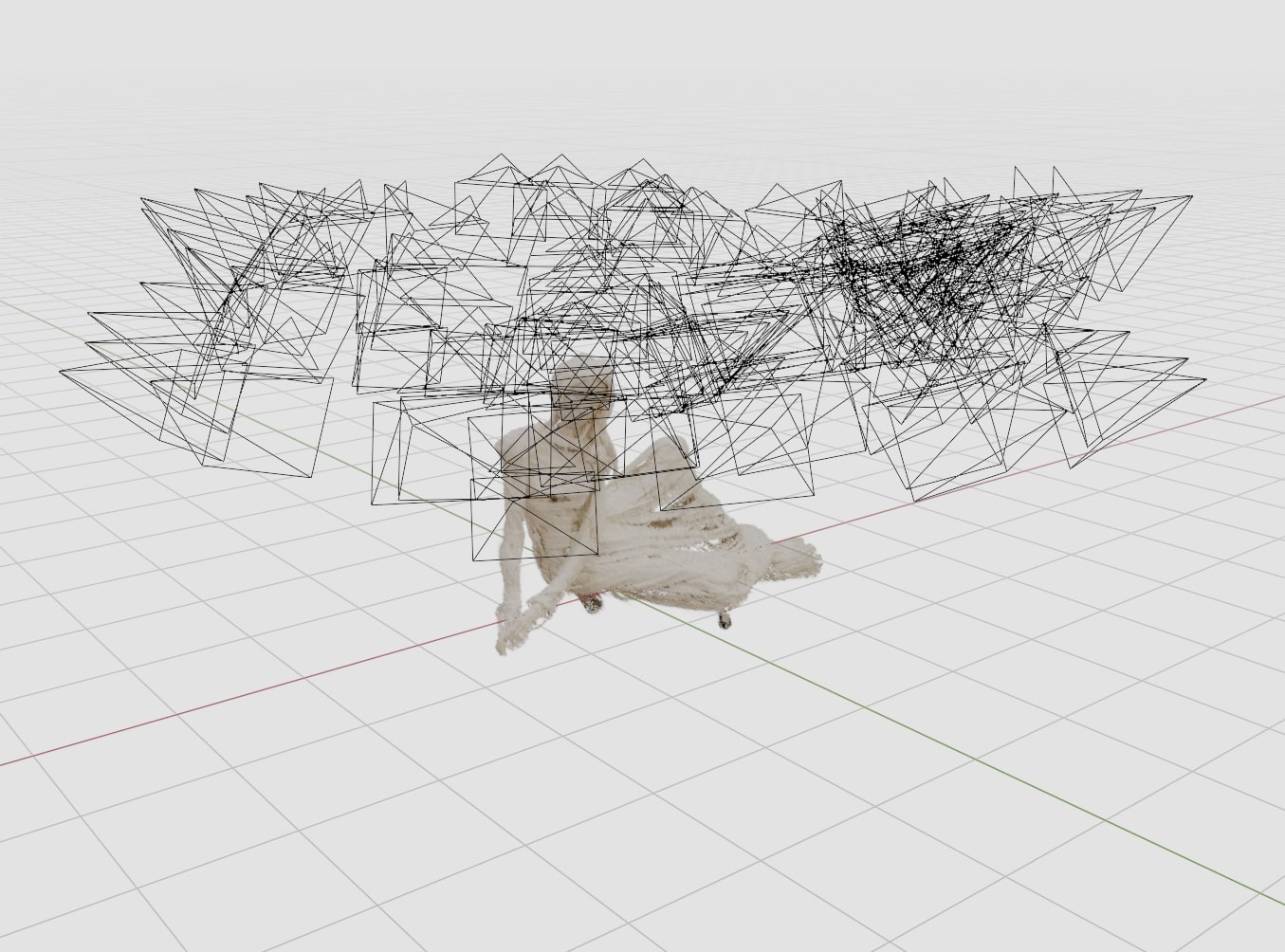} & 
        \includegraphics[width=0.33\linewidth]{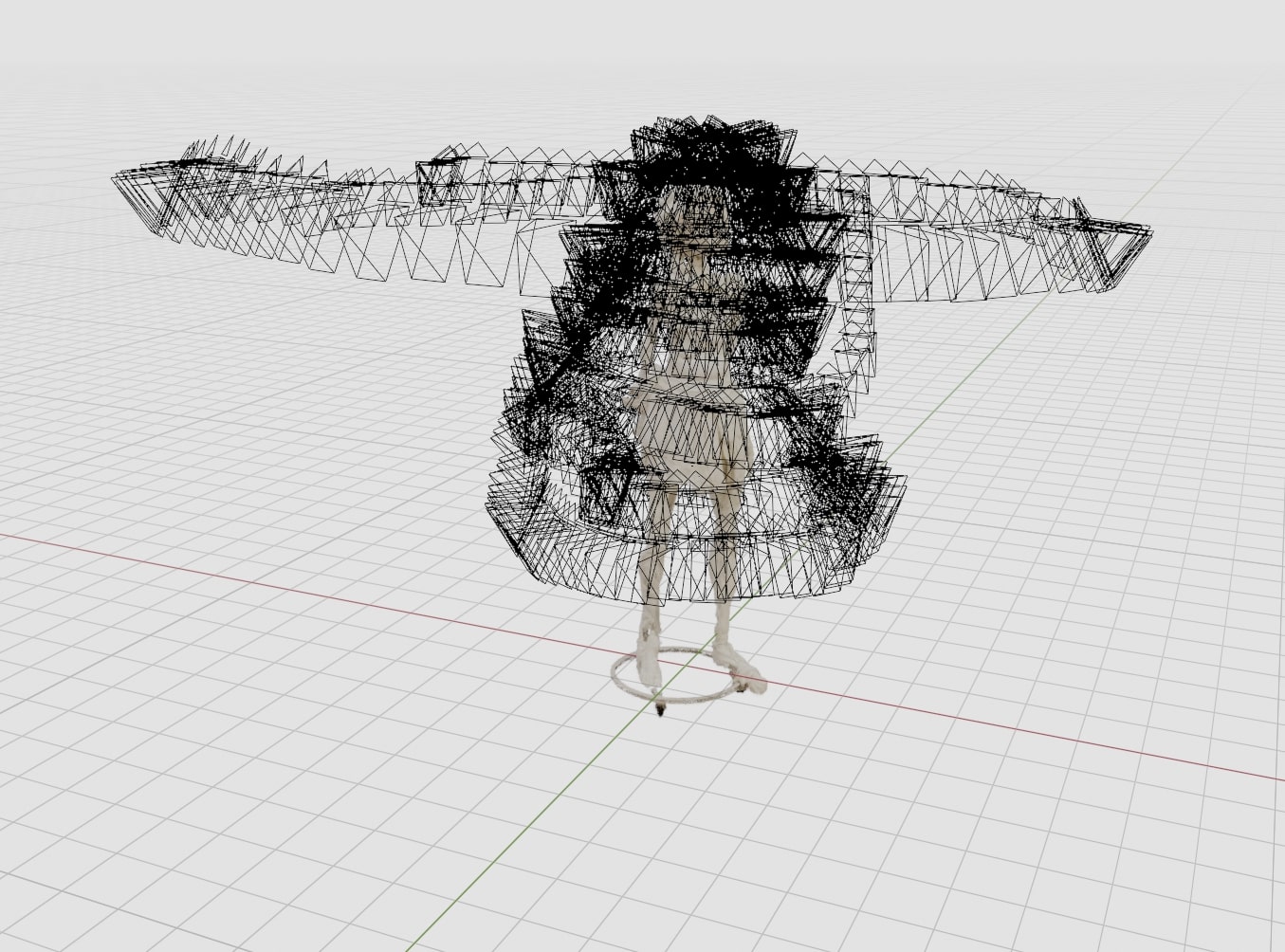} 
        
        \end{tabular}
    \caption{Visualisation of cameras in different scenes in \our{}. The visualisation was made in Blender \cite{blender} using addons \cite{PhotogrammetryForModeling2021,kiri}.}
    \label{fig:cameras}
\end{figure}

\begin{table*}[t]
\centering
\caption{The comparison between a size of \our{} and other common datasets used in 3D reconstruction task. Our method has a comparable number of scenes to other datasets (excluding CO3D). }
\label{tab:dataset_stats}
\resizebox{\textwidth}{!}{%
\begin{tabular}{l|cccc}
\toprule
\textbf{Dataset} & \textbf{No of scenes} & \textbf{Mean no of images per scene} & \textbf{Min no of } & \textbf{Max no of images per scene} \\
\midrule
\textbf{\our{}} & 6 & 330.7 & 66 & 680 \\
\textbf{Mip360} & 9 & 215.3 & 125 & 311 \\
\textbf{LLFF}   & 8 & 38.1  & 20 & 62 \\
\textbf{NeRF-Synthetic} & 8 & 400 & 400 & 400 \\
\textbf{CO3D} & 19k & depends$^*$ & depends$^*$ & depends$^*$ \\
\textbf{DeepBlending} & 19 & 244\textsuperscript{\textdaggerdbl} & 225\textsuperscript{\textdaggerdbl} & 263\textsuperscript{\textdaggerdbl} \\
\bottomrule
\multicolumn{5}{l}{\textsuperscript{*}\footnotesize We did not include statistics for CO3D since it is a big dataset that is divided into different classes.} \\
\multicolumn{5}{l}{\textsuperscript{\textdaggerdbl}\footnotesize We included statistics for only two scenes that are usually used in benchmarking 3D reconstruction task, that is Playroom and DrJohnson.} \\
\end{tabular}%
}
\end{table*}

Moving to real-world data, datasets like the Mip-NeRF 360 dataset \cite{multinerf2022} exemplify the challenge of large-scale, unbounded scenes. This dataset significantly advanced the handling of 360-degree captures, presenting complex outdoor and indoor environments with detailed backgrounds. Its primary focus is on achieving high fidelity in large-scale reconstructions. In contrast, while also featuring real captures, \our{} concentrates on different complexities: specifically, objects with challenging material properties (highly detailed, low variety in texture, white sculptures) captured with consumer-grade devices, leading to artifacts like white balance inconsistencies and dynamic background.

The LLFF (Local Light Field Fusion) dataset \cite{mildenhall2019llff} is representative of benchmarks focusing on reconstruction from sparse, forward-facing views captured with handheld cameras. This highlighted the challenge of generating plausible results from limited input. \our{} incorporates sparsity (e.g. Kacper and Paulina scenes) but frames it within a 360 view compared to only forward-facing views in LLFF. 

Large-scale, category-centric datasets are exemplified by CO3D (Common Objects in 3D) \cite{reizenstein21co3d}. Providing numerous multi-view sequences across many object categories with point cloud annotations, it was a significant progress in category-specific reconstruction. 

Finally, datasets featuring controlled indoor environments, such as the commonly used scenes from DeepBlending \cite{HPPFDB18}, represent scenarios with artificial lighting. They are not object-centered and more taggle a problem of reconstructing whole rooms.

In summary, while the selected representative datasets have driven significant advancements by targeting specific challenges like synthetic perfection, scale, sparsity, category-specific reconstruction, or reconstructing whole rooms, \our{} fills a crucial niche. Currently, the method of 3D reconstruction came so far that we believe that their progress should be focused on tailoring toward the inexperienced end-user. 

\section{\our{} Dataset}

This section details the acquisition process and the inherent characteristics of \our{} that make it a distinctive and realistic benchmark for the 3D reconstruction task.
\paragraph{Data acquisition and challenges}
The \our{} dataset consists of six scenes (see Figure \ref{fig:demo}, each featuring a different sculpture made by the Polish sculptor Paweł Althamer. The dataset was acquired in the Foksal Gallery Foundation headquarters. Captured sculptures were made in 2012 and are a part of a Almech project. All of sculptures are white with multiple perforations. In the background, there are printed and taped AprilTags from the tag36h11 family with a 10 cm width. Among the dataset there are three ways of capturing the data with camera positions presented in Figure \ref{fig:cameras}.

\begin{wrapfigure}[16]{r}{0.48\textwidth}
\vspace{-0.5cm}
\centering
    \renewcommand{\arraystretch}{0}
    \setlength{\tabcolsep}{0.4pt}
    \begin{tabular}{c@{}c@{}}
        Paulina & Kacper \\ \vspace{0.1cm} \\
        \includegraphics[width=0.48\linewidth]{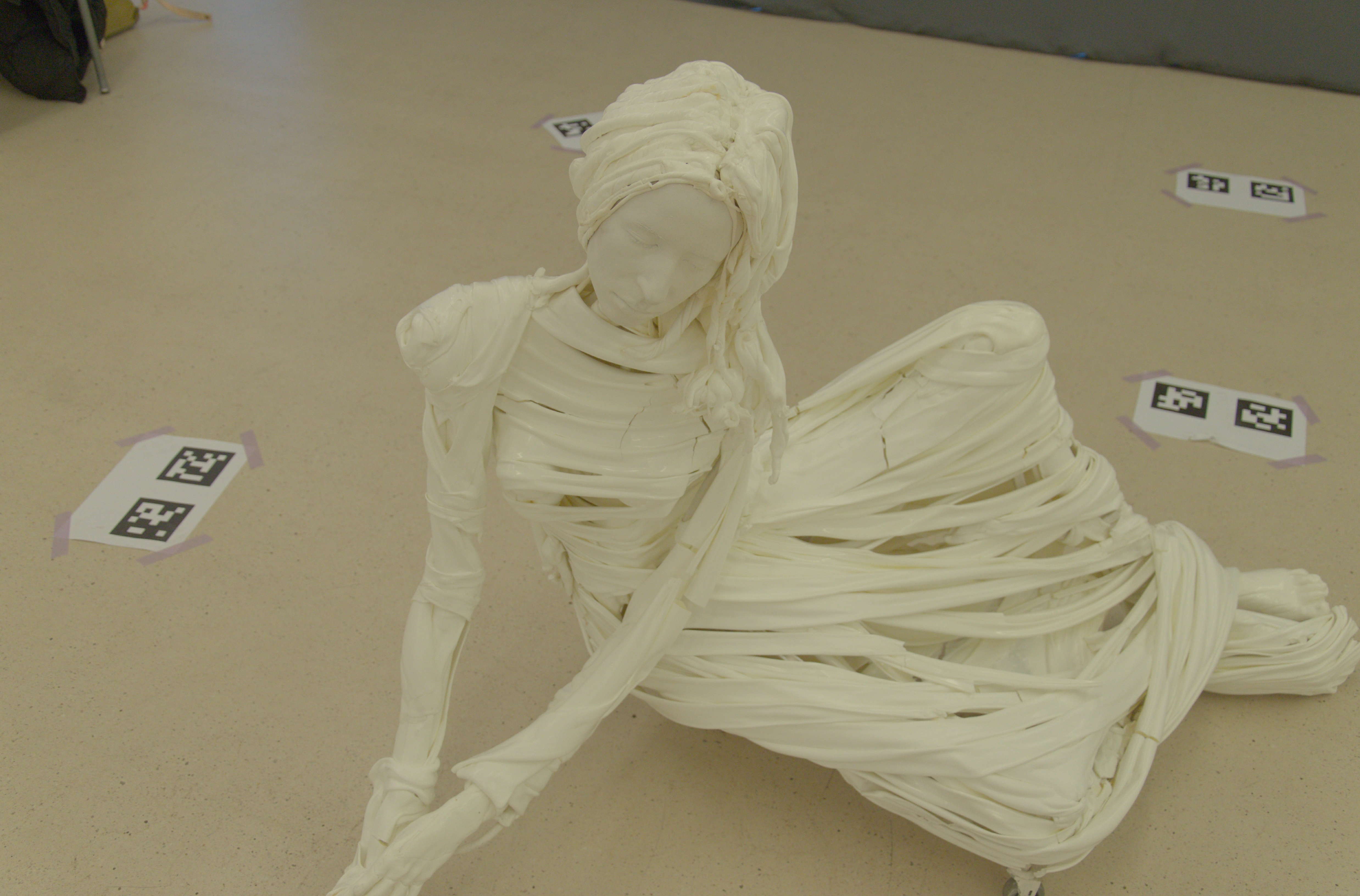} &
        \includegraphics[width=0.48\linewidth]{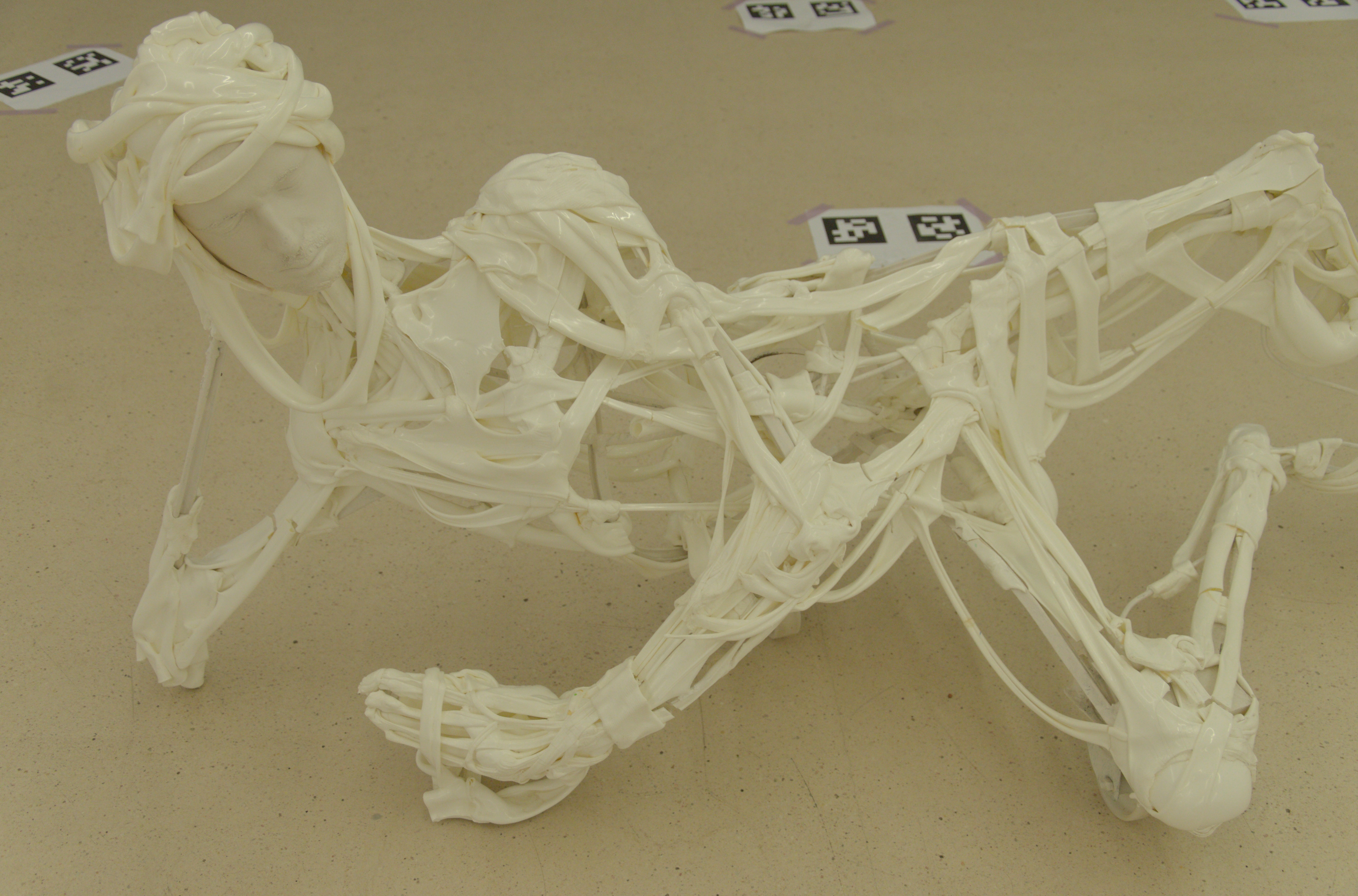}  \\
        \includegraphics[width=0.48\linewidth]{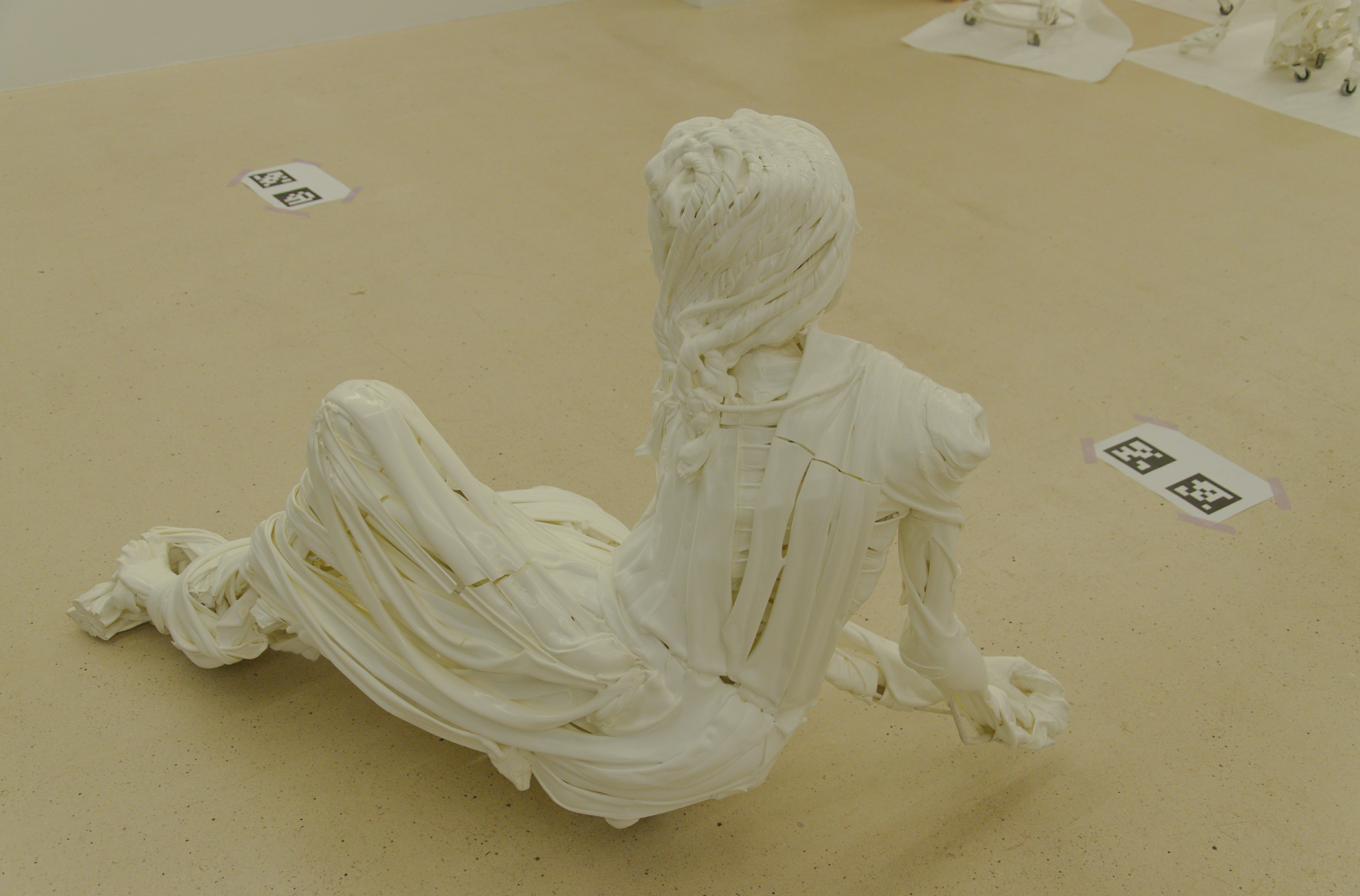} &
        \includegraphics[width=0.48\linewidth]{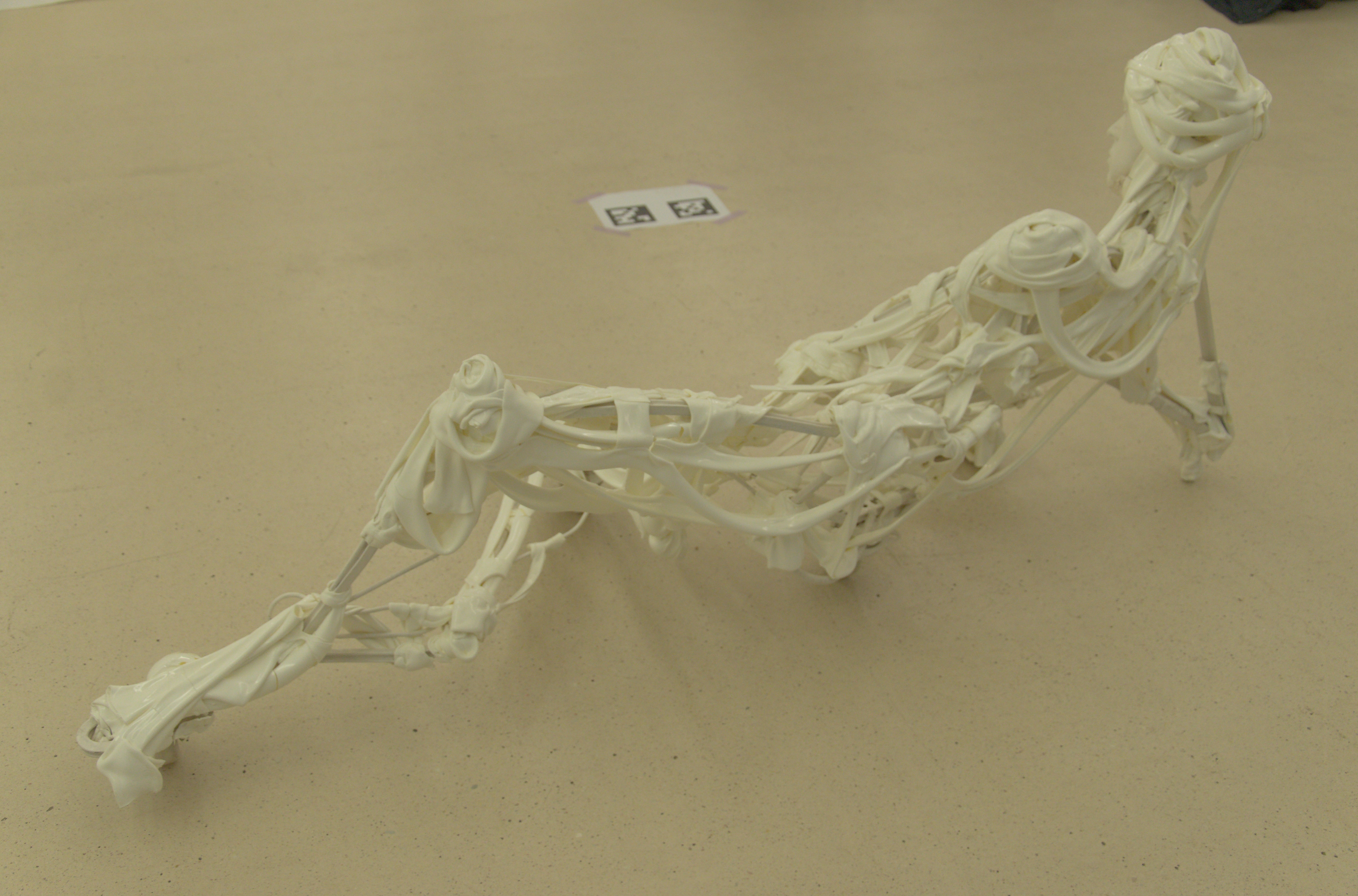} 
        \\  
    \end{tabular}
    \caption{Examples of Paulina and Kacper scenes' images.}
    \label{fig:paulina_kacper}
\end{wrapfigure}

\begin{figure}{}
    \centering
    \renewcommand{\arraystretch}{0}
    \setlength{\tabcolsep}{0.4pt}
    \begin{tabular}{c@{}c@{}c@{}c@{}}
    GT  & Splatfacto-MCMC & Nerfacto \\
        \includegraphics[width=0.33\linewidth]{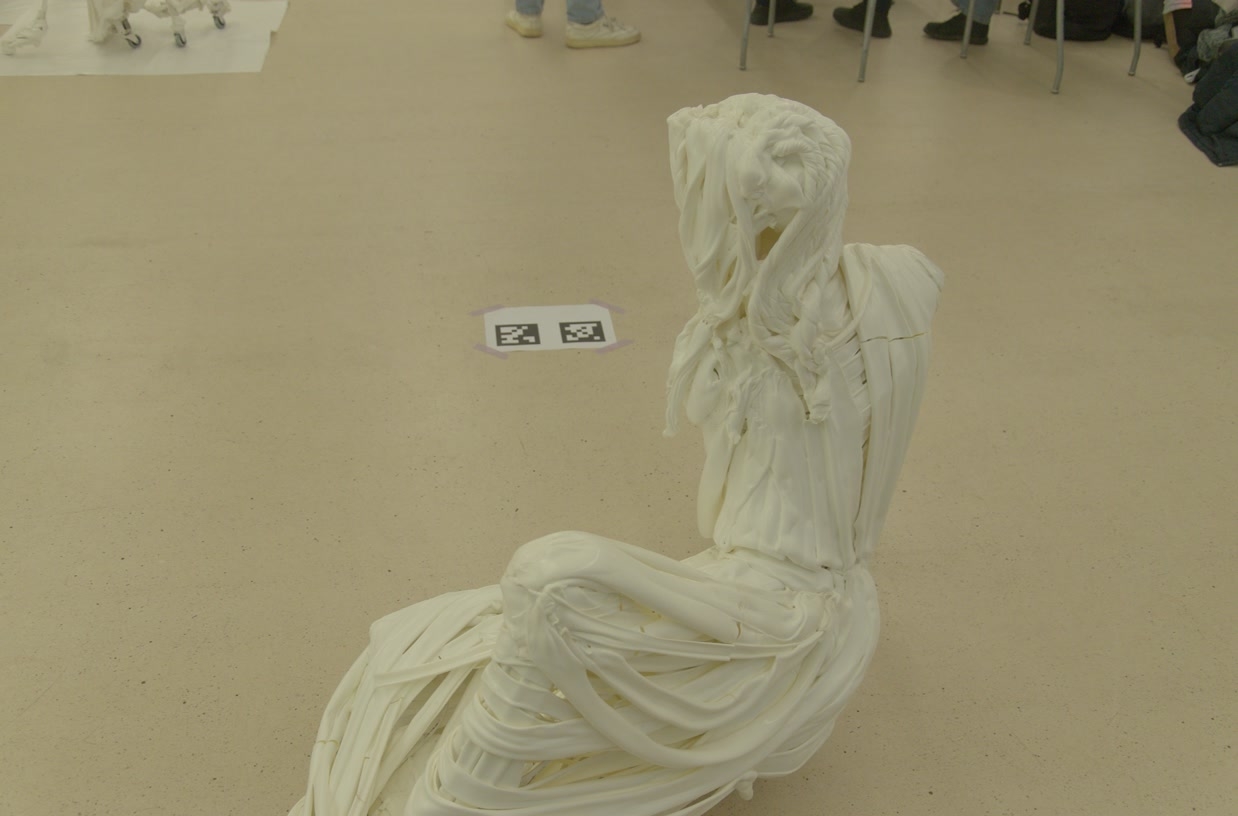} &
        \includegraphics[width=0.33\linewidth]{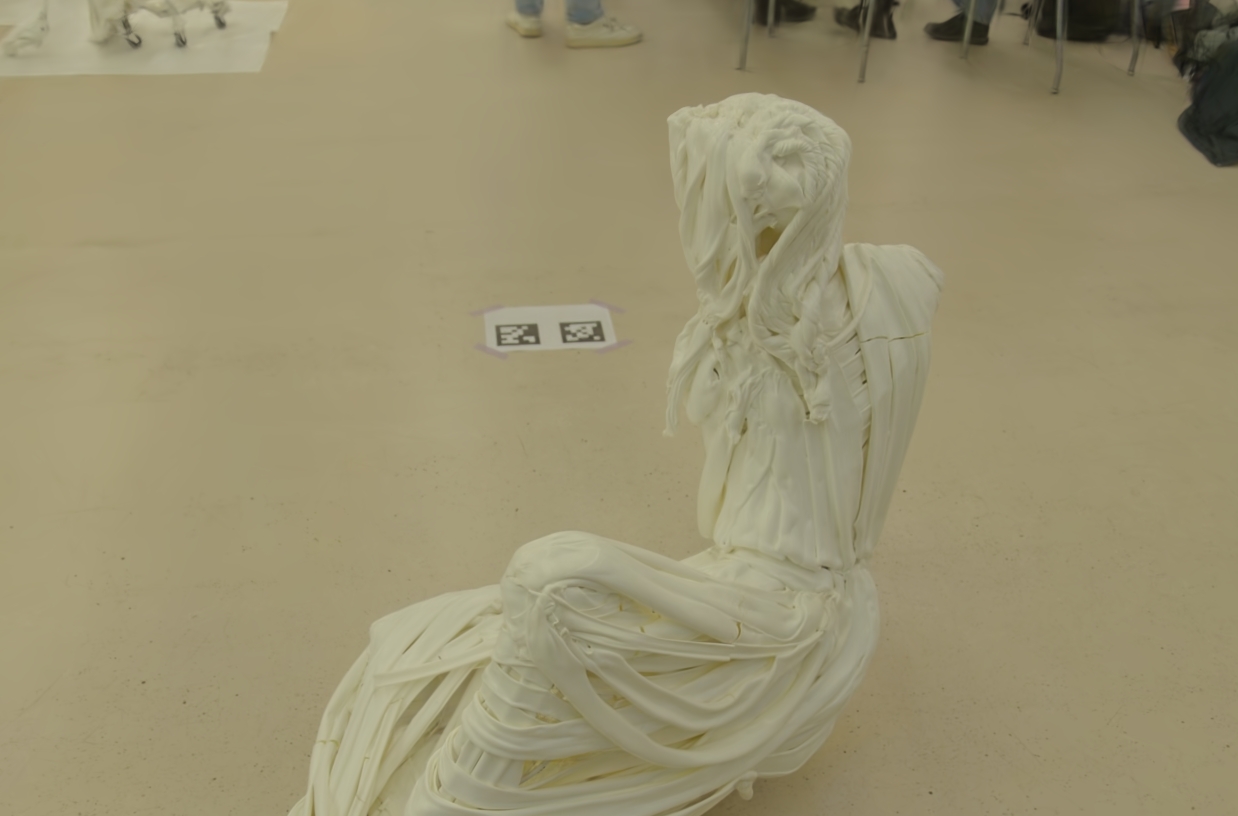}  &
        \includegraphics[width=0.33\linewidth]{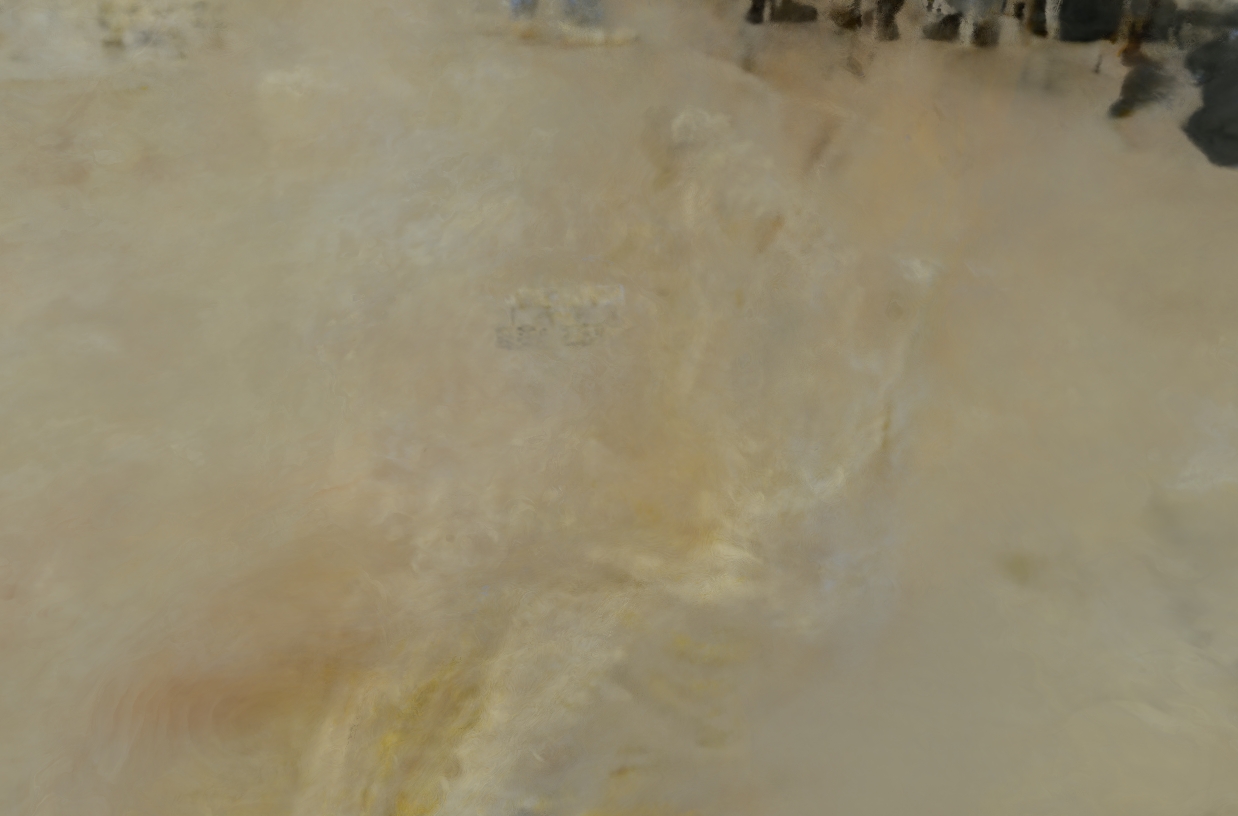} 
        
    \end{tabular}
    \caption{In scenes with low training samples (i.e. Paulina scene) methods based on NeRF struggle with reconstruction.}
    \label{fig:nerfacto_low_samples}
\end{figure}

\paragraph{Kacper and Paulina scenes} To simulate challenges arising from data acquisition limitations, the Kacper and Paulina scenes introduce the difficulty of sparse input views. These sets were made with photos from NICON D7000 and AF-S DX Nikkor 18-105mm f/3.5-5.6G ED VR lens using fixed ISO, aperture and shutter speed. The autofocus was locked between sets of photos. The main challenge of this part of the dataset is the low number of photos which were chosen by preprocessing using SfM via COLMAP. The final sets consist of 66 photos for Kacper and 104 for Paulina (see Figure \ref{fig:paulina_kacper}). This sparsity presents a significant challenge, demanding that reconstruction algorithms effectively leverage limited information to achieve high-fidelity results, particularly in capturing the intricate details characteristic of our sculptures. Especially, methods such as NeRF proved to be extremely sensitive to this challenge yielding the worst results (see Table \ref{tab:results} and Figure \ref{fig:nerfacto_low_samples}, especially when compared to Splatfacto-MCMC.

\begin{figure}[t]
    \centering
    \renewcommand{\arraystretch}{0}
    \setlength{\tabcolsep}{0.5pt}
    \begin{tabular}{c@{}c@{}c@{}c@{}}
    GT before resizing & GT after resizing & Splatfacto-MCMC \\
        \includegraphics[width=0.33\linewidth]{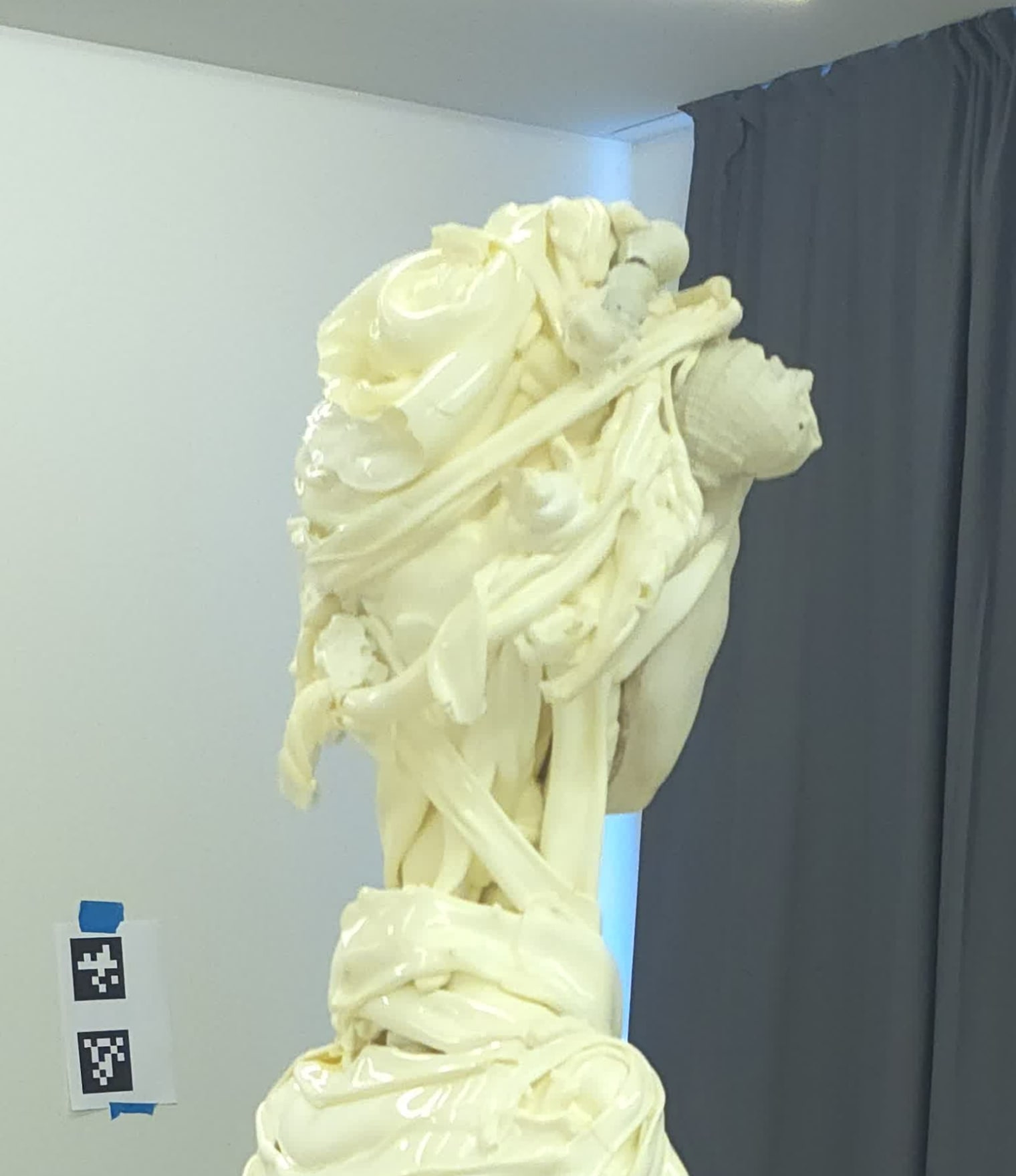} &
        \includegraphics[width=0.33\linewidth]{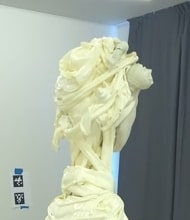}  &
        \includegraphics[width=0.33\linewidth]{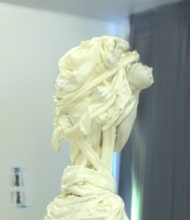} 
        
    \end{tabular}
    \caption{The necessary 8x downsizing of input images led to substantial detail loss, preventing even the best model from achieving satisfactory reconstruction fidelity. Ideally, the 3D reconstruction models should be able to handle subsampling of such data to avoid data loss.}
    \label{fig:hanna_resize}
\end{figure}

\paragraph{Jarosław and Hanna scenes} Two exemplary scenes within \our{} underscore a critical bottleneck in contemporary 3D reconstruction: the effective utilization of readily available ultra-high-resolution imagery. Photos were made with Samsung Galaxy S24 Ultra phone from a handheld position with PRO mode on with fixed ISO, aperture and shutter speed. This sets consists of 348 (Jarosław) and 190 (Hanna) high-resolution photos 6004x8018 px and 8017x6005 px consecutively.  While such images promise exceptional detail, their practical application is severely hampered by computational demands. In our experiments, an 8x downscaling was crucial for feasibility, inherently discarding significant fine-grained information. These scenes thus serve as a crucial stress test, demanding that reconstruction methods either pioneer techniques for efficient high-resolution processing or demonstrate resilience to information loss inherent in aggressive but necessary downsampling. Ideally, the methods used for reconstruction task should somehow account for large images to make it more feasible to create highly detailed reconstructions on private computers. Even after looking at results of Splatfacto-MCMC which acquired the highest scores, it still lost a great level of detail, see Figure~\ref{fig:hanna_resize}.

\newpage

\begin{wrapfigure}[19]{l}{0.4\textwidth}
\vspace{-0.5cm}   
\centering
    \renewcommand{\arraystretch}{0}
    \setlength{\tabcolsep}{0.4pt}
    \begin{tabular}{c@{}c@{}}
        Superior AWB & Incorrect AWB \\ \vspace{0.1cm} \\ 
        \includegraphics[width=0.48\linewidth]{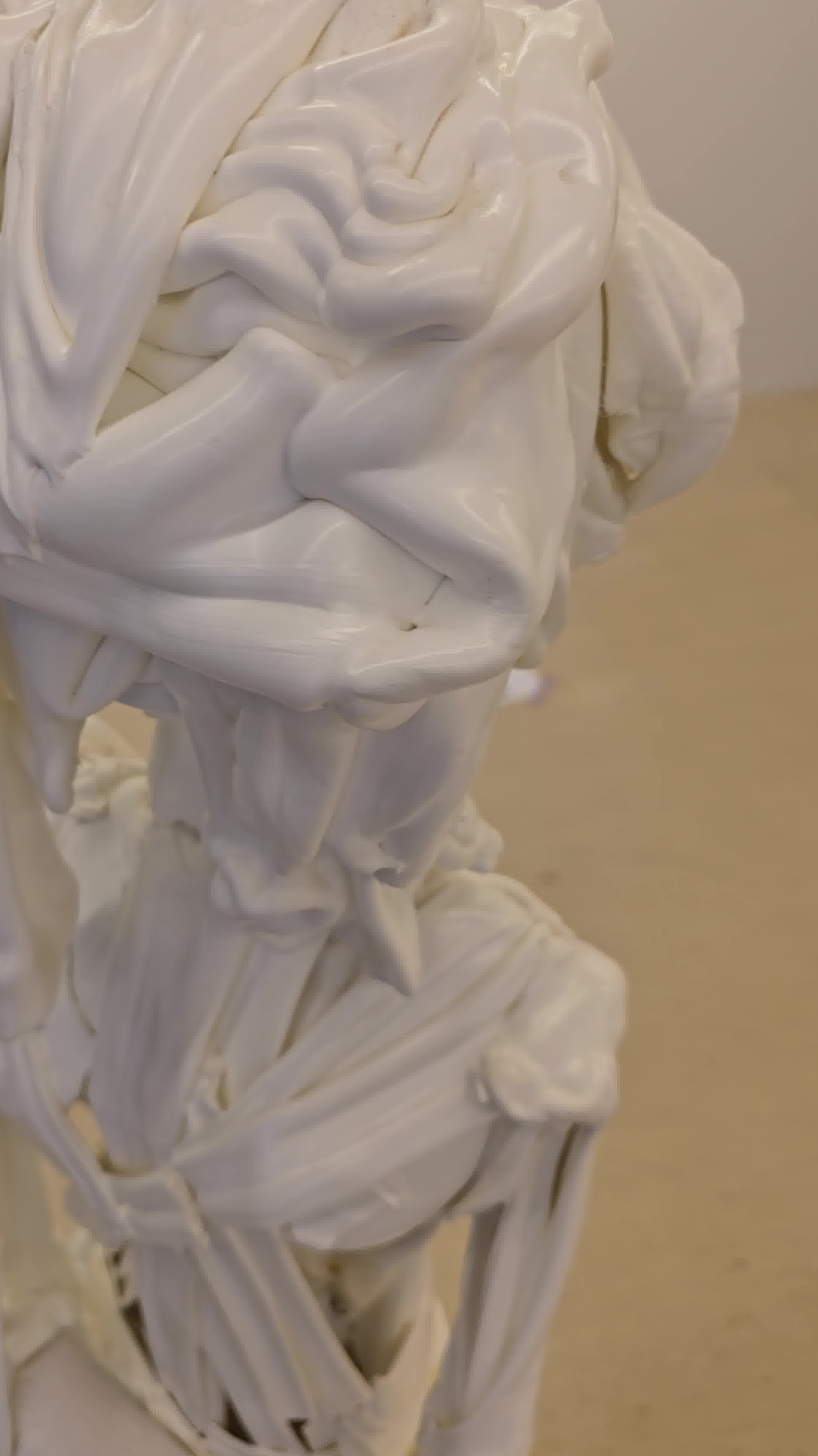} &
        \includegraphics[width=0.48\linewidth]{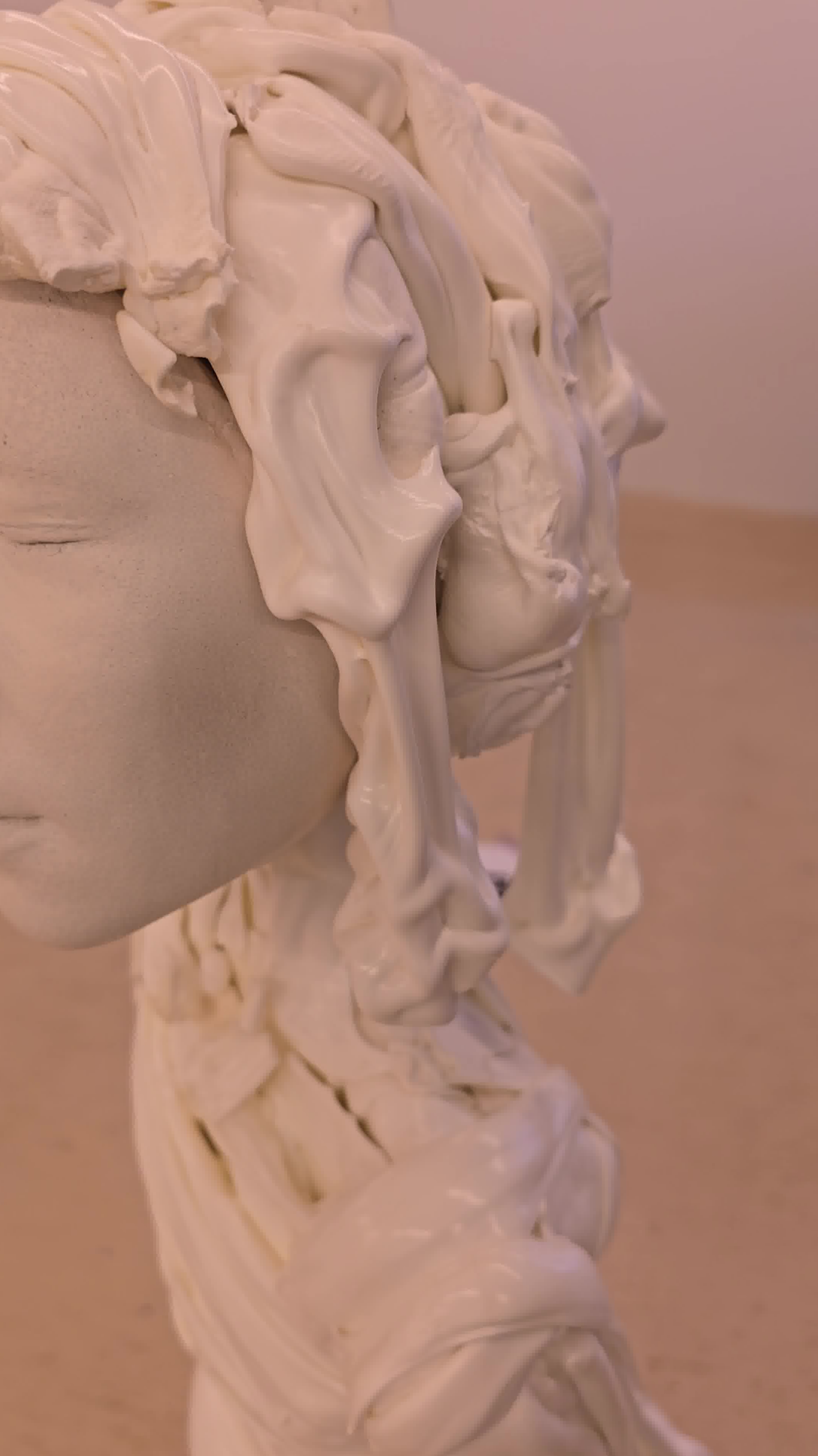} \\
    \end{tabular}
    \caption{Example of variety in automatic white balance (AWB) in Tosia scene. Especially, when considering the wall in the background, a human eye can find incorrect AWB photos easily.}
    \label{fig:white_balance}
\end{wrapfigure} 

\paragraph{Tosia and Wiktor scenes} To simulate common acquisition issues with current mobile devices and most common method, two scenes were created using video from a Samsung Galaxy S24 Ultra. From these videos, 680 and 596 images were respectively obtained by extracting every 20th frame. A significant challenge introduced by these scenes is the photometric inconsistency caused by the phone's automatic white balance (AWB) system. During capture, this system, typically unalterable by the end-user, dynamically shifted the white balance, leading to frames with varying color casts, notably a reddish tint in some instances. This directly reflects real-world scenarios where on-device post-processing introduces data variability that reconstruction methods must handle. Tosia scene has 13.2\% of scenes with misleading white balance and Wiktor 28.4\%. The photos with wrong white balance can be easily distinguished by the human eye (see Figure~\ref{fig:white_balance}) while reconstruction algorithms would have required prior manual preprocessing of input data.


\section{Experiments}

To benchmark performance on \our{}, six widely adopted 3D reconstruction algorithms were evaluated: 2DGS \cite{Huang2DGS2024}, Mip-Splatting \cite{Yu2024GOF}, NeRF (implementation Nerfacto from Nerfstudio \cite{nerfstudio}), 3DGS (implementation Splatfacto from Nerfstudio), 3DGS-MCMC (implementation Splatfacto-MCMC from Nerfstudio) \cite{kheradmand20243d}, and Instant-NGP \cite{mueller2022instant} (implementation from Nerfstudio). These methods were selected as representative baselines in the field, encompassing diverse paradigms ranging from explicit point-based rendering to implicit volumetric representations.

\paragraph{Experimental Setup}

Reconstruction quality and efficiency were assessed using the following metrics: PSNR (Peak Signal-to-Noise Ratio, SSIM (Structural Similarity Index), LPIPS (Learned Perceptual Image Patch Similarity), training time, and FPS.


All methods are implemented using publicly available codebases with minimal modifications to ensure reproducibility. For 2DGS and Mip-Splatting, implementations from their original repositories were used, retaining default hyperparameters as prescribed by the authors. Similarly, Nerfacto, Splatfacto, Splatfacto-MCMC, and Instant-NGP are executed via the Nerfstudio framework, adhering to its default configurations. All scenes are trained for 30,000 iterations, following the community-wide paradigm. To balance computational efficiency and reconstruction fidelity, input images are downsampled by factor of 4 for most scenes. For Hanna and Jaroslaw scenes a downsampling factor of 8 is applied. Experiments are conducted on an NVIDIA RTX 4090 GPU with 256 GB of system memory, running Ubuntu 24.04 LTS.

Following the protocol of 3DGS 12.5\% of images from each scene are reserved as a test set. Validation metrics (SSIM, PSNR, LPIPS, FPS) are computed on this subset. 

\begin{table*}[t]
\centering
\caption{Quantitative results across \our{}. Best scores per metric are bolded. OOM = Out-of-Memory. Time is represented in format minutes:seconds. Splatfacto-MCMC achieved superior qualitative metrics, however, it had the longest training time. Additionally, Splatfacto-MCMC had significantly smaller FPS compared to Mip-Splatting, 2DGS and Splatfacto. We did not include average score for Instant-NGP due to OOM for 2 out of 6 scenes.}
\label{tab:results}
\resizebox{\textwidth}{!}{%
\begin{tabular}{l|c|cccccc|c}
\toprule
\textbf{Method} & \textbf{Metric} & \textbf{Jaroslaw} & \textbf{Hanna} & \textbf{Kacper} & \textbf{Paulina} & \textbf{Tosia} & \textbf{Wiktor} & \textbf{Avg} \\
\midrule
\multirow{5}{*}{\textbf{2DGS}} 
& SSIM & 0.917 & 0.936 & 0.887 & 0.895 & 0.942 & 0.944 & 0.920 \\
& PSNR & 28.95 & 29.60 & 28.13 & 30.96 & 30.83 & 31.29 & 29.96 \\
& LPIPS & 0.235 & 0.184 & 0.312 & 0.295 & 0.169 & 0.138 & 0.222 \\
& Time & 11:17 & 11:20 & 11:27 & 11:27 & 7:32 & \textbf{8:20} & 10:21 \\
& FPS & 151.21 & 132.86 & \textbf{163.18} & \textbf{189.26} & 260.42 & 226.08 & 187.17 \\
\midrule

\multirow{5}{*}{\textbf{Mip-Splatting}} 
& SSIM & 0.923 & 0.934 & 0.898 & 0.915 & 0.954 & 0.958 & 0.930 \\
& PSNR & 29.23 & 29.01 & 28.12 & 31.62 & 32.46 & 33.25 & 30.62 \\
& LPIPS & 0.211 & 0.176 & 0.290 & 0.252 & 0.128 & 0.093 & 0.192 \\
& Time & 14:44 & 11:42 & 11:20 & 10:48 & 10:14 & 10:06 & 11:29 \\
& FPS & \textbf{183.06} & 134.29 & 126.28 & 157.90 & \textbf{296.39} & \textbf{252.50} & \textbf{191.74} \\
\midrule

\multirow{5}{*}{\textbf{Nerfacto}} 
& SSIM & 0.857 & 0.829 & 0.797 & 0.856 & 0.730 & 0.770 & 0.807 \\
& PSNR & 22.19 & 21.80 & 22.32 & 25.16 & 17.10 & 23.30 & 21.98 \\
& LPIPS & 0.285 & 0.222 & 0.457 & 0.322 & 0.637 & 0.233 & 0.359 \\
& Time & 13:12 & 11:25 & 12:40 & 13:12 & 11:15 & 11:29 & 12:12 \\
& FPS & 1.28 & 1.35 & 0.89 & 0.86 & 1.94 & 2.04 & 1.39 \\
\midrule

\multirow{5}{*}{\textbf{Splatfacto}} 
& SSIM & 0.924 & 0.937 & 0.942 & 0.956 & 0.954 & 0.952 & 0.944 \\
& PSNR & 29.23 & 29.52 & 31.23 & 32.44 & 33.23 & 32.63 & 31.38 \\
& LPIPS & 0.162 & 0.115 & 0.183 & 0.154 & 0.110 & 0.090 & 0.136 \\
& Time & \textbf{7:03} & \textbf{6:19} & \textbf{7:45} & \textbf{7:36} & \textbf{6:21} & 10:45 & \textbf{7:38} \\
& FPS & 177.57 & \textbf{189.21} & 136.97 & 134.88 & 182.02 & 173.64 & 165.71 \\
\midrule

\multirow{5}{*}{\textbf{Splatfacto-MCMC}} 
& SSIM & \textbf{0.931} & \textbf{0.942} & \textbf{0.957} & \textbf{0.966} & \textbf{0.960} & \textbf{0.960} & \textbf{0.953} \\
& PSNR & \textbf{29.82} & \textbf{30.12} & \textbf{32.93} & \textbf{34.55} & \textbf{33.67} & \textbf{33.62} & \textbf{32.45} \\
& LPIPS & \textbf{0.139} & \textbf{0.104} & \textbf{0.125} & \textbf{0.113} & \textbf{0.100} & \textbf{0.074} & \textbf{0.109} \\
& Time & 14:08 & 13:53 & 17:33 & 19:24 & 13:34 & 18:01 & 16:06 \\
& FPS & 94.95 & 83.43 & 60.26 & 56.56 & 100.69 & 98.13 & 82.34 \\
\midrule

\multirow{5}{*}{\textbf{Instant-NGP}} 
& SSIM & 0.885 & OOM & 0.927 & OOM & 0.909 & 0.912 & -- \\
& PSNR & 24.96 & OOM & 29.85 & OOM & 27.87 & 27.06 & -- \\
& LPIPS & 0.321 & OOM & 0.205 & OOM & 0.216 & 0.181 & -- \\
& Time & 9:48 & OOM & 9:44 & OOM & 8:51 & 9:48 & -- \\
& FPS & 0.88 & OOM & 0.60 & OOM & 1.76 & 1.67 & -- \\
\bottomrule
\end{tabular}%
}
\end{table*}




\begin{wrapfigure}{l}{0.5\textwidth}
    \centering
        \centering
    \hspace{0.02\linewidth}
    \begin{minipage}[c]{\linewidth} 
        \centering
        \renewcommand{\arraystretch}{0.15} 
        \setlength{\tabcolsep}{1pt}     

         \begin{tabular}{@{}c@{}c@{\;}c@{}c@{}c@{}}
            & & Jaroslaw & Kacper & Paulina \\
            
            &
            \rotatebox{90}{\ \ \ \ \ \  \small GT}  &
            \includegraphics[width=0.28\linewidth]{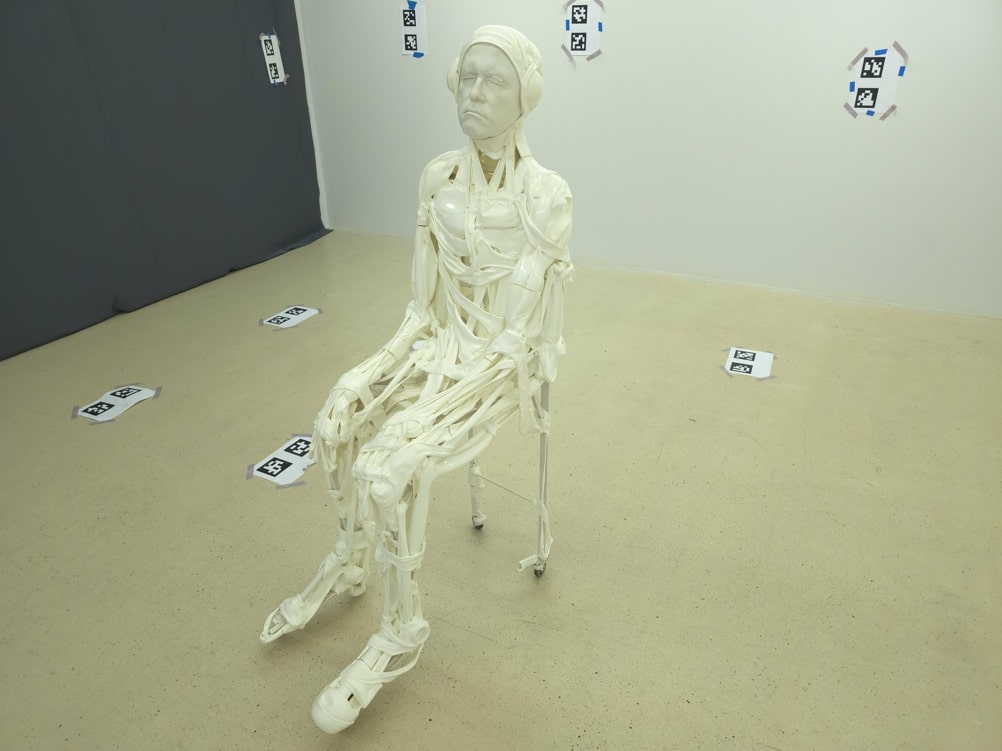} &
            \includegraphics[width=0.32\linewidth]{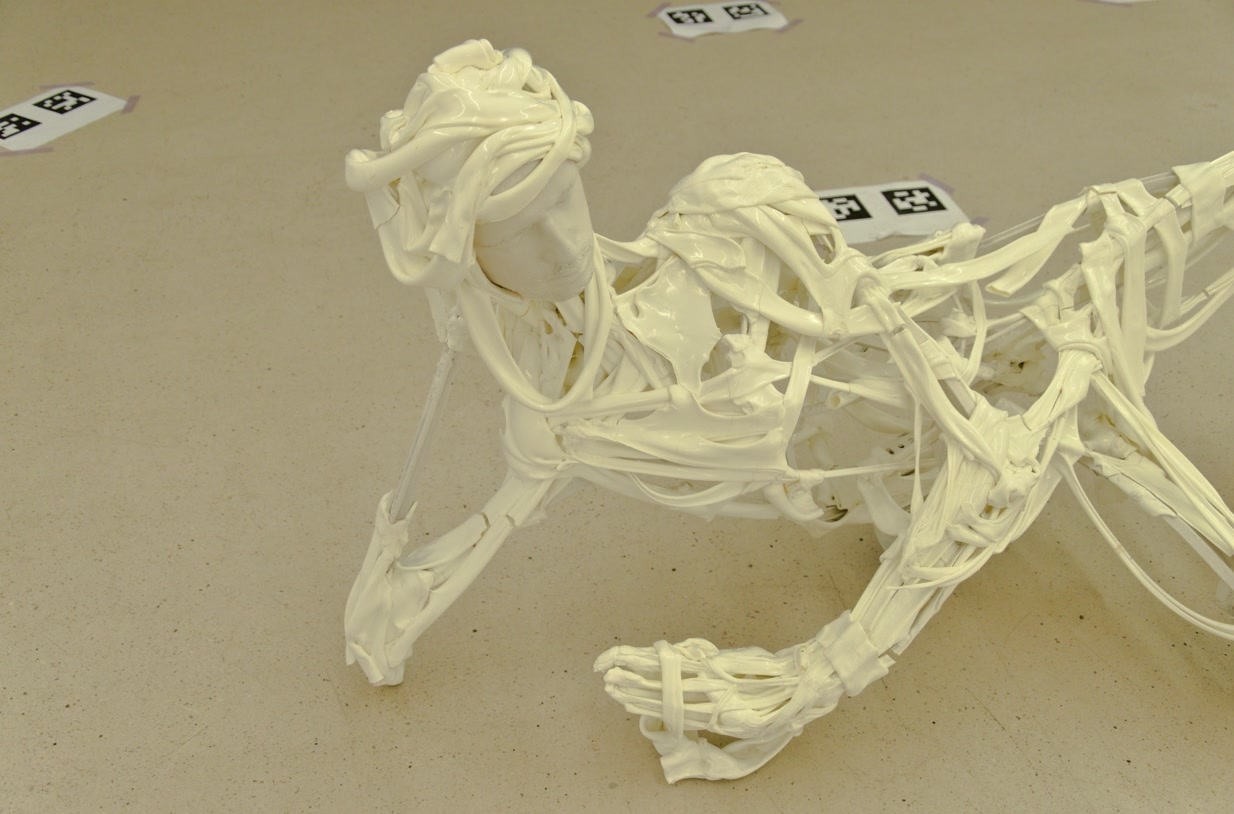} &
            \includegraphics[width=0.32\linewidth]{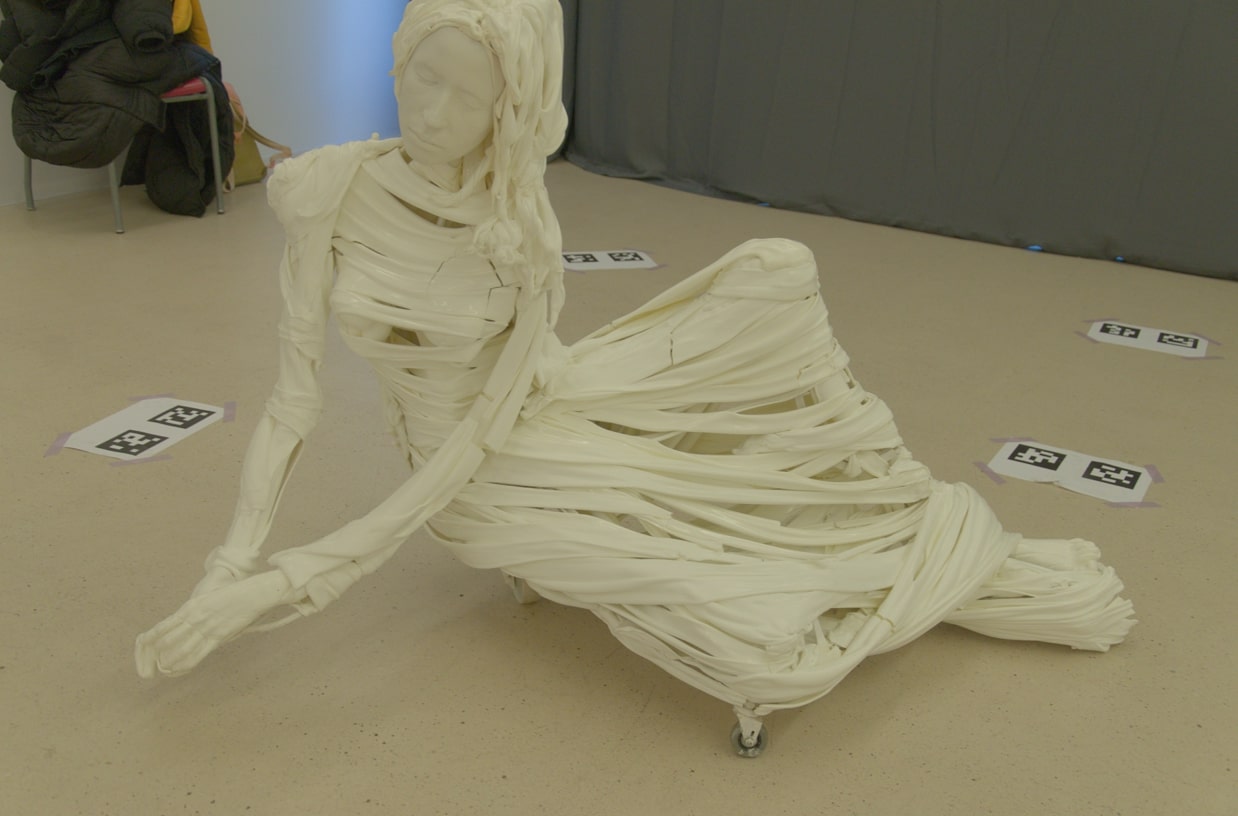} \\

            &
            \rotatebox{90}{\ \ \ \small 2DGS} &
            \includegraphics[width=0.28\linewidth]{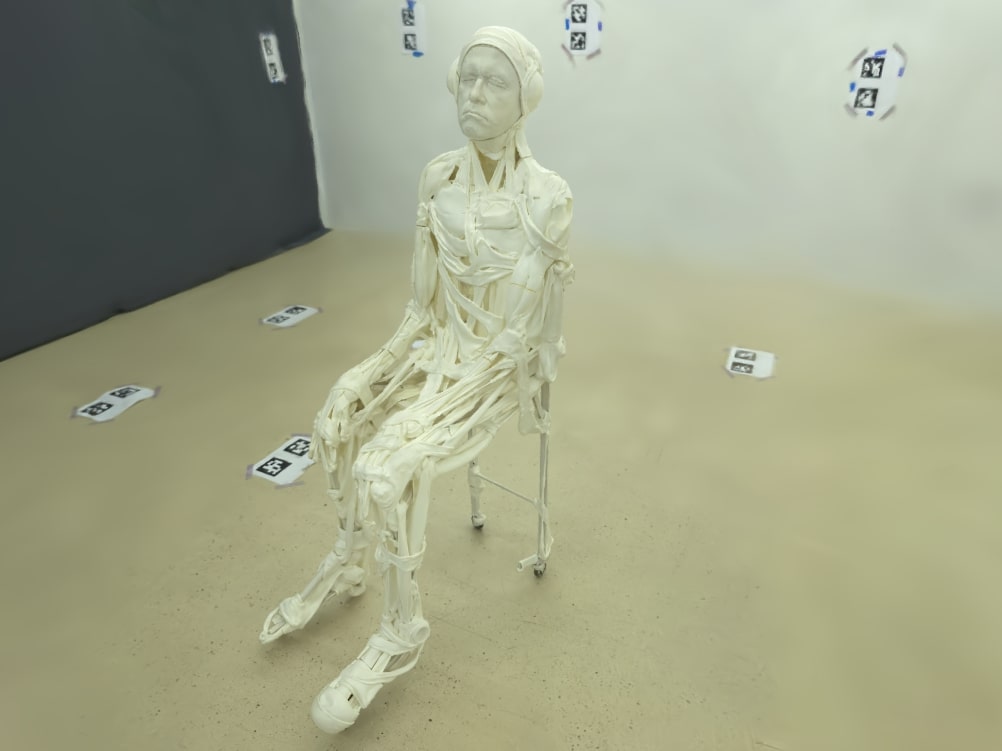} &
            \includegraphics[width=0.32\linewidth]{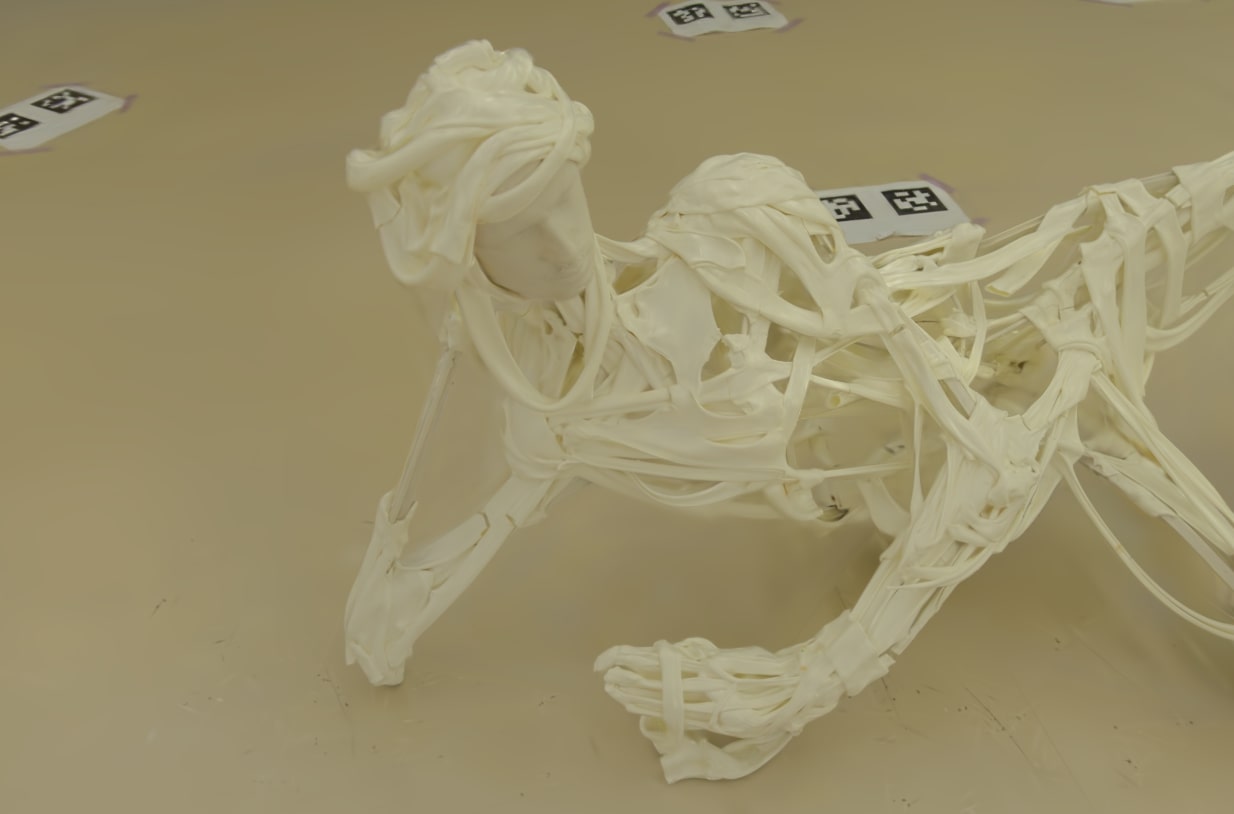} &
            \includegraphics[width=0.32\linewidth]{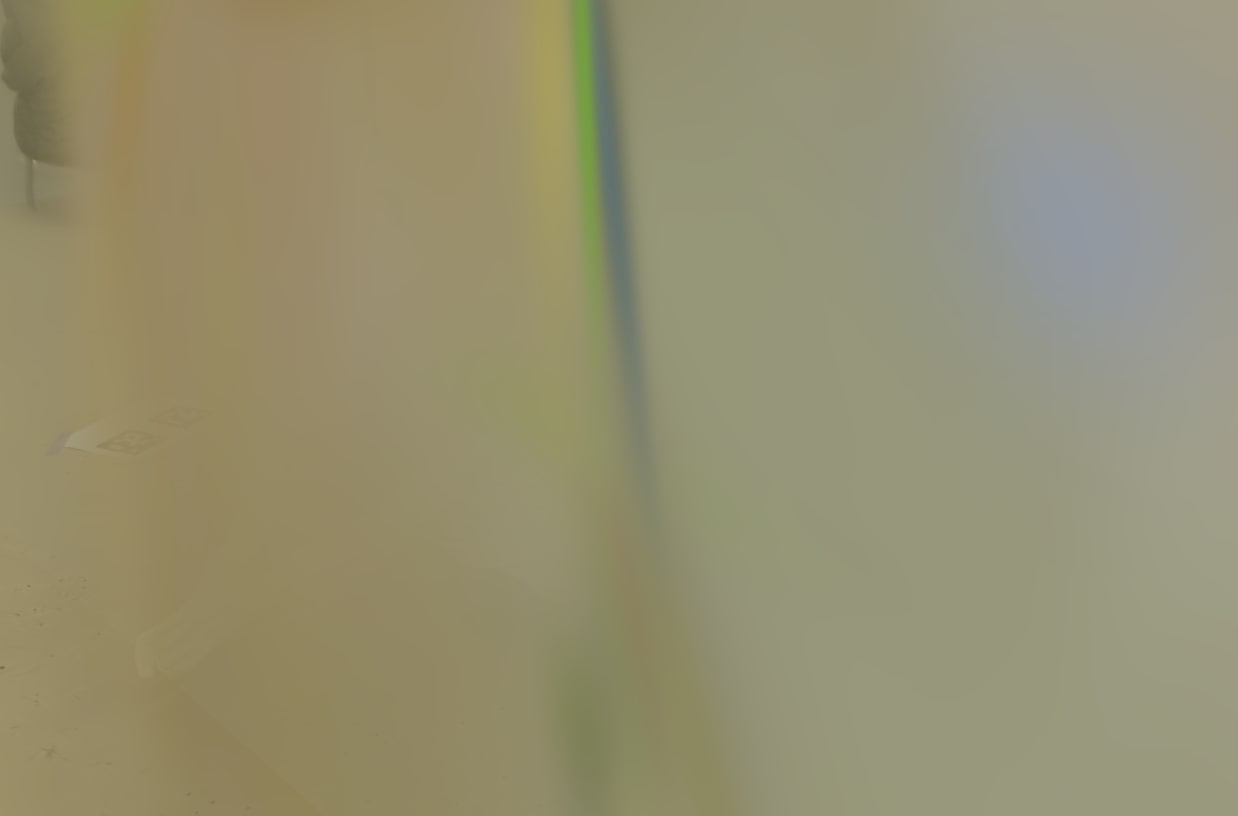} \\

            
            \rotatebox{90}{\ \quad \small Mip}
            &
            \rotatebox{90}{\ \ \small Splatting} & 
            \includegraphics[width=0.28\linewidth]{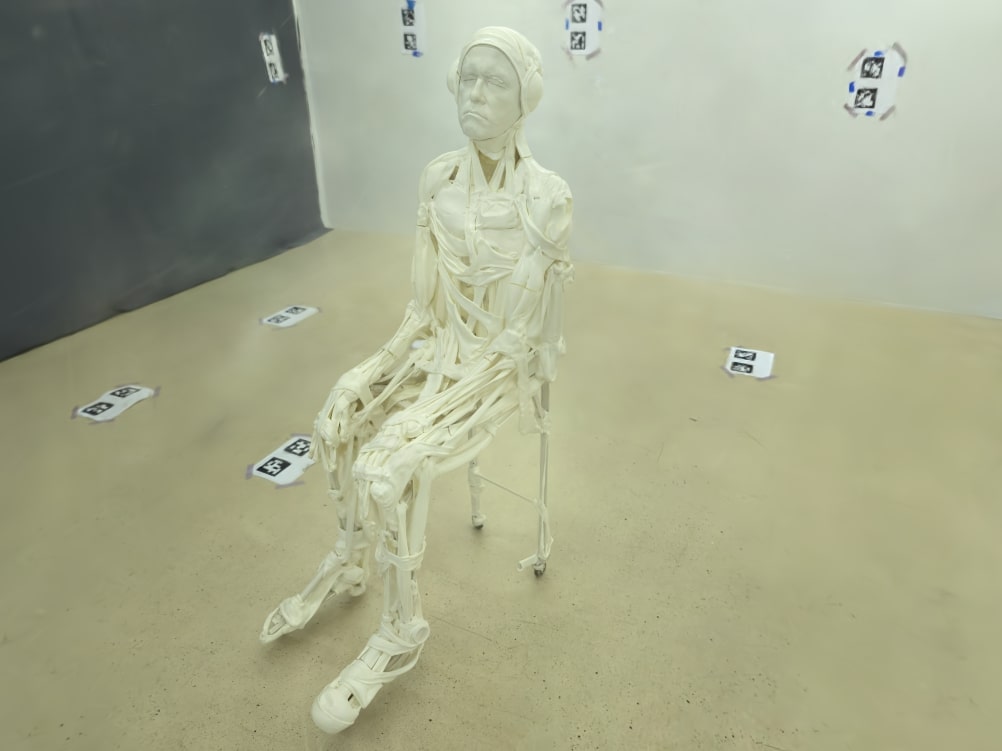} &
            \includegraphics[width=0.32\linewidth]{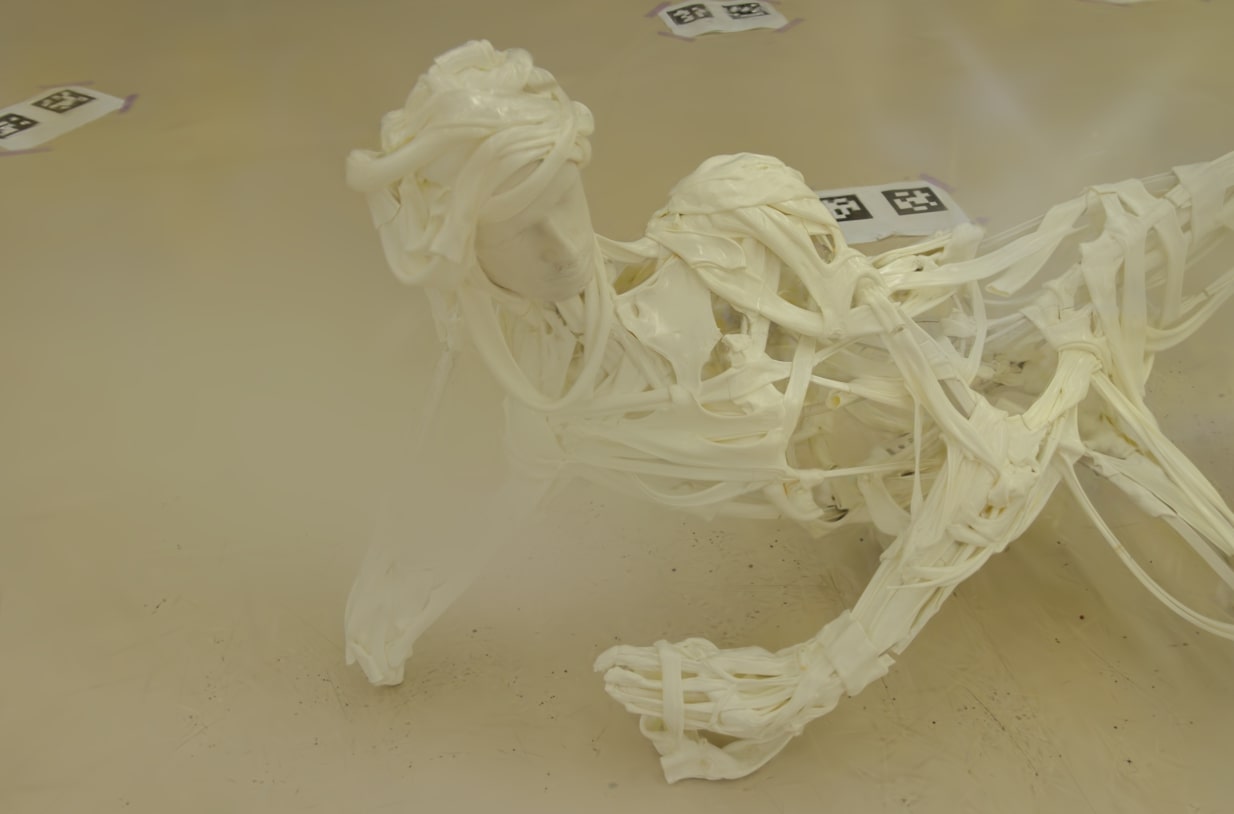} &
            \includegraphics[width=0.32\linewidth]{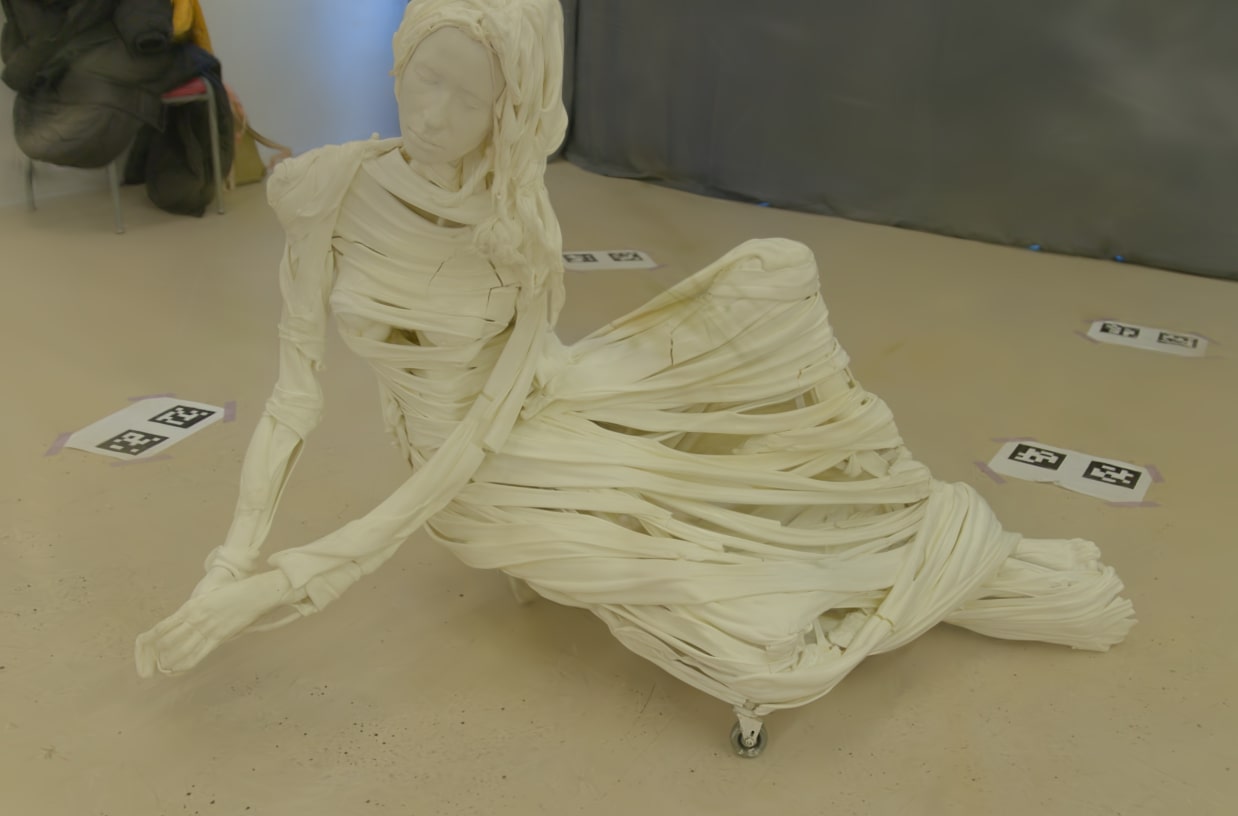} \\

            &
            \rotatebox{90}{\ \ \small Nerfacto} &
            \includegraphics[width=0.28\linewidth]{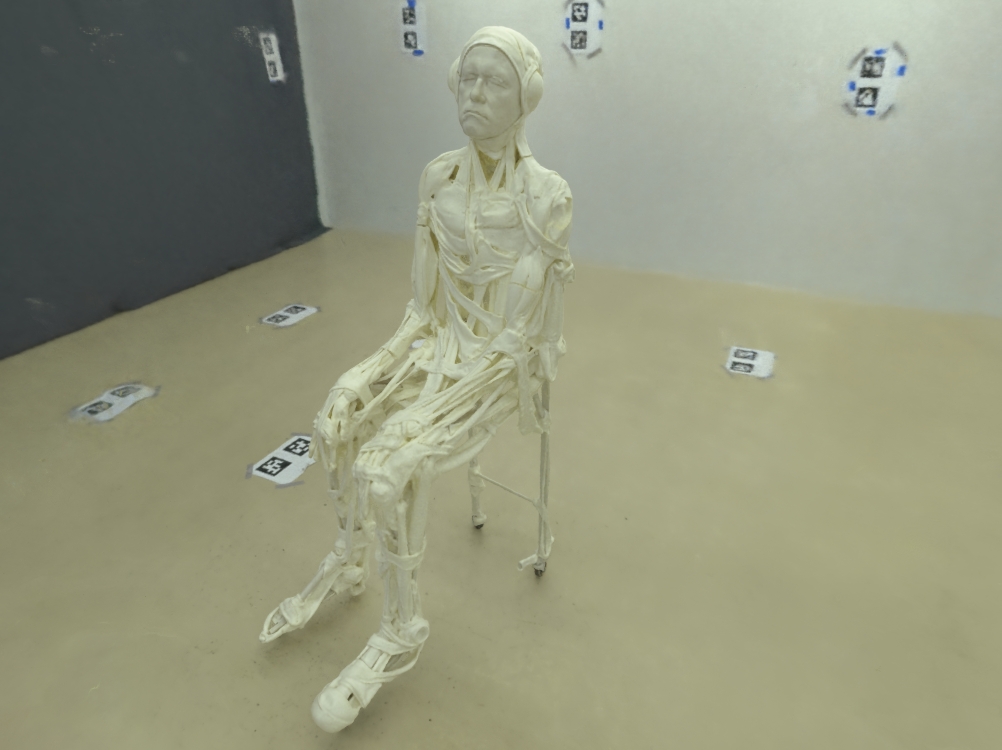} &
            \includegraphics[width=0.32\linewidth]{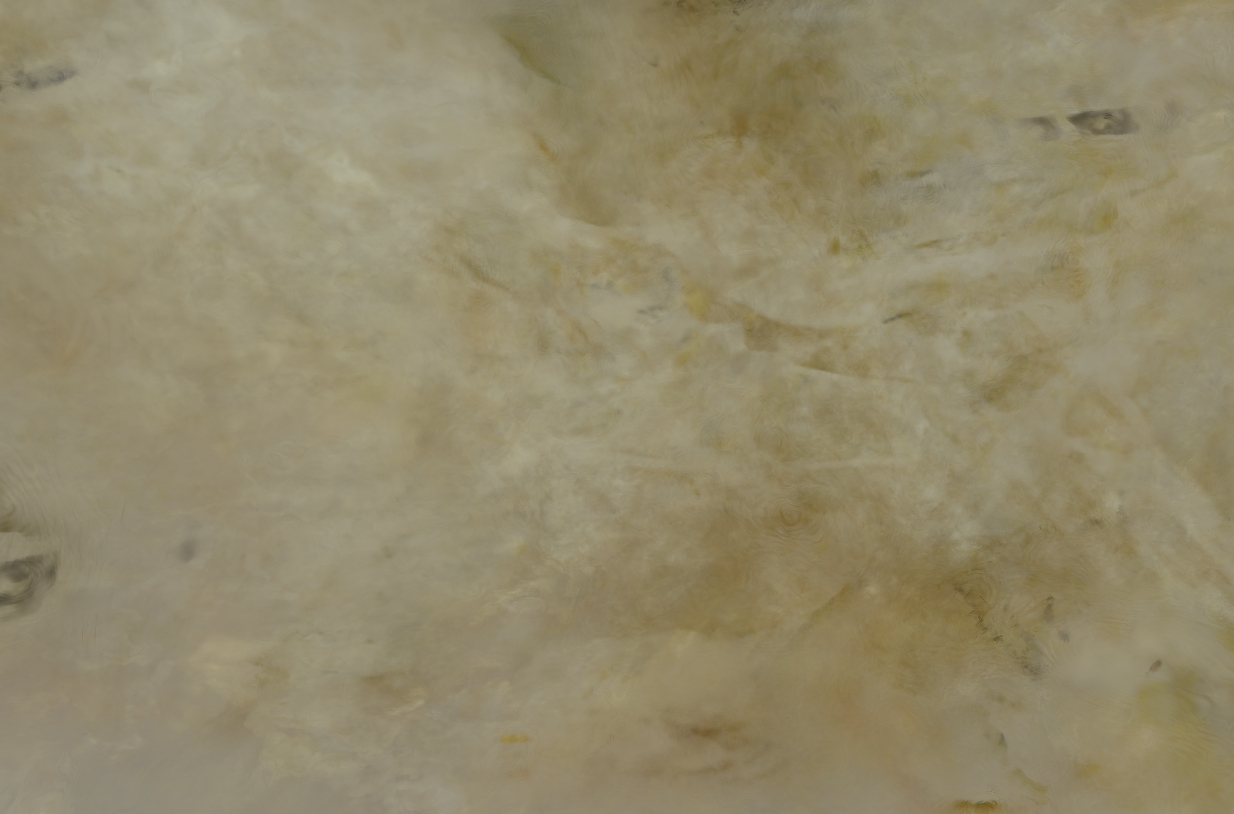} &
            \includegraphics[width=0.32\linewidth]{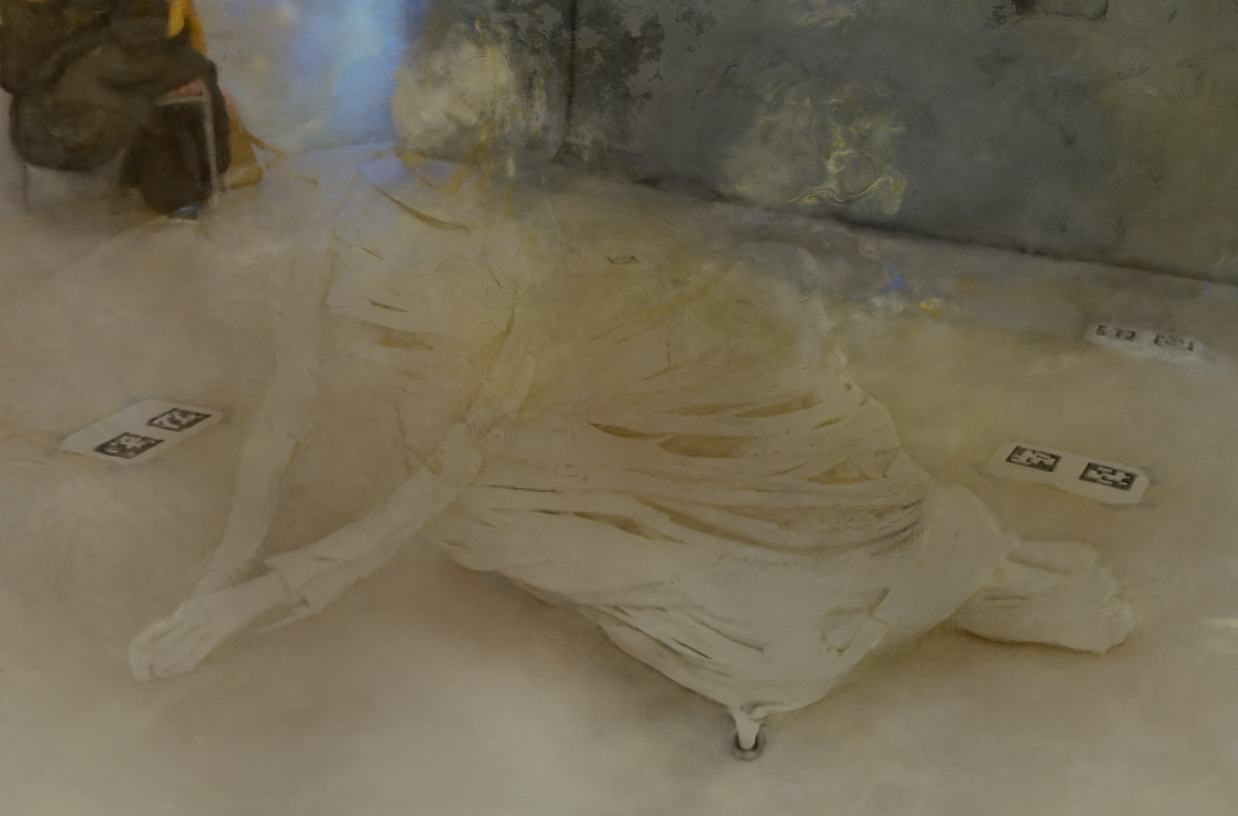} \\
            
            &
            \rotatebox{90}{\ \small Splatfacto} &
            \includegraphics[width=0.28\linewidth]{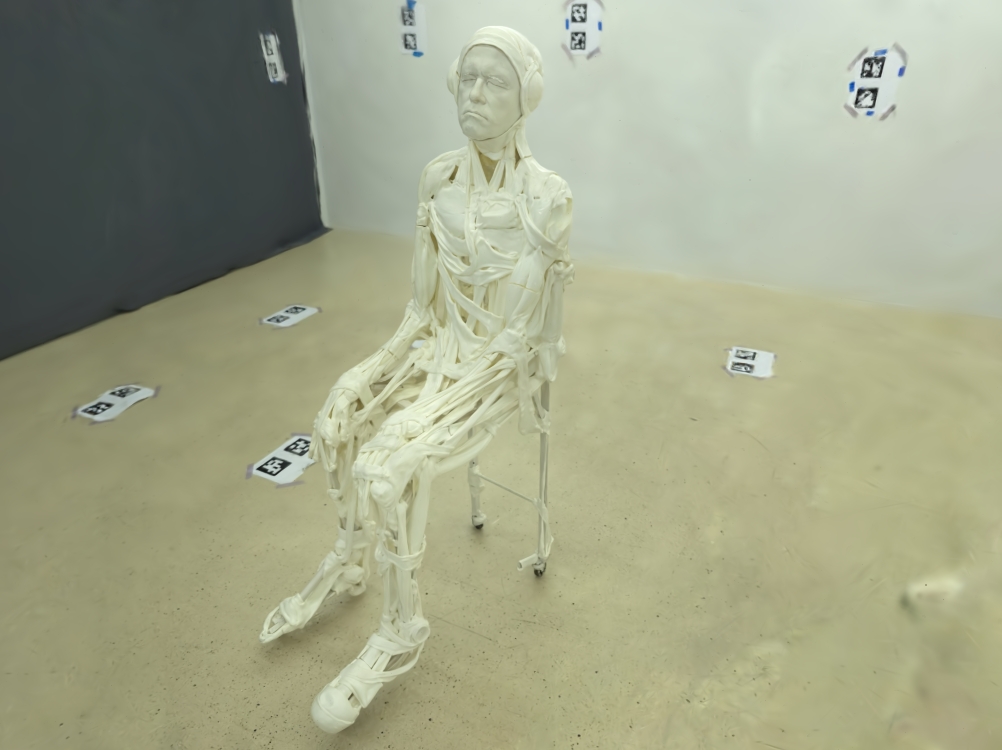} &
            \includegraphics[width=0.32\linewidth]{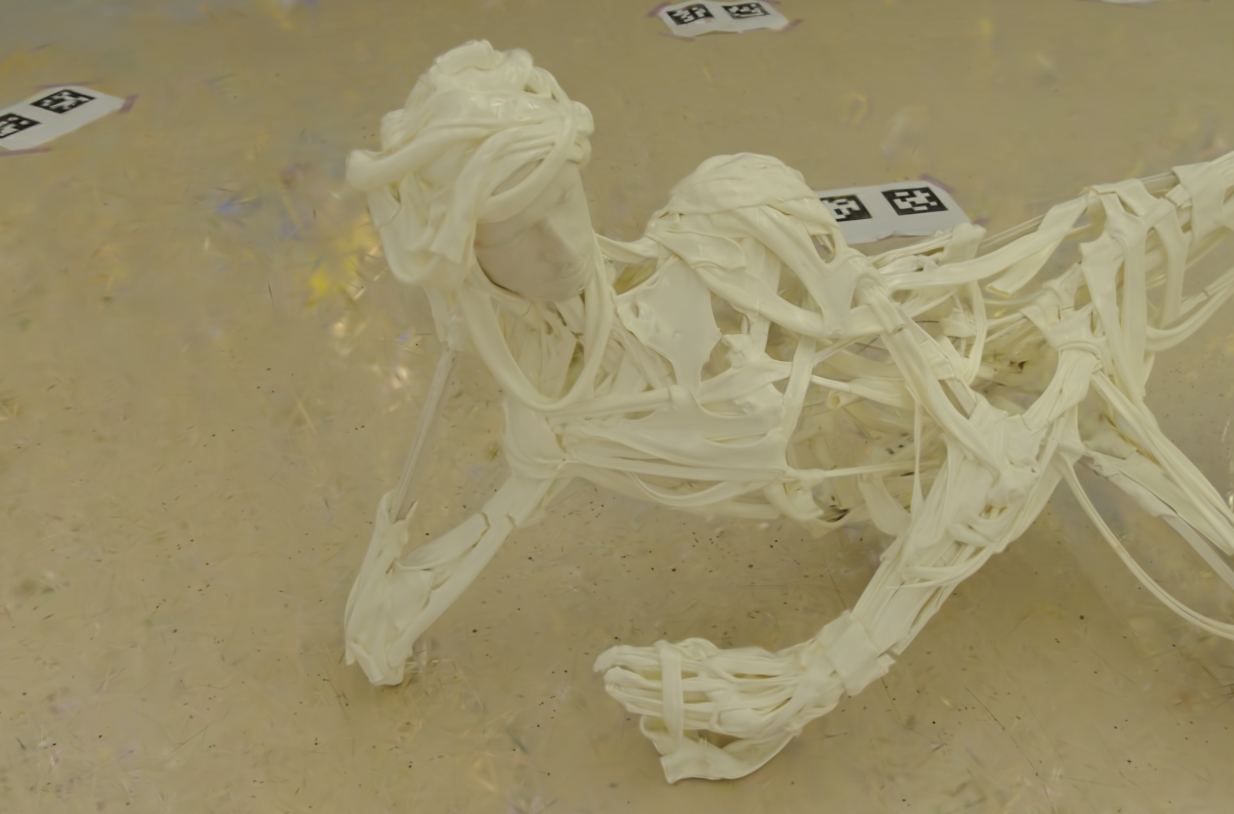} &
            \includegraphics[width=0.32\linewidth]{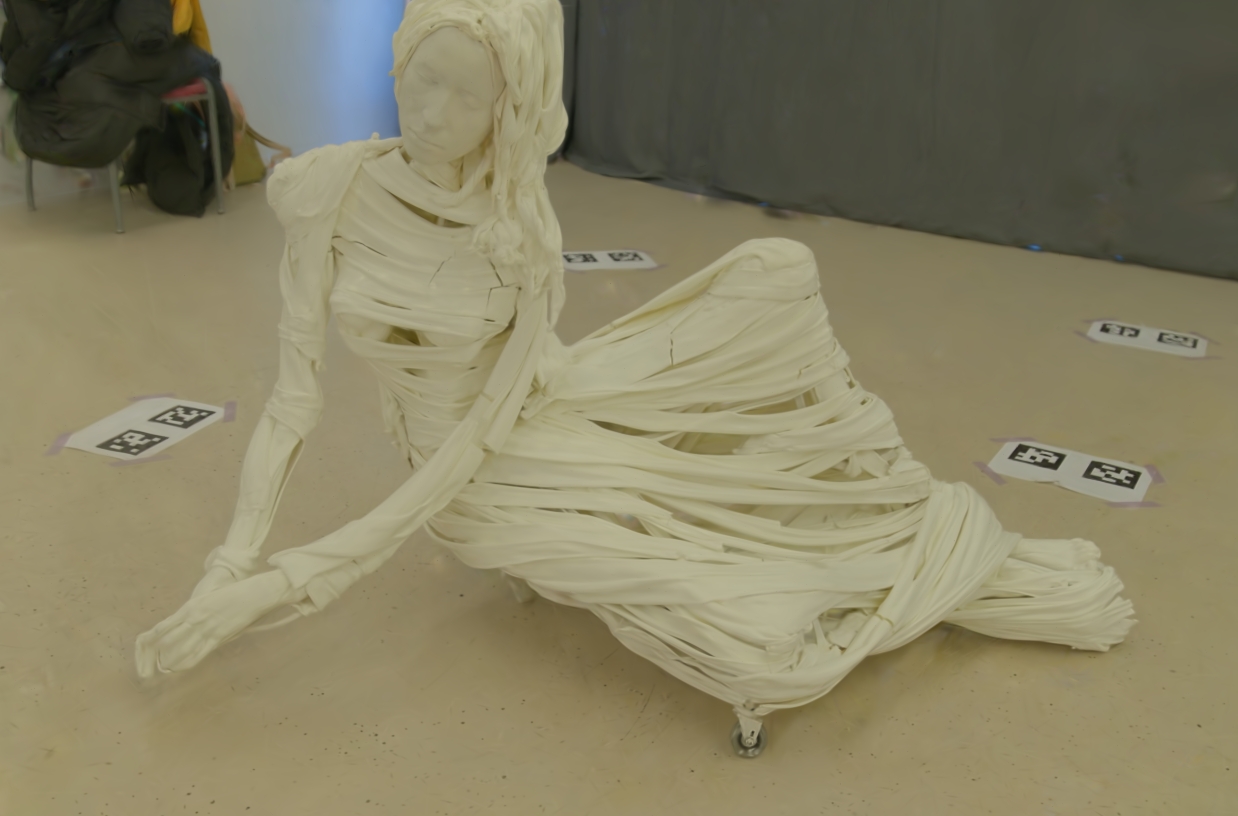} \\

            
            \rotatebox{90}{\ \small Splatfacto}
            &
            \rotatebox{90}{\ \ \small MCMC} &
            
            \includegraphics[width=0.28\linewidth]{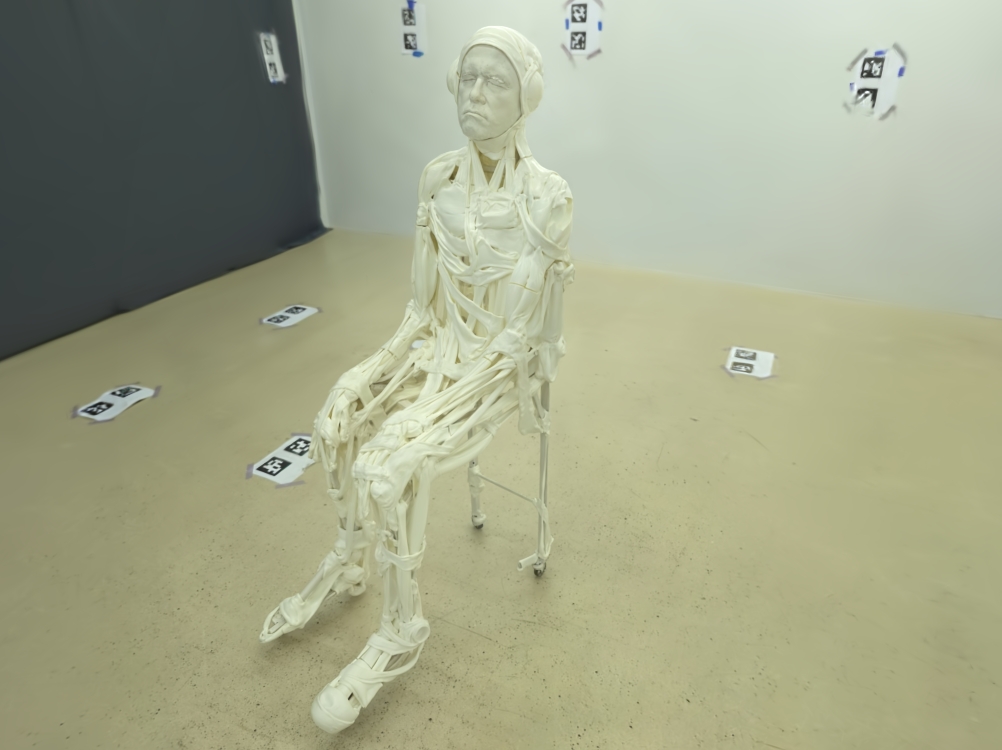} &
            \includegraphics[width=0.32\linewidth]{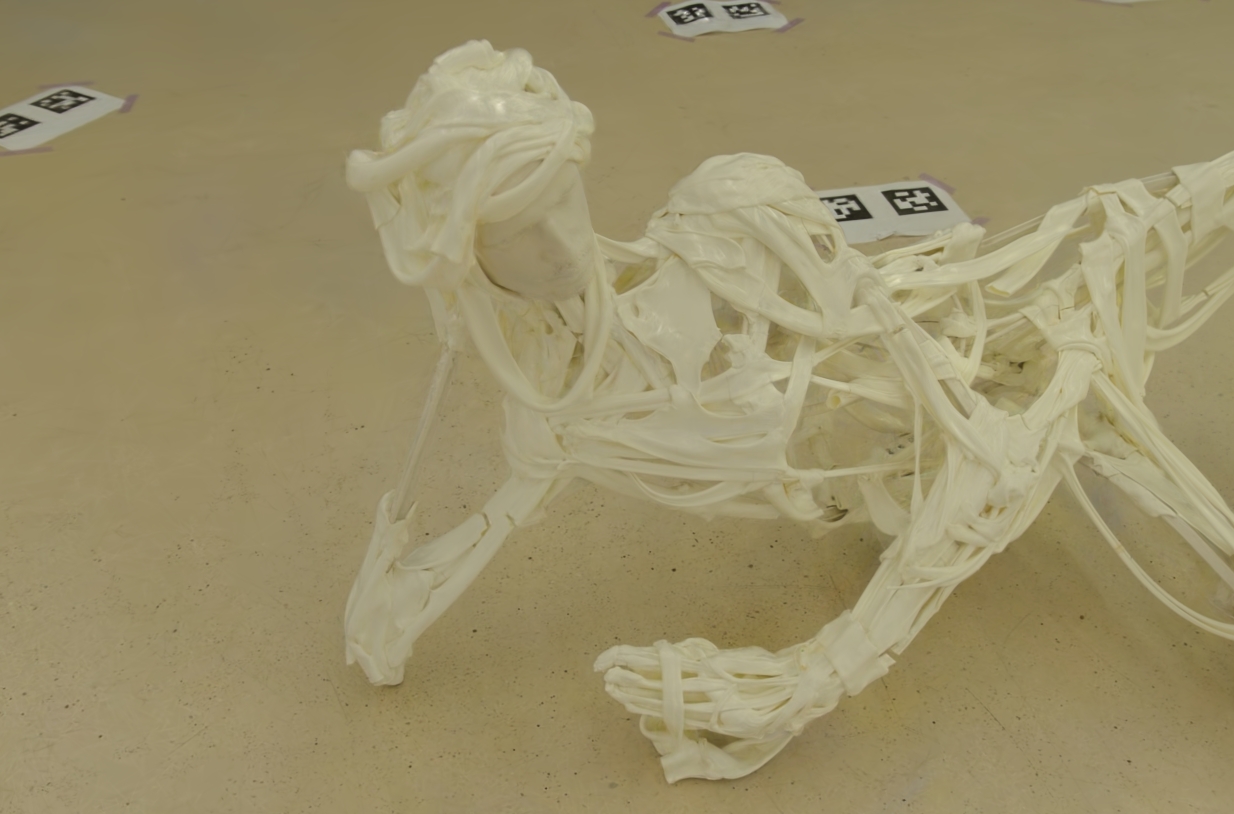} &
            \includegraphics[width=0.32\linewidth]{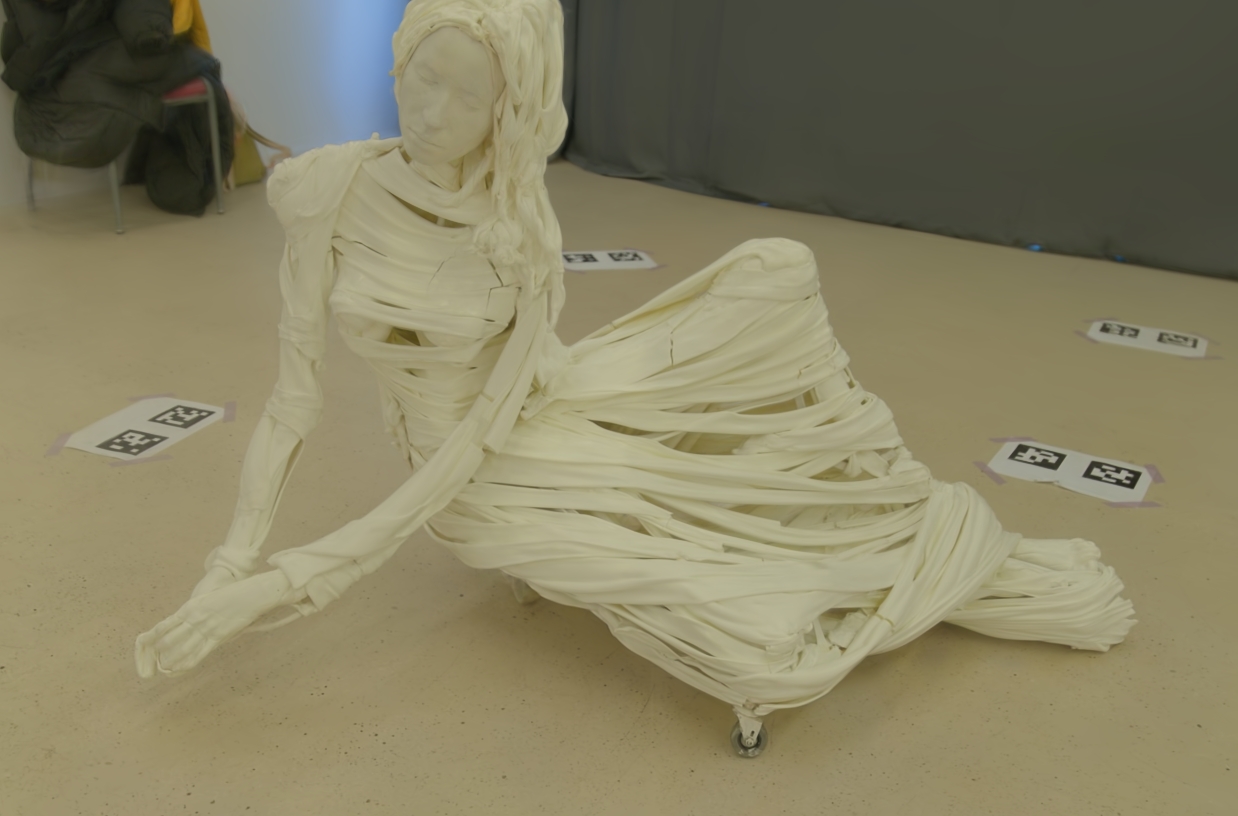} \\

            \rotatebox{90}{\quad \small Instant}
            &
            \rotatebox{90}{\ \quad \small NGP} &
            \includegraphics[width=0.28\linewidth]{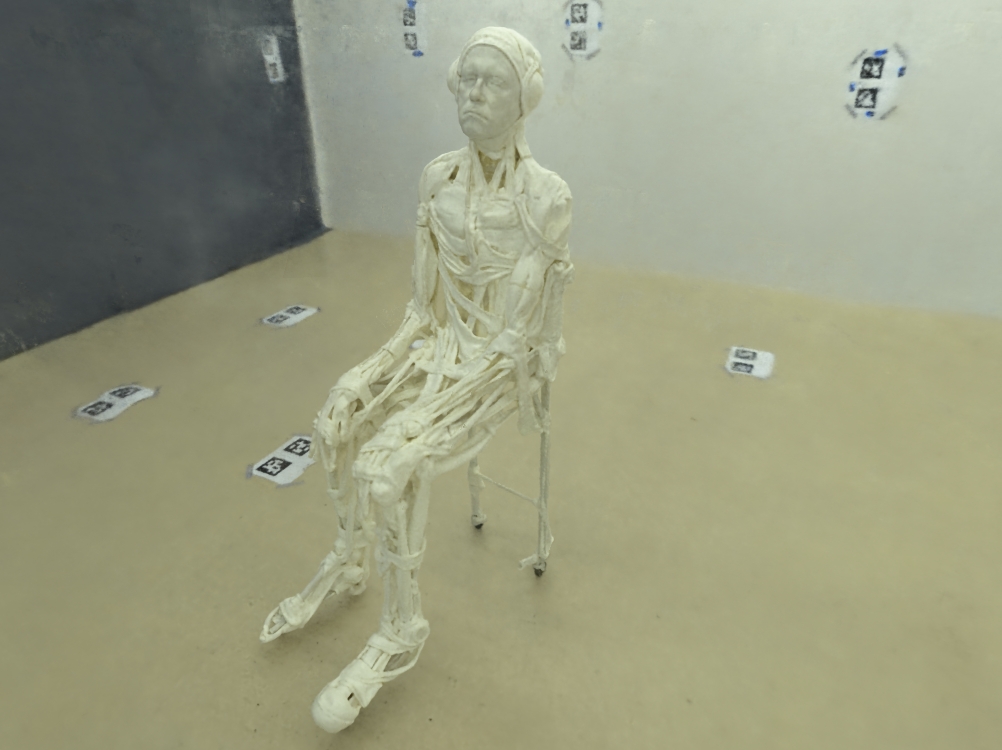} &
            \includegraphics[width=0.32\linewidth]{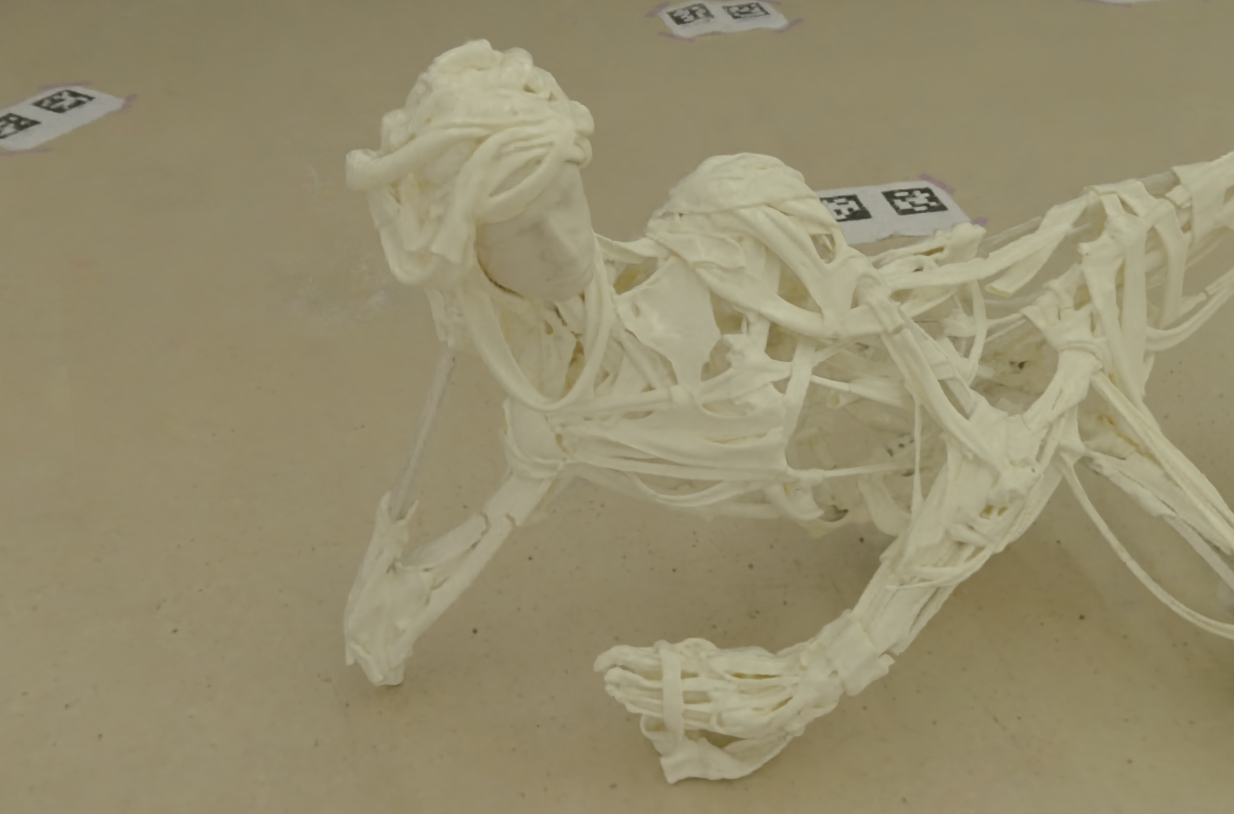} &
            \includegraphics[width=0.32\linewidth]{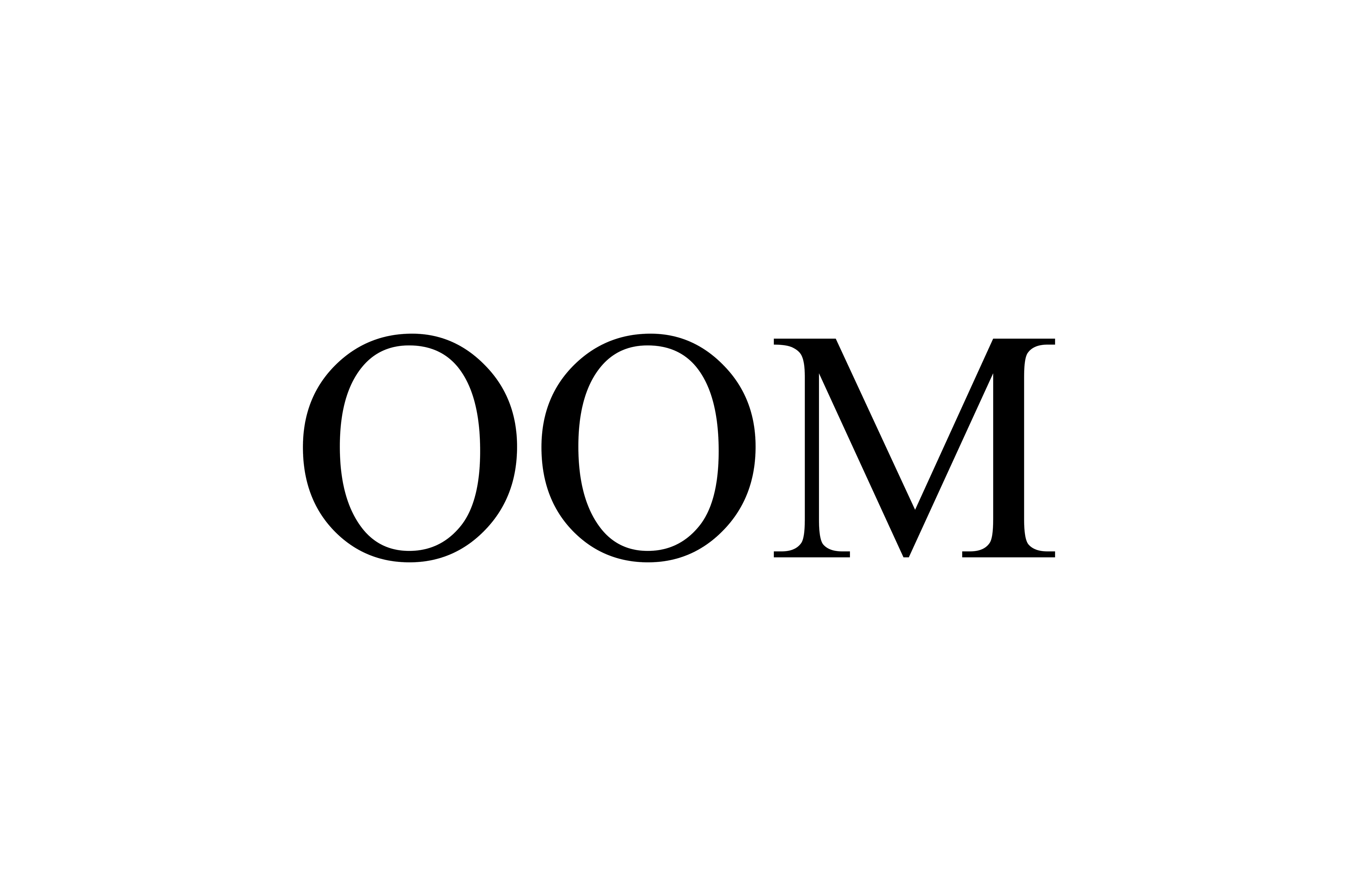}\\
        \end{tabular}
    \end{minipage}
    \caption{Qualitative results on Jaroslaw, Kacper and Paulina scenes. The majority of methods struggled on scenes with sparse number of training views (Kacper and Paulina). We have chosen the same frames for a fair comparison, however, Mip-Splatting also showed artifacts (see Figure \ref{fig:sparse_views}).}
    \label{fig:results_2}
    \vspace{-0.5cm}
\end{wrapfigure}

\paragraph{Quantitative results} The quantitative performance of the selected 3D reconstruction methods on the \our{} dataset is presented in Table~\ref{tab:results}. This table summarizes key image-based fidelity metrics (SSIM, PSNR, LPIPS), training duration, and rendering speed (FPS) for each of the six scenes and their averages.

The general superiority of Gaussian Splating (GS) based methods (2DGS, Mip-Splatting, Splatfacto, Splatfacto-MCMC) in reconstruction quality and rendering speed suggests that their explicit, point-based representations are currently more robust to the combination of challenges presented. Splatfacto-MCMC achieved the highest scored in PSNR, LPIPS and SSIM metrics indicating its proficiency in navigating the low contrast objects with low textures and low training samples. Nonetheless, it had the longest training time, longer even than Nerfacto.  

Conversely, the difficulties encountered by NeRF-based approaches (Nerfacto, Instant-NGP) are particularly revealing about the dataset's nature. Nerfacto's considerably lower fidelity scores across all metrics suggest that \our{} effectively exposes sensitivities common in implicit volumetric methods. Instant-NGP sadly ended with OOM for two out of six scenes on our hardware. Additionally, even among trained successfully trained scenes it suffered OOM when saving final renders. These sensivities might arise from the dynamic background elements present in some captures, low training sample or the discrepancies in white balance across different input views, both of which can lead to inconsistencies or artifacts in the learned neural field. The uniform white color and low texture of the sculptures likely exacerbate these issues, making it harder for the network to converge to a globally consistent and detailed representation. 

The dataset's emphasis on low contrast objects of interest (white sculptures) against potentially complex or dynamic backgrounds, coupled with variations in data quantity and quality (e.g., few matched photos, white balance shifts), creates a benchmark where the nuances between algorithmic approaches become clearly visible in the quality of the final reconstruction.

In summary, the performance spectrum observed on \our{} validates its design philosophy. It successfully creates a benchmark where the limitations of existing methods - particularly their sensitivity to fine geometric details in low-texture environments, robustness to color inconsistencies, and scalability with data volume, are brought to the forefront, thereby serving as a valuable resource for driving future advancements in the field. 
\paragraph{Comparative Analysis of Dataset Characteristics}

To contextualize the specific contributions of \our{}, Table \ref{tab:dataset_comparison} provides a comparative overview against several representative datasets commonly used in 3D reconstruction community. The table evaluates these datasets along key dimensions of real-world challenges that often hinder reconstruction quality, particularly those stemming from casual data acquisition using consumer-grade devices like smartphones. These dimension include:

\begin{itemize}
    \setlength\itemsep{-.1em}
    \item Real-world versus synthetic data, distinguishing between captures from physical environments versus computer-generated scenes, with the former introducing natural noise and lighting complexities.
    \setlength\itemsep{-.1em}
    \item A focus on consumer device capture artifacts, highlighting challenges arising from smartphone cameras such as automatic white balance adjustments, unfocused images, or aggressive post-processing, as opposed to more controlled professional captures.
    \setlength\itemsep{-.1em}
    \item Photometric inconsistencies, such as varying color casts (e.g., white balance shifts) or brightness across images of the same static point, which violate the common assumption of consistent appearance.
    \setlength\itemsep{-.1em}
    \item Sparse views resulting from Structure-from-Motion (SfM) processing failures, where only a limited subset of input images can be successfully registered, reducing the available data for reconstruction.
    \setlength\itemsep{-.1em}
    \item The presence of dynamic background elements, meaning moving people behind the primary subject that can cause artifacts if not explicitly handled.
    \setlength\itemsep{-.1em}
    \item And challenging object properties such as low variation in color and texture, where surfaces lack distinct patterns making feature matching for SfM and dense reconstruction difficult. Additionally, it has intricate geometry, referring to complex shapes with fine details, perforations, or thin structures that are hard to model accurately.
    
\end{itemize}
\begin{figure}[t]
    \centering
    \renewcommand{\arraystretch}{0.2}
    \setlength{\tabcolsep}{0.4pt}
    \begin{tabular}{c@{}c@{}c@{}c@{}c@{}c@{}c@{}}
        GT &
        2DGS & 
        Mip &
        Nerfacto &
        Splatfacto &
        Splatfacto &
        Instant \\
        &
        & 
        Splatting &
        &
        &
        MCMC &
        NGP \\ [0.2cm]
        \includegraphics[width=0.13\linewidth]{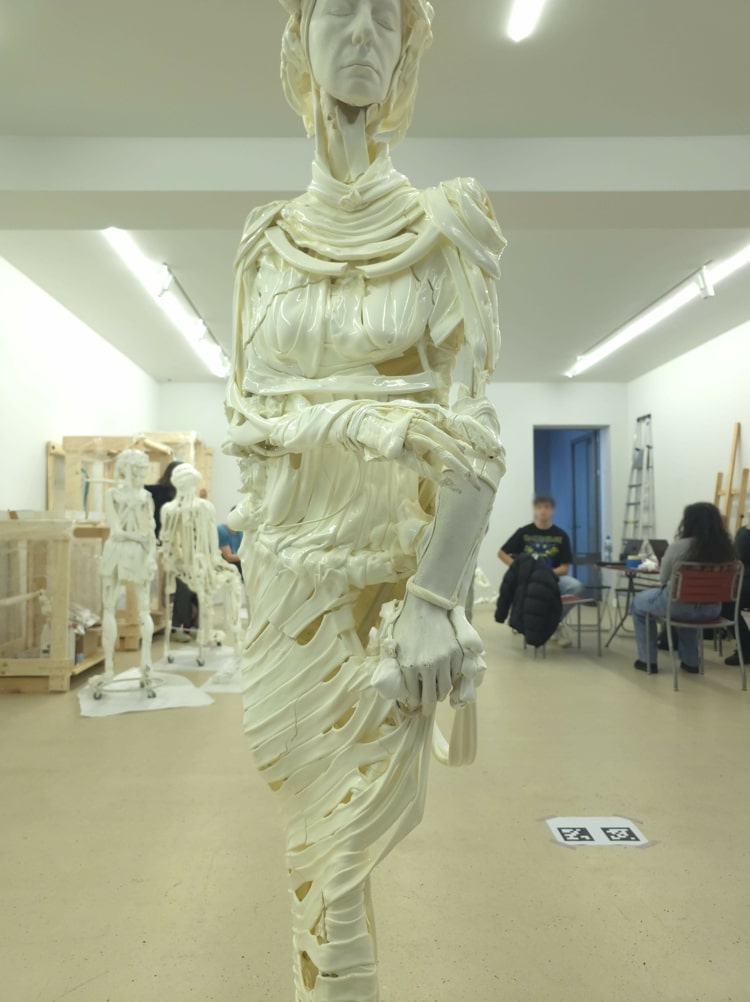} &
        \includegraphics[width=0.13\linewidth]{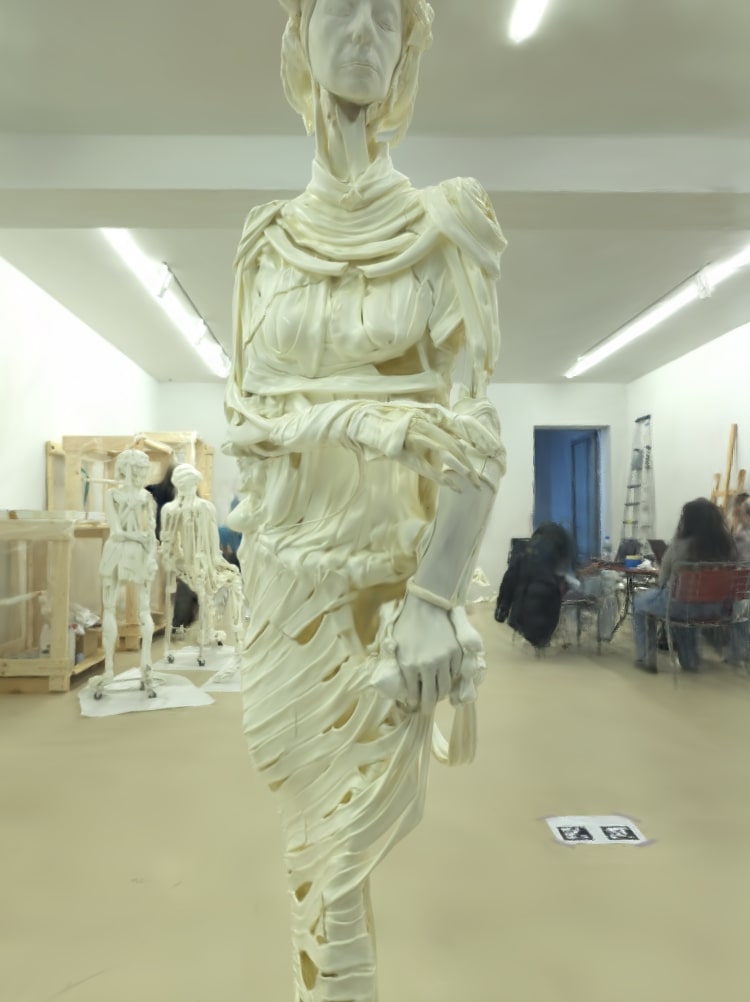} & 
        \includegraphics[width=0.13\linewidth]{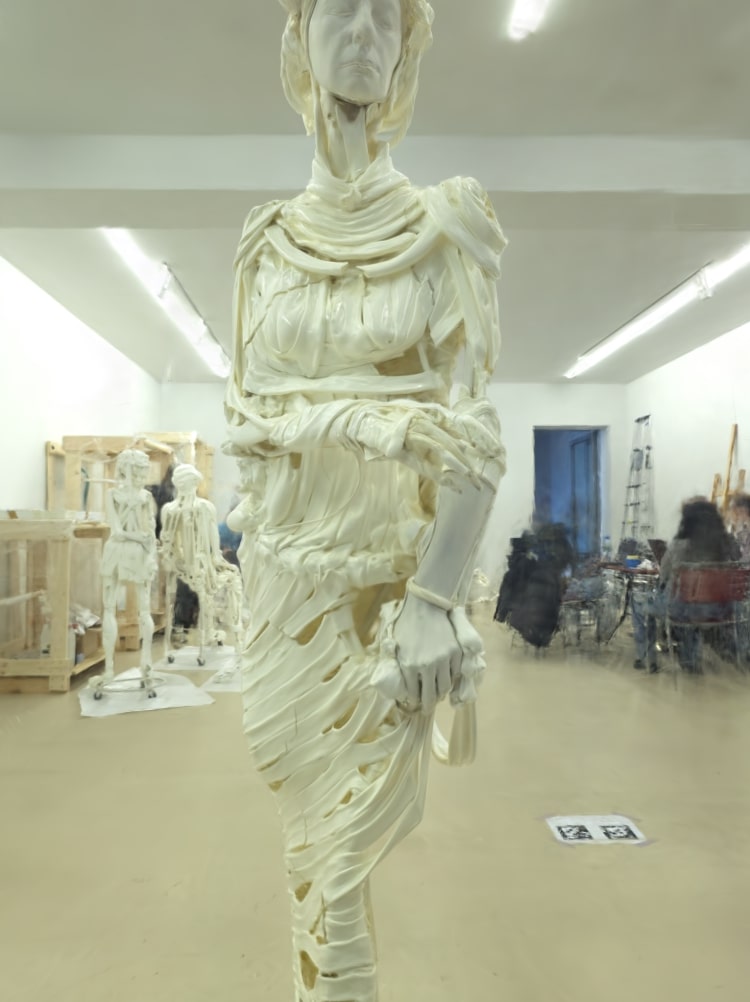} & 
        \includegraphics[width=0.13\linewidth]
        {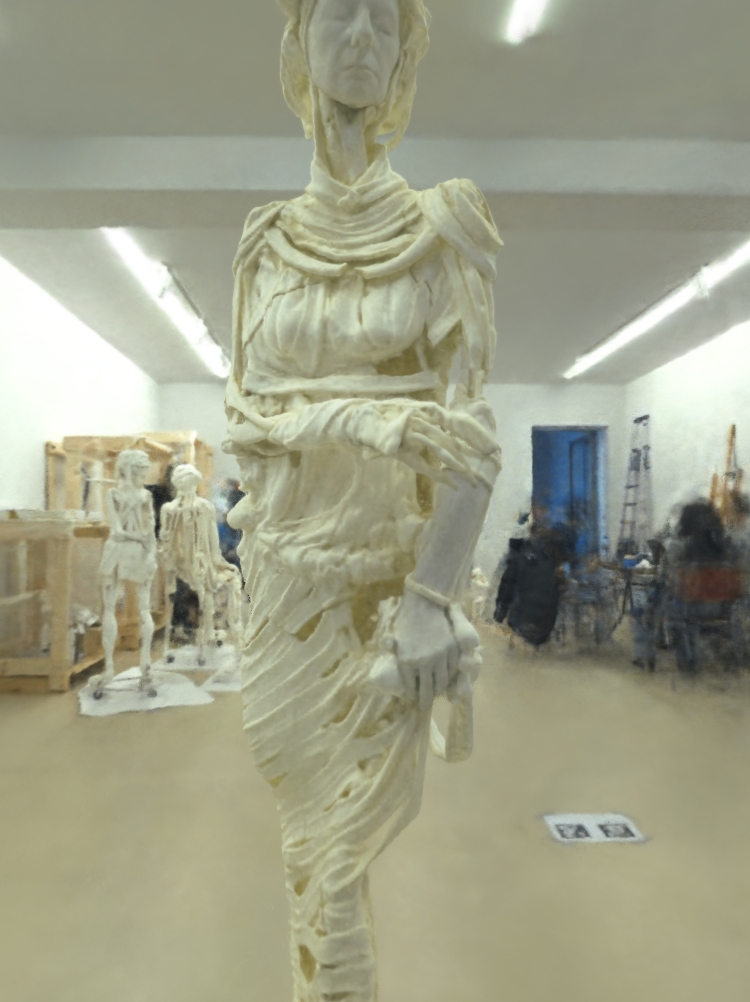} & 
        \includegraphics[width=0.13\linewidth]{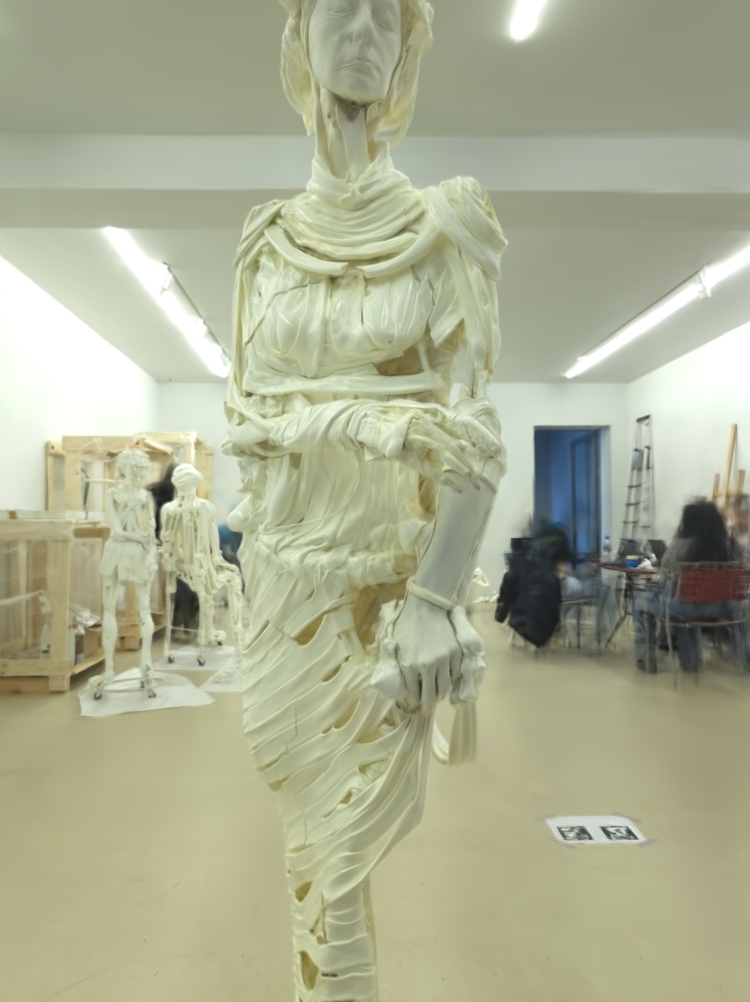} &
        \includegraphics[width=0.13\linewidth]{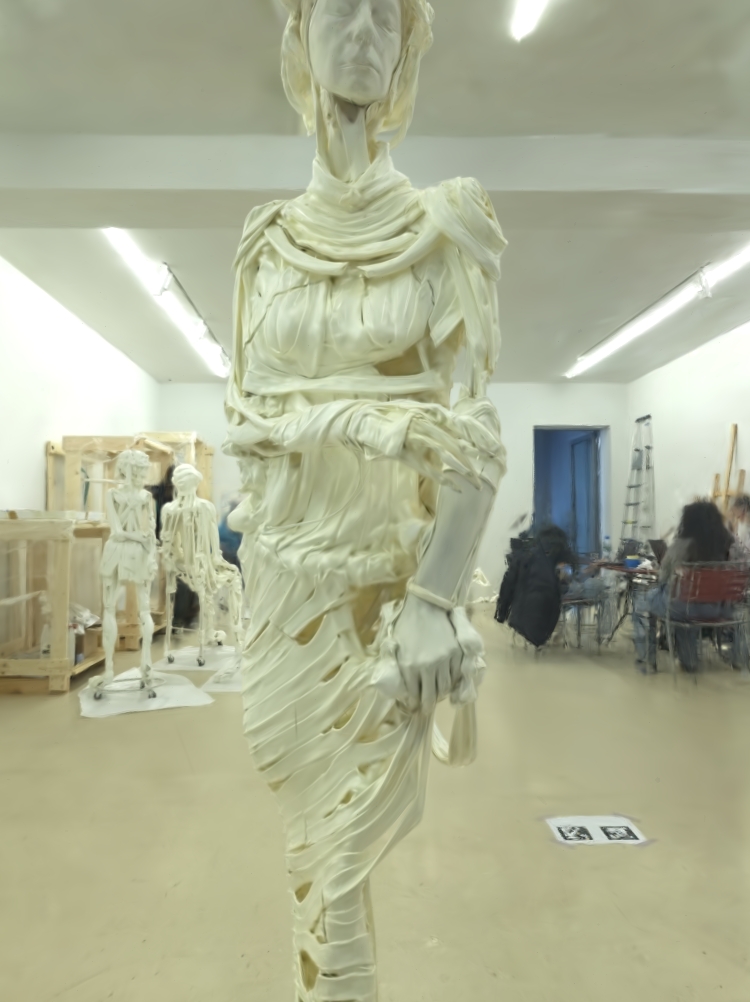} &
        \includegraphics[width=0.13\linewidth]{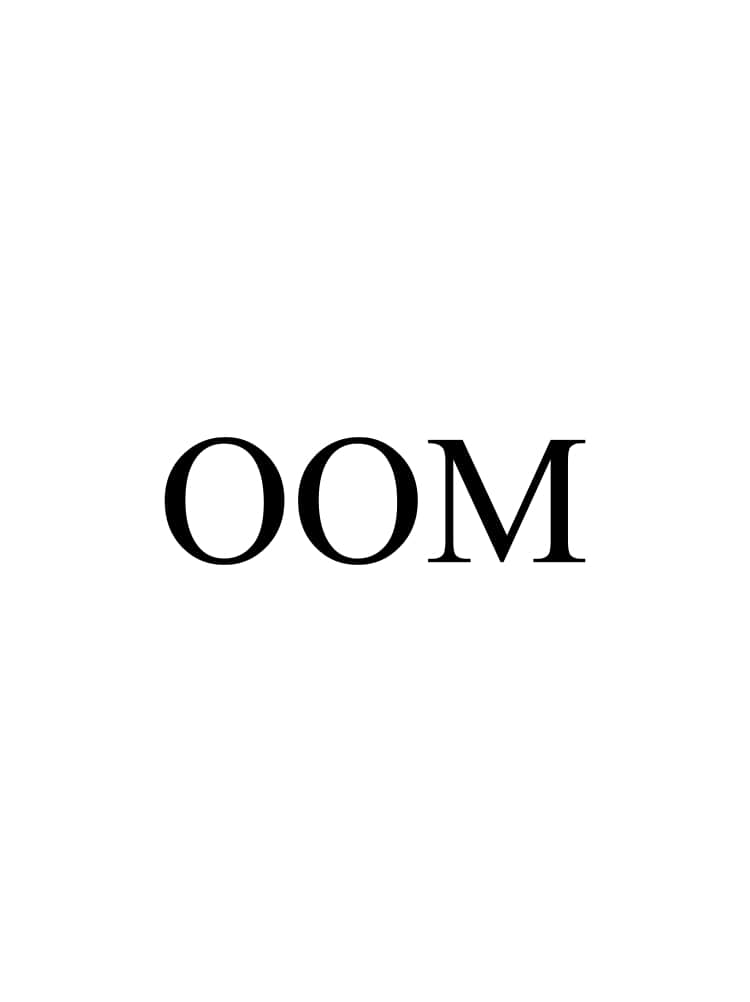} \\
        
        \includegraphics[width=0.13\linewidth]{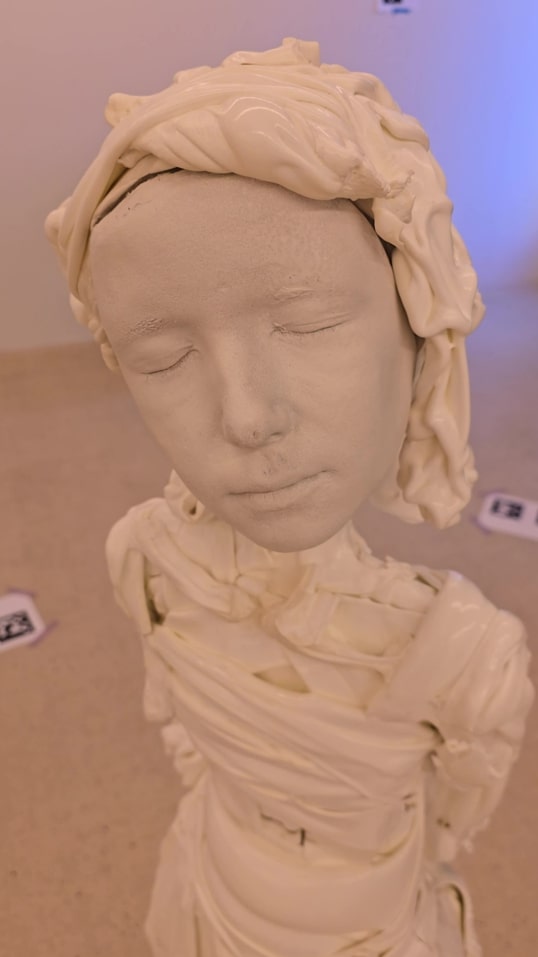} &
        \includegraphics[width=0.13\linewidth]{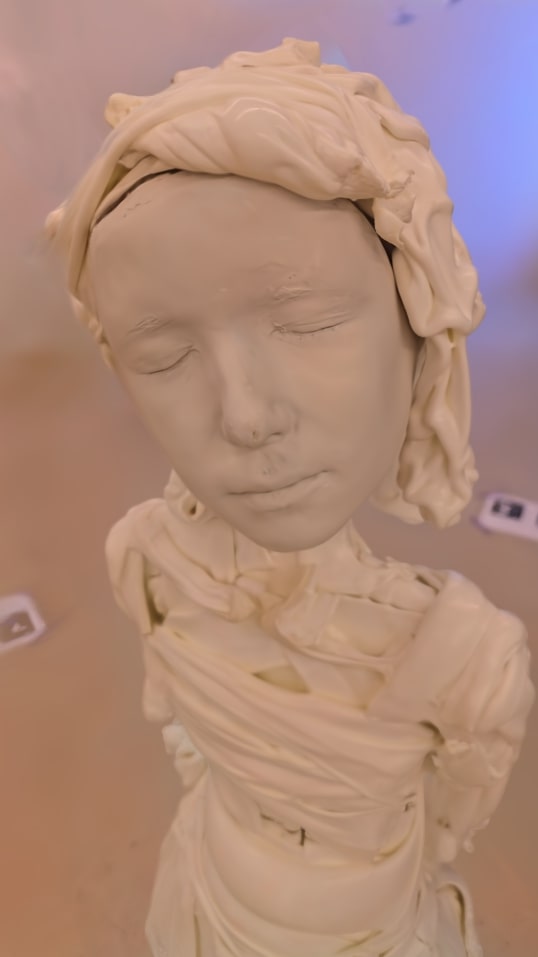} & 
        \includegraphics[width=0.13\linewidth]{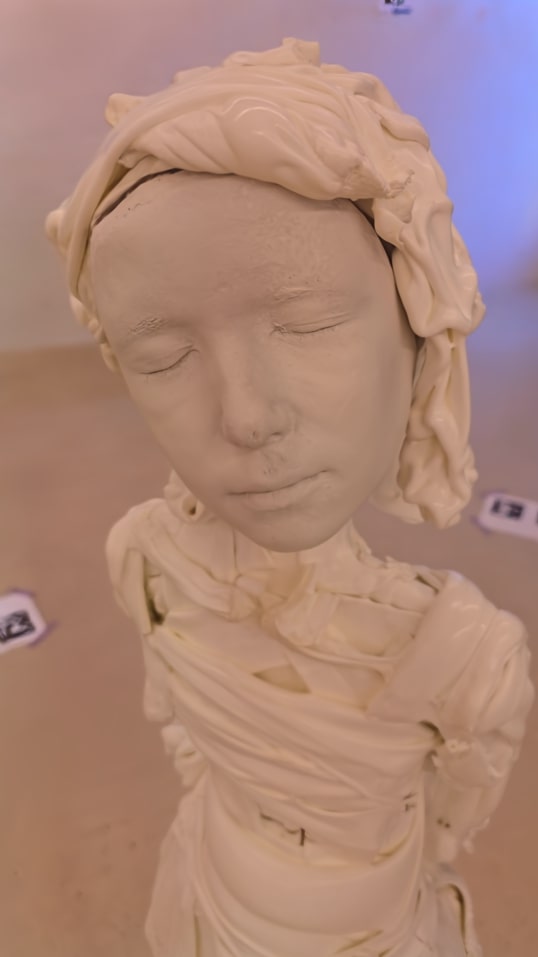} & 
        \includegraphics[width=0.13\linewidth]
        {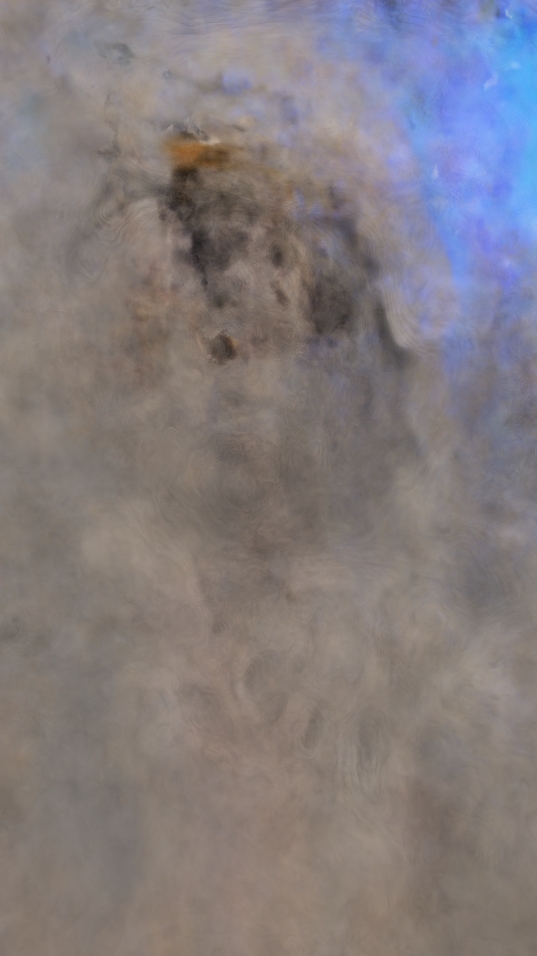} & 
        \includegraphics[width=0.13\linewidth]{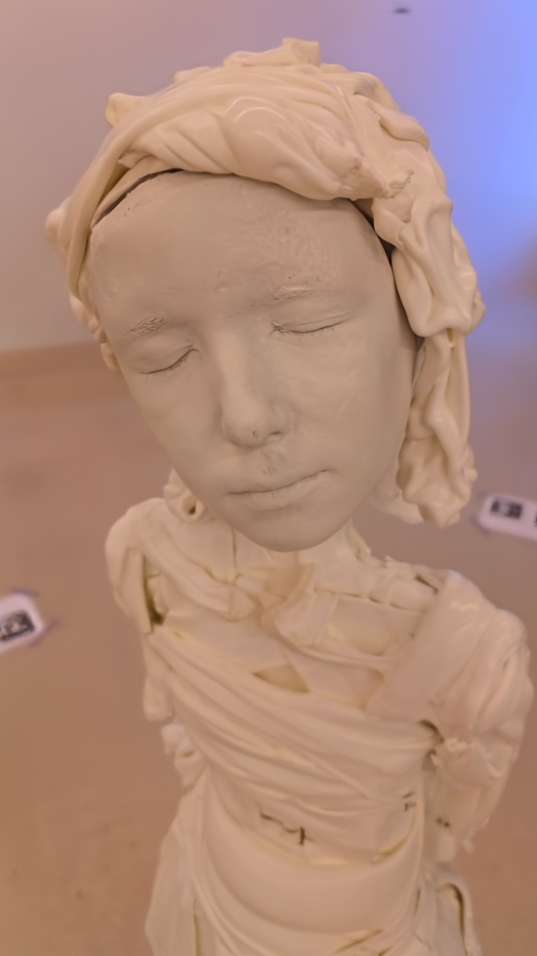} &
        \includegraphics[width=0.13\linewidth]{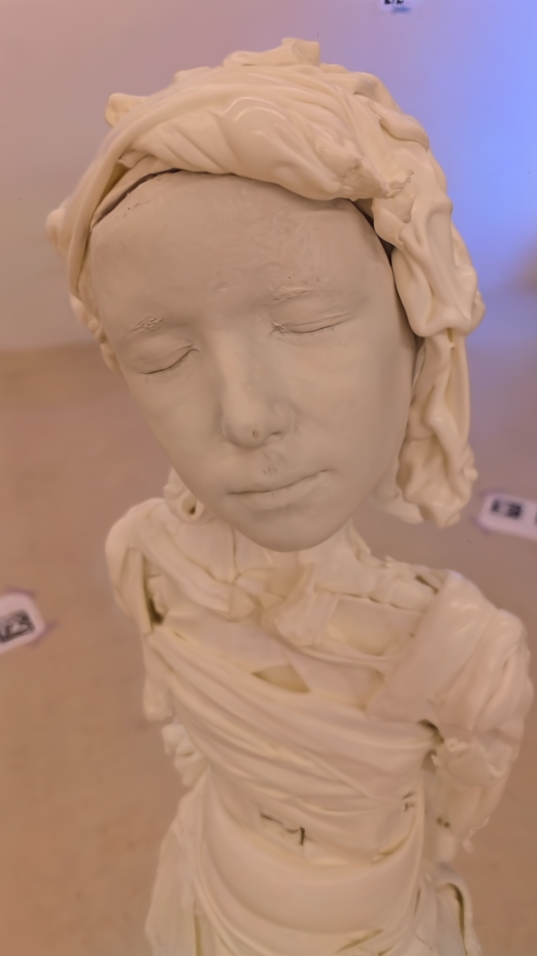} &
        \includegraphics[width=0.13\linewidth]{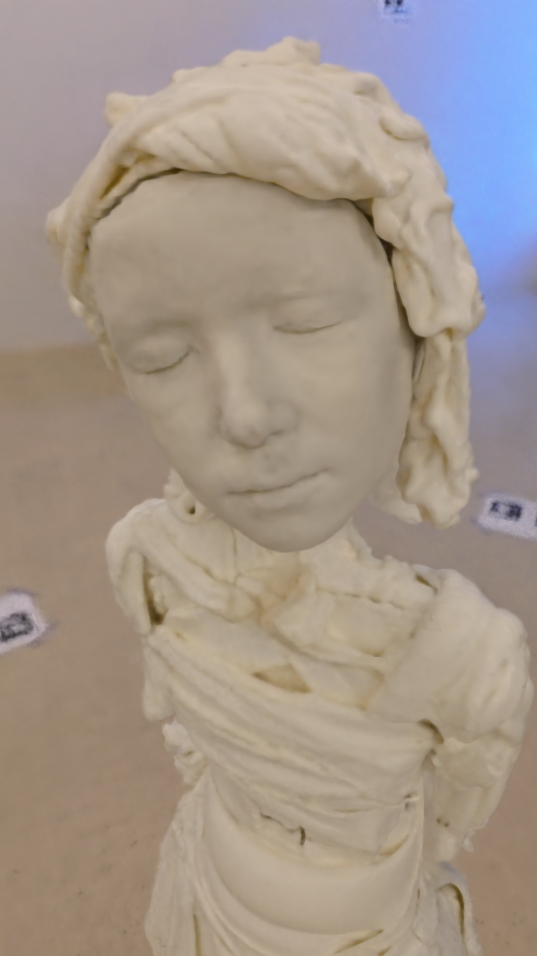} \\
        
        \includegraphics[width=0.13\linewidth]{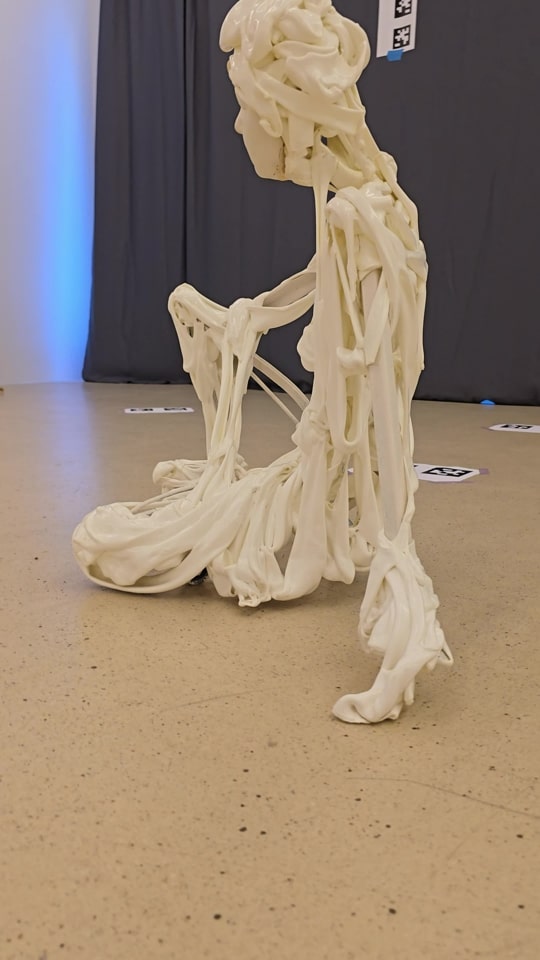} &
        \includegraphics[width=0.13\linewidth]{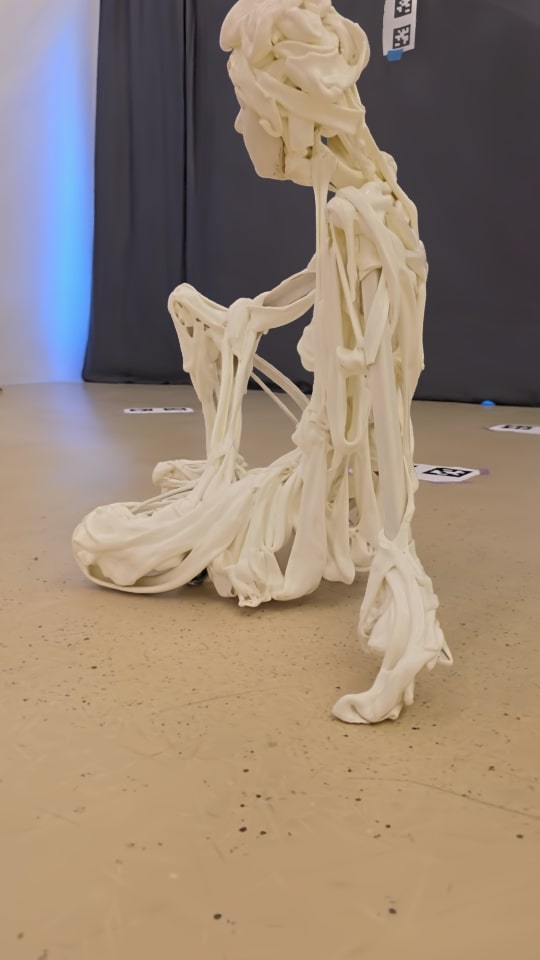} & 
        \includegraphics[width=0.13\linewidth]{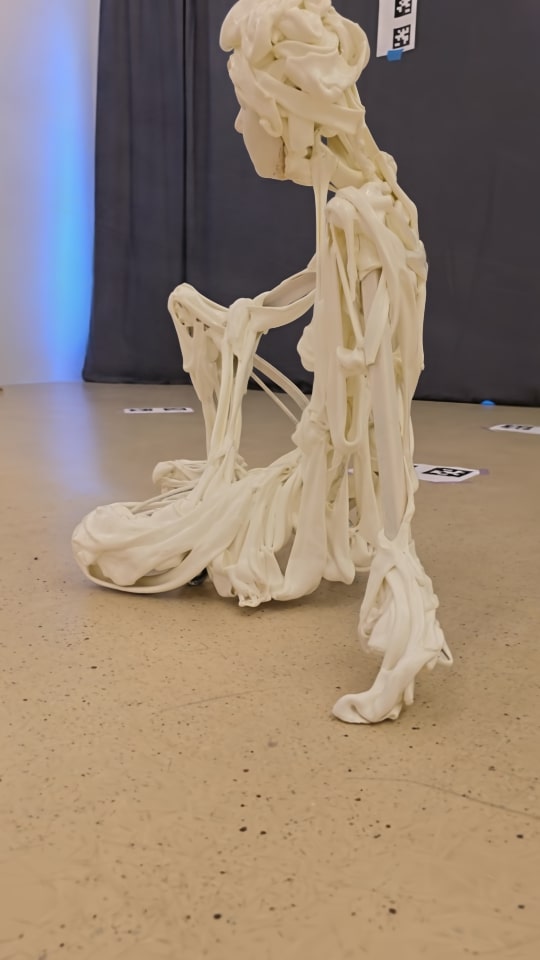} & 
        \includegraphics[width=0.13\linewidth]
        {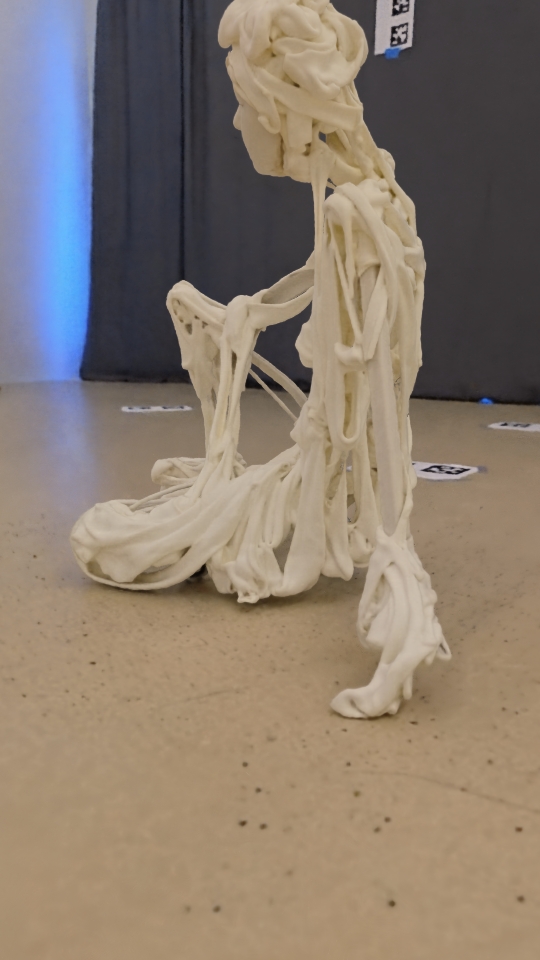} & 
        \includegraphics[width=0.13\linewidth]{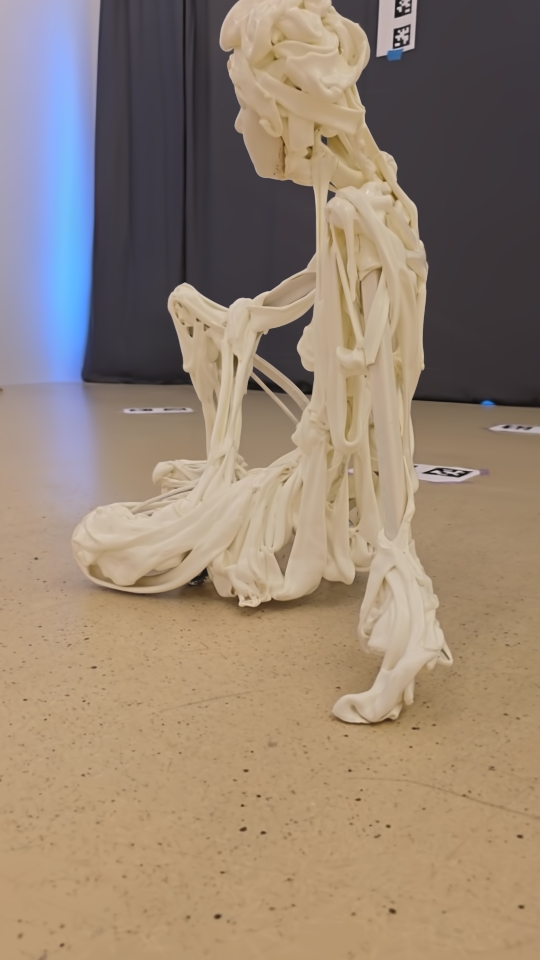} &
        \includegraphics[width=0.13\linewidth]{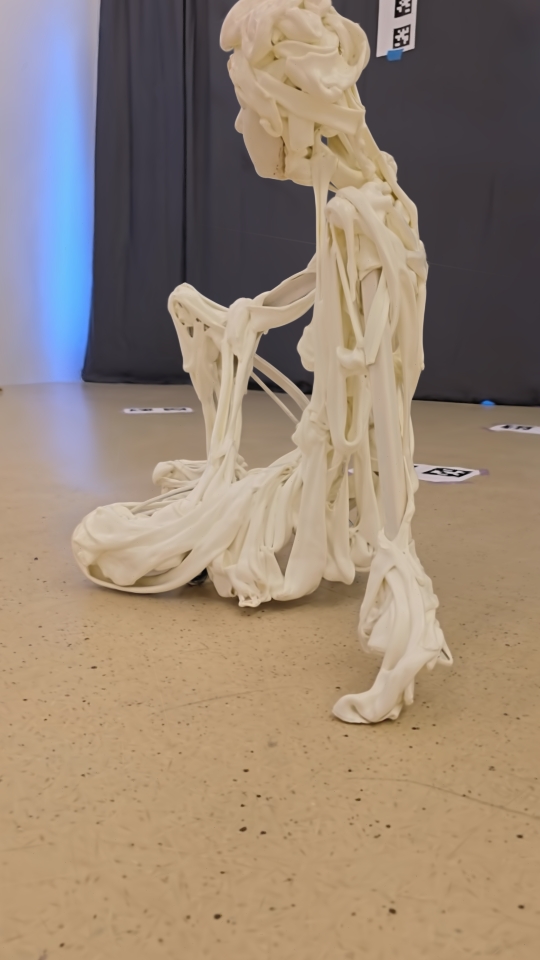} &
        \includegraphics[width=0.13\linewidth]{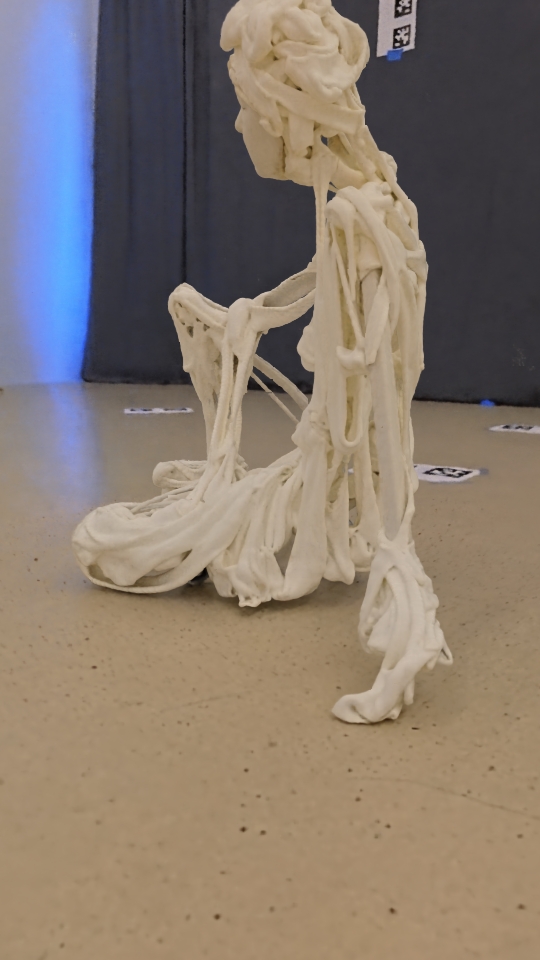} 
        
    \end{tabular}
    \caption{Qualitative results on Hanna, Tosia and Wiktor scenes. Splatfacto-MCMC achieved results closest to the original. However, on Tosia scene Instant-NGP even though its white balance differs from GT, it is closer to the wanted white balance that was seen in the majority of training data. Nerfacto created a significantly blurred photo, however, it also kept stable white balance (WB). It seems like NeRF-based methods are superior on capturing the stable essence of the object. 3DGS-based methods displayed higher ability to reconstruct sudden changes in WB.}
    \label{fig:results_1}
\end{figure}


\begin{table*}[ht]
\centering
\caption{Comparison of \our{} with representative existing datasets across key real-world challenge dimensions. \our{} uniquely combines challenges related to object properties (low texture variation, intricate geometry) with acquisition artifacts common in consumer-grade captures (photometric inconsistency, dynamic elements, variable data density including sparsity and high-resolution issues). (\cmark = Feature present/addressed, \xmark = Feature largely absent or not a primary focus)}
\label{tab:dataset_comparison}
\resizebox{\textwidth}{!}{
\begin{tabular}{lccccccc}
\toprule
\textbf{Dataset} & \textbf{Real-World} & \textbf{Consumer Device} & \textbf{Photometric} & \textbf{Sparse Views} & \textbf{Dynamic} & \textbf{Low Texture and} & \textbf{Intricate} \\
& \textbf{Data} & \textbf{Capture Focus} & \textbf{Inconsistency} & \textbf{(SfM Result)} & \textbf{Background} & \textbf{Color Variation} & \textbf{Geometry} \\
\midrule
NeRF-Synthetic & \xmark & \xmark & \xmark & \xmark & \xmark & Partial\textsuperscript{*} & Partial\textsuperscript{*} \\
LLFF & \cmark & \cmark & \xmark & \cmark & \xmark & \xmark & \xmark \\
Mip-NeRF 360 & \cmark & \xmark & \xmark & \xmark & Partial\textsuperscript{\textdaggerdbl} & \xmark & Partial \\
CO3D & \cmark & Partial\textsuperscript{\S} & \xmark & Partial\textsuperscript{\P} & Partial\textsuperscript{\P} & Partial\textsuperscript{\P} & Partial\textsuperscript{\P} \\
DeepBlending (common scenes) & \cmark & \xmark & \xmark & \xmark & \xmark & \xmark & \xmark \\
\midrule
\textbf{\our{}} & \textbf{\cmark} & \textbf{\cmark} & \textbf{\cmark} & \textbf{\cmark} & \textbf{\cmark} & \textbf{\cmark} & \textbf{\cmark} \\
\bottomrule
\multicolumn{8}{l}{\textsuperscript{*}\footnotesize Synthetic objects can be designed this way, but lack real-world sensor noise/lighting effects.} \\
\multicolumn{8}{l}{\textsuperscript{\textdaggerdbl}\footnotesize May contain some motion but typically not the primary challenge or uncurated passersby.} \\
\multicolumn{8}{l}{\textsuperscript{\S}\footnotesize Contains varied captures, some likely from phones, but curation might mitigate some artifacts; not the specific focus.} \\
\multicolumn{8}{l}{\textsuperscript{\P}\footnotesize Large scale dataset, some instances might exhibit these traits, but not a guaranteed/designed-in challenge for all/most scenes.} \\
\end{tabular}%
}
\end{table*}

As this comparison in Table \ref{tab:dataset_comparison} highlights, existing datasets, while foundational, typically address only a subset of these challenges. 

\paragraph{Qualitative results}
Qualitative results are presented in Figure \ref{fig:results_2} and Figure \ref{fig:results_1}. It represents renders made on test set with different methods. Even though, Splatfacto-MCMC achieved the best results both qualitatively and quantitatively, it still could misses intricate details as previously mentioned and shown in Figure \ref{fig:hanna_resize}. 

\begin{figure}[t]
    \centering
    \renewcommand{\arraystretch}{0.2}
    \setlength{\tabcolsep}{0.4pt}
    \begin{tabular}{c@{}c@{}c@{}}
        \\[0.2cm]
        2DGS & Mip-Splatting & Nerfacto \\
        
        \includegraphics[width=0.33\linewidth]{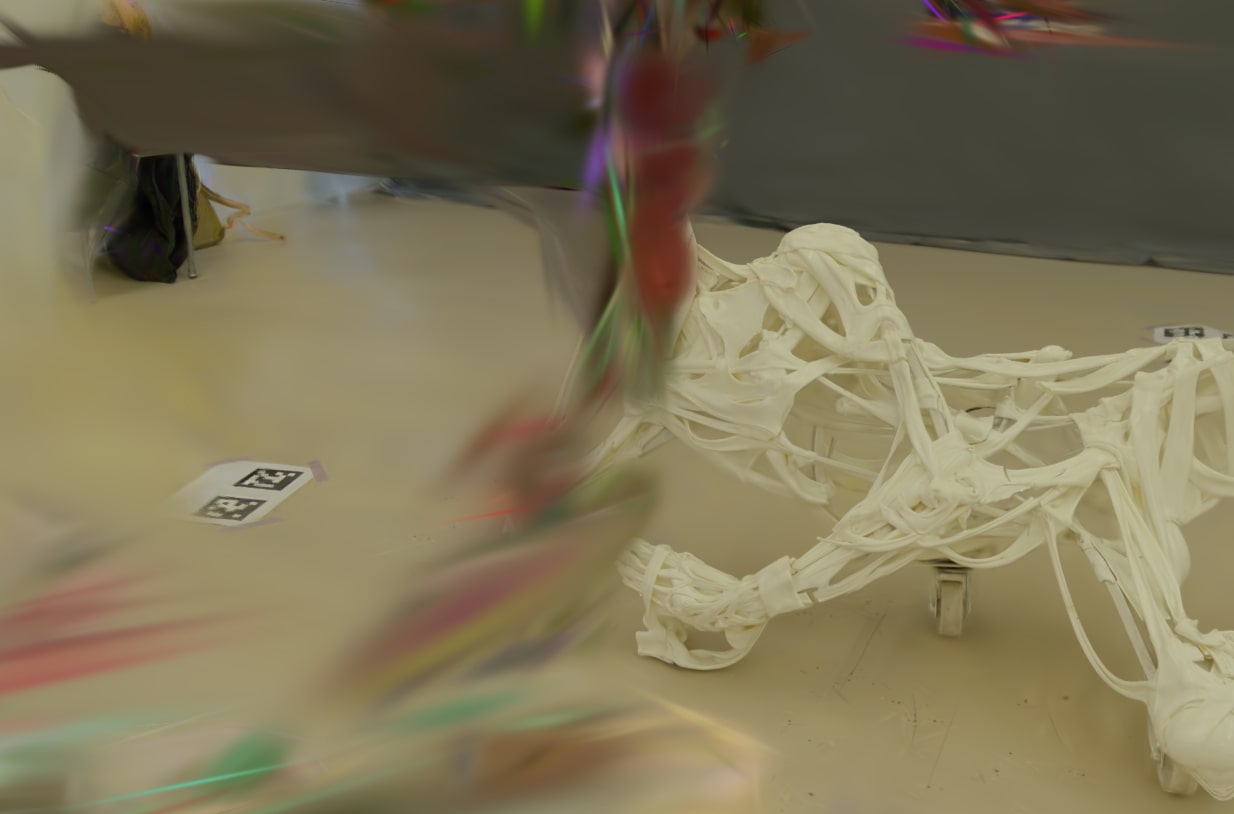} &
        \includegraphics[width=0.33\linewidth]{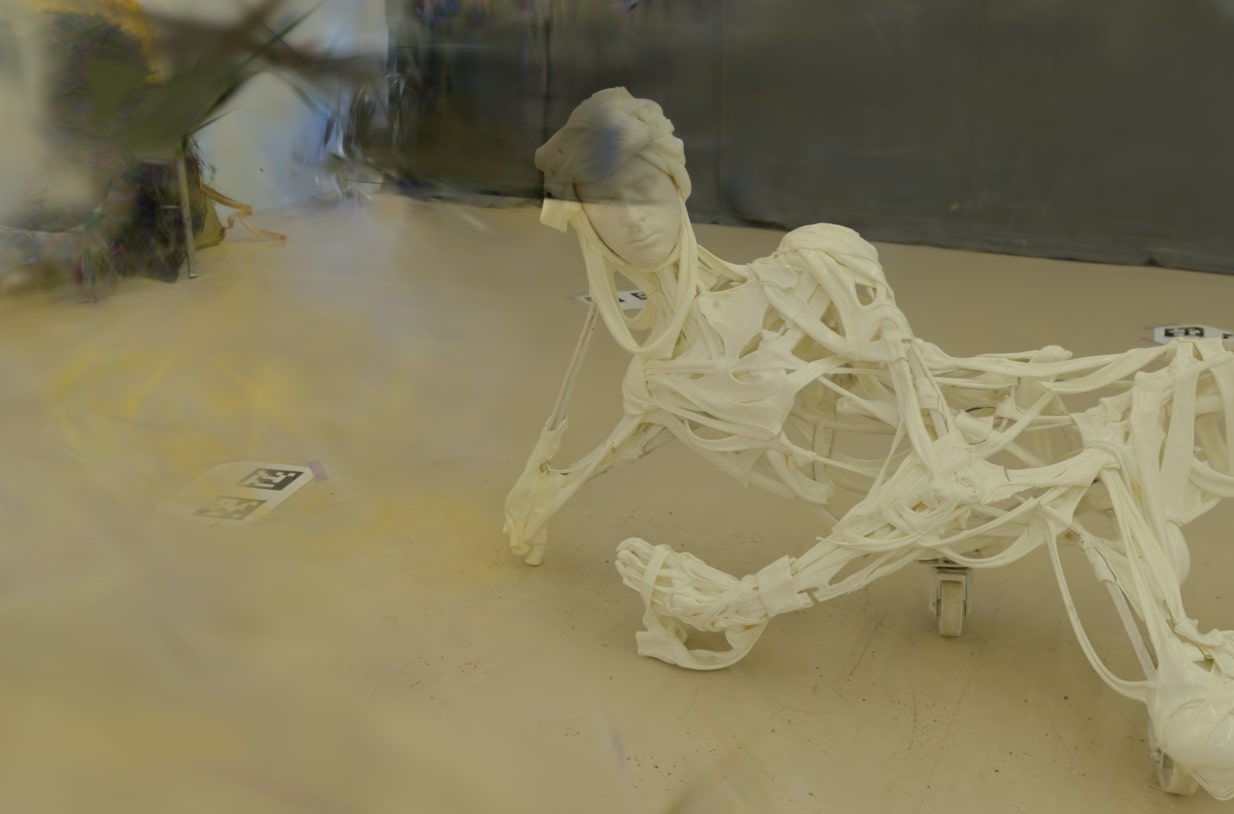} & 
        \includegraphics[width=0.33\linewidth]{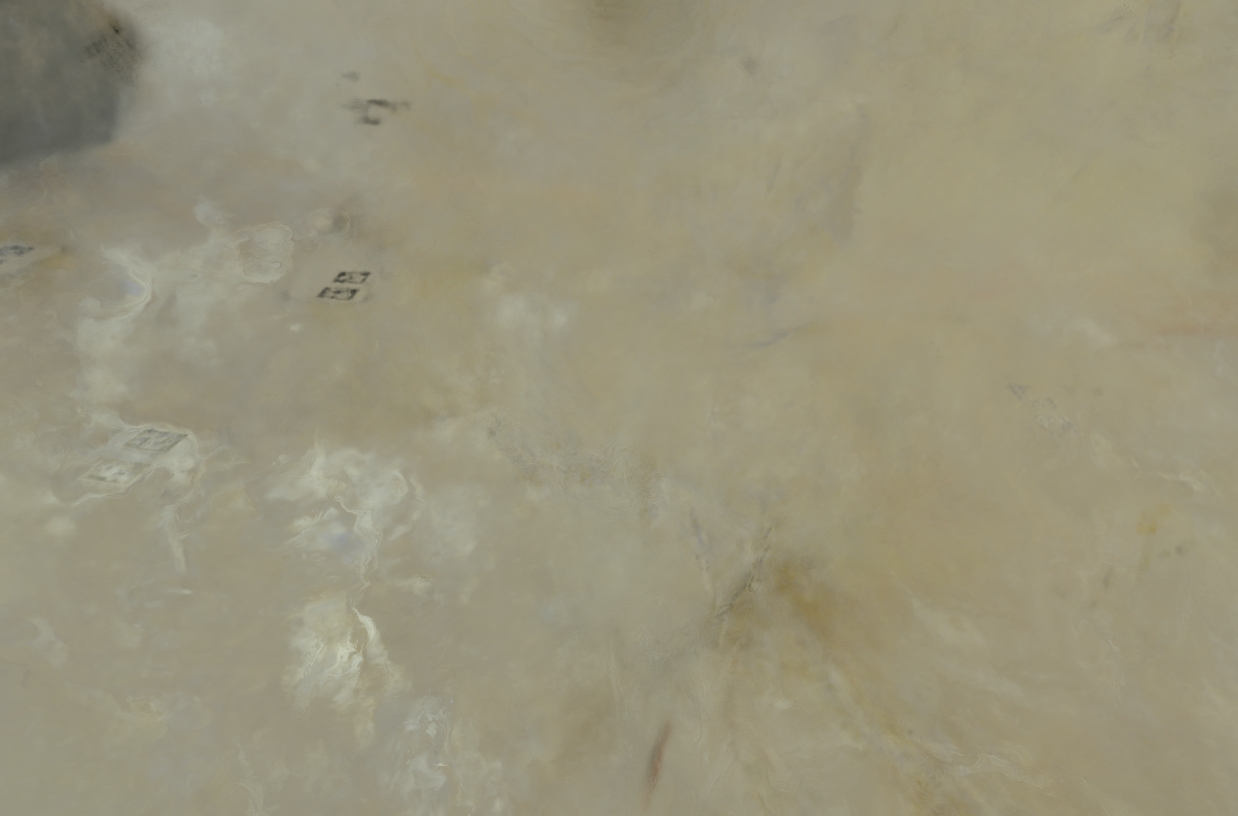}  
        
    \end{tabular}
    \caption{Examples of views from Paulina and Kacper scenes (sparse training views) in 2DGS, Mip-Splatting and Nerfacto which showcases a significant amount of artifacts which greatly disturbs an object of interest.}
    \label{fig:sparse_views}
\end{figure}

Multiple methods failed at reconstructing correctly Paulina and Kacper scenes with sparse views. For fair comparison, we used the same frame throughout every method, however, multiple methods failed to correctly reconstruct every point of view in testing sets creating a significant amount of artifacts located on the object of interest. The methods that had at least one frame failed were 2DGS, Mip-Splatting, Nerfacto (which failed on almost every image). Additionally, Instant-NGP OOM on Paulina scene and on multiple renders in Kacper. We showcased these exemplary problematic frames in Figure \ref{fig:sparse_views}. 

\section{Conclusion}
\our{} represents a novel dataset designed to address the limitations of existing benchmarks in 3D reconstruction from 2D images.  Unlike previous datasets that often feature idealized synthetic or meticulously captured data, \our{} specifically incorporates realistic challenges commonly encountered by non-expert users, such as dynamic backgrounds, a low number of images, and inconsistencies in image acquisition parameters. We think that recent advancements in 3D reconstruction methodologies have progressed to a stage where these real-world complexities should be effectively addressed.  The introduction of \our{} provides a valuable tool for evaluating and advancing the robustness of 3D reconstruction models under these conditions.

\paragraph{Limitations and Future Work}
The dataset size is the biggest limitation of \our{}. Future Work could include more sculptures, presumably from different artists. Additionally, every scene could contain captures showcasing every combination of the mentioned challenges.

\section{Acknoledgments}
We are deeply grateful to the amazing artist Paweł Althamer for its wonderful art and his curtosy in enabling us to scan and share his work. We also want to thank Foksal Gallery Foundation for providing us with space for scanning, a warm welcome, and helping in contacting the artist. 

\bibliographystyle{unsrt}

\begin{thebibliography}{10}

\bibitem{mildenhall2020nerf}
Ben Mildenhall, Pratul~P. Srinivasan, Matthew Tancik, Jonathan~T. Barron, Ravi
  Ramamoorthi, and Ren Ng.
\newblock Nerf: Representing scenes as neural radiance fields for view
  synthesis.
\newblock In {\em ECCV}, 2020.

\bibitem{kerbl3Dgaussians}
Bernhard Kerbl, Georgios Kopanas, Thomas Leimk{\"u}hler, and George Drettakis.
\newblock 3d gaussian splatting for real-time radiance field rendering.
\newblock {\em ACM Transactions on Graphics}, 42(4), July 2023.

\bibitem{schoenberger2016mvs}
Johannes~Lutz Sch\"{o}nberger, Enliang Zheng, Marc Pollefeys, and Jan-Michael
  Frahm.
\newblock Pixelwise view selection for unstructured multi-view stereo.
\newblock In {\em European Conference on Computer Vision (ECCV)}, 2016.

\bibitem{schoenberger2016sfm}
Johannes~Lutz Sch\"{o}nberger and Jan-Michael Frahm.
\newblock Structure-from-motion revisited.
\newblock In {\em Conference on Computer Vision and Pattern Recognition
  (CVPR)}, 2016.

\bibitem{multinerf2022}
Ben Mildenhall, Dor Verbin, Pratul~P. Srinivasan, Peter Hedman, Ricardo
  Martin-Brualla, and Jonathan~T. Barron.
\newblock {MultiNeRF}: {A} {Code} {Release} for {Mip-NeRF} 360, {Ref-NeRF}, and
  {RawNeRF}, 2022.

\bibitem{verbin2022refnerf}
Dor Verbin, Peter Hedman, Ben Mildenhall, Todd Zickler, Jonathan~T. Barron, and
  Pratul~P. Srinivasan.
\newblock {Ref-NeRF}: Structured view-dependent appearance for neural radiance
  fields.
\newblock {\em CVPR}, 2022.

\bibitem{pumarola2020d}
Albert Pumarola, Enric Corona, Gerard Pons-Moll, and Francesc Moreno-Noguer.
\newblock {D-NeRF: Neural Radiance Fields for Dynamic Scenes}.
\newblock In {\em Proceedings of the IEEE/CVF Conference on Computer Vision and
  Pattern Recognition}, 2020.

\bibitem{Hong_2022_CVPR}
Yang Hong, Bo~Peng, Haiyao Xiao, Ligang Liu, and Juyong Zhang.
\newblock Headnerf: A real-time nerf-based parametric head model.
\newblock In {\em Proceedings of the IEEE/CVF Conference on Computer Vision and
  Pattern Recognition (CVPR)}, pages 20374--20384, June 2022.

\bibitem{Deng_2022_CVPR}
Kangle Deng, Andrew Liu, Jun-Yan Zhu, and Deva Ramanan.
\newblock Depth-supervised nerf: Fewer views and faster training for free.
\newblock In {\em Proceedings of the IEEE/CVF Conference on Computer Vision and
  Pattern Recognition (CVPR)}, pages 12882--12891, June 2022.

\bibitem{wu20234dgaussians}
Guanjun Wu, Taoran Yi, Jiemin Fang, Lingxi Xie, Xiaopeng Zhang, Wei Wei, Wenyu
  Liu, Qi~Tian, and Wang Xinggang.
\newblock 4d gaussian splatting for real-time dynamic scene rendering.
\newblock {\em arXiv preprint arXiv:2310.08528}, 2023.

\bibitem{huang2023sc}
Yi-Hua Huang, Yang-Tian Sun, Ziyi Yang, Xiaoyang Lyu, Yan-Pei Cao, and Xiaojuan
  Qi.
\newblock Sc-gs: Sparse-controlled gaussian splatting for editable dynamic
  scenes.
\newblock {\em arXiv preprint arXiv:2312.14937}, 2023.

\bibitem{waczynska2024d}
Joanna Waczy{\'n}ska, Piotr Borycki, Joanna Kaleta, S{\l}awomir Tadeja, and
  Przemys{\l}aw Spurek.
\newblock D-miso: Editing dynamic 3d scenes using multi-gaussians soup.
\newblock {\em arXiv preprint arXiv:2405.14276}, 2024.

\bibitem{Huang2DGS2024}
Binbin Huang, Zehao Yu, Anpei Chen, Andreas Geiger, and Shenghua Gao.
\newblock 2d gaussian splatting for geometrically accurate radiance fields.
\newblock In {\em SIGGRAPH 2024 Conference Papers}. Association for Computing
  Machinery, 2024.

\bibitem{MALARZ2025104273}
Dawid Malarz, Weronika Smolak-Dyżewska, Jacek Tabor, Sławomir Tadeja, and
  Przemysław Spurek.
\newblock Gaussian splatting with nerf-based color and opacity.
\newblock {\em Computer Vision and Image Understanding}, 251:104273, 2025.

\bibitem{blender}
Blender~Online Community.
\newblock {\em Blender - a 3D modelling and rendering package}.
\newblock Blender Foundation, Stichting Blender Foundation, Amsterdam, 2018.

\bibitem{PhotogrammetryForModeling2021}
Sebastian Bullinger, Christoph Bodensteiner, and Michael Arens.
\newblock A photogrammetry-based framework to facilitate image-based modeling
  and automatic camera tracking.
\newblock {\em International Conference on Computer Graphics Theory and
  Applications}, 2021.

\bibitem{kiri}
Kiri-Innovation.
\newblock 3dgs render blender addon by kiri engine.
\newblock \url{https://github.com/Kiri-Innovation/3dgs-render-blender-addon},
  2025.

\bibitem{mildenhall2019llff}
Ben Mildenhall, Pratul~P. Srinivasan, Rodrigo Ortiz-Cayon, Nima~Khademi
  Kalantari, Ravi Ramamoorthi, Ren Ng, and Abhishek Kar.
\newblock Local light field fusion: Practical view synthesis with prescriptive
  sampling guidelines.
\newblock {\em ACM Transactions on Graphics (TOG)}, 2019.

\bibitem{reizenstein21co3d}
Jeremy Reizenstein, Roman Shapovalov, Philipp Henzler, Luca Sbordone, Patrick
  Labatut, and David Novotny.
\newblock Common objects in 3d: Large-scale learning and evaluation of
  real-life 3d category reconstruction.
\newblock In {\em International Conference on Computer Vision}, 2021.

\bibitem{HPPFDB18}
Peter Hedman, Julien Philip, True Price, Jan-Michael Frahm, George Drettakis,
  and Gabriel Brostow.
\newblock Deep blending for free-viewpoint image-based rendering.
\newblock {\em ACM Transactions on Graphics (SIGGRAPH Asia Conference
  Proceedings)}, 37(6), November 2018.

\bibitem{Yu2024GOF}
Zehao Yu, Torsten Sattler, and Andreas Geiger.
\newblock Gaussian opacity fields: Efficient high-quality compact surface
  reconstruction in unbounded scenes.
\newblock {\em arXiv:2404.10772}, 2024.

\bibitem{nerfstudio}
Matthew Tancik, Ethan Weber, Evonne Ng, Ruilong Li, Brent Yi, Justin Kerr,
  Terrance Wang, Alexander Kristoffersen, Jake Austin, Kamyar Salahi, Abhik
  Ahuja, David McAllister, and Angjoo Kanazawa.
\newblock Nerfstudio: A modular framework for neural radiance field
  development.
\newblock In {\em ACM SIGGRAPH 2023 Conference Proceedings}, SIGGRAPH '23,
  2023.

\bibitem{kheradmand20243d}
Shakiba Kheradmand, Daniel Rebain, Gopal Sharma, Weiwei Sun, Yang-Che Tseng,
  Hossam Isack, Abhishek Kar, Andrea Tagliasacchi, and Kwang~Moo Yi.
\newblock 3d gaussian splatting as markov chain monte carlo.
\newblock In {\em Advances in Neural Information Processing Systems (NeurIPS)},
  2024.
\newblock Spotlight Presentation.

\bibitem{mueller2022instant}
Thomas M\"uller, Alex Evans, Christoph Schied, and Alexander Keller.
\newblock Instant neural graphics primitives with a multiresolution hash
  encoding.
\newblock {\em ACM Trans. Graph.}, 41(4):102:1--102:15, July 2022.

\end{thebibliography}

\end{document}